\documentclass[journal]{IEEEtran}
\usepackage{float}
\usepackage{amssymb}
\usepackage{amsthm}
\usepackage{amsfonts}
\usepackage{mathrsfs}
\usepackage{graphicx}
\usepackage{amsmath,bm}
\usepackage{algorithmic}
\usepackage{algorithm}
\usepackage{setspace}
\usepackage{color}
\usepackage{subfig}
\usepackage{caption}
\usepackage{url}
\usepackage{booktabs}
\usepackage{multirow}
\usepackage{arydshln}

\begin{document}

\title{S$^2$FS: Spatially-Aware Separability-Driven Feature Selection in Fuzzy Decision Systems}

\author{Suping Xu, Chuyi Dai, Ye Liu,\\
Lin Shang,~\IEEEmembership{Member,~IEEE}, Xibei Yang, and Witold Pedrycz,~\IEEEmembership{Life Fellow,~IEEE}

\thanks{S. Xu, C. Dai, Y. Liu, and W. Pedrycz are with the Department of Electrical and Computer Engineering, University of Alberta, Edmonton, AB T6G 2R3, Canada. (E-mail: supingxu@yahoo.com;suping2@ualberta.ca, cdai4@ualberta.ca, ye35@ualberta.ca, wpedrycz@ualberta.ca)}
\thanks{L. Shang is with the School of Computer Science, Nanjing University, Nanjing 210023, China, and also with the State Key Laboratory for Novel Software Technology, Nanjing University, Nanjing 210023, China. (E-mail: shanglin@nju.edu.cn)}
\thanks{X. Yang is with the School of Computer, Jiangsu University of Science and Technology, Zhenjiang 212003, China. (E-mail: jsjxy\_yxb@just.edu.cn)}

\thanks{Manuscript received X X, 2025; revised X X, 2025.}}

\markboth{IEEE TRANSACTIONS ON X,~Vol.~X, No.~X, X~2025}%
{Shell \MakeLowercase{\textit{et al.}}: Bare Demo of IEEEtran.cls for IEEE Journals}

\maketitle

\begin{abstract}
Feature selection is crucial for fuzzy decision systems (FDSs), as it identifies informative features and eliminates rule redundancy, thereby enhancing predictive performance and interpretability. Most existing methods either fail to directly align evaluation criteria with learning performance or rely solely on non-directional Euclidean distances to capture relationships among decision classes, which limits their ability to clarify decision boundaries. However, the spatial distribution of instances has a potential impact on the clarity of such boundaries. Motivated by this, we propose \textbf{S}patially-aware \textbf{S}eparability-driven \textbf{F}eature \textbf{S}election (S$^2$FS), a novel framework for FDSs guided by a spatially-aware separability criterion. This criterion jointly considers within-class compactness and between-class separation by integrating scalar-distances with spatial directional information, providing a more comprehensive characterization of class structures. S$^2$FS employs a forward greedy strategy to iteratively select the most discriminative features. Extensive experiments on ten real-world datasets demonstrate that S$^2$FS consistently outperforms eight state-of-the-art feature selection algorithms in both classification accuracy and clustering performance, while feature visualizations further confirm the interpretability of the selected features.
\end{abstract}


%
\IEEEpeerreviewmaketitle

\section{Introduction}
\label{sec:1}

\IEEEPARstart{F}{uzzy} decision systems (FDSs) provide an effective paradigm for classification under uncertainty by modeling gradual transitions and ambiguous boundaries. Their performance, however, heavily depends on the quality of input features. Redundant or noisy features inflate the fuzzy rule base and obscure decision boundaries, resulting in increased complexity, degraded predictive performance, and reduced model transparency. Consequently, as a crucial preprocessing step for improving efficiency, accuracy, and interpretability \cite{YangQ2025,ZhaoY2025,ZhangX2024} of learning models, feature selection is particularly important for FDSs, often referred to as fuzzy feature selection.

Fuzzy feature selection usually considers both the structural characteristics of fuzzy representations and the semantic ambiguity of fuzzy data. Existing solutions generally adopt three types of evaluation criteria: fuzzy entropy-based, fuzzy similarity/correlation-based, and fuzzy rough set-based. To address the combinatorial complexity of subset selection, many of these are integrated with heuristic or meta-heuristic search strategies. Despite notable progress, a persistent challenge remains: how to ensure that feature evaluation criteria, such as those based on uncertainty characterized by fuzzy entropy or fuzzy rough approximations, correlate well with the actual classification performance of the resulting feature subsets in FDSs, independently of any specific classifier \cite{WangC2022,XuS15250}. In practice, uncertainty minimization does not always guarantee consistently improved classification performance, revealing a potential misalignment between the evaluation criteria and the learning objective.

To bridge this gap, recent research has shifted toward the development of performance-driven evaluation criteria and selection frameworks. For example, Wang et al. \cite{WangC2022} proposed an inner product dependency (IPD) measure to simultaneously maximize the dependency of an instance on its decision class and minimize classification error. Hu et al. \cite{HuM2022} introduced a separability-based measure that explicitly models the relationship between instances and decision classes by enhancing intra-class aggregation and inter-class dispersion. More recently, Xu et al. \cite{XuS15250} presented a highly scalable margin-aware fuzzy rough feature selection (MAFRFS) framework. By jointly considering within-class margins (compactness) and between-class margins (separation), MAFRFS not only effectively reduces the uncertainty of classification tasks but also guides the feature selection process toward more separable and discriminative class structures.

Although the above methods consider structural relationships in the feature space, including both the relationships among instances within each decision class and those between class centroids, they are fundamentally based on scalar-distance metrics such as Euclidean distance \cite{WangC2022,XuS15250} or membership degrees \cite{HuM2022} derived from it. These magnitude-only criteria, while effective for measuring separation, are inherently insensitive to the directional distribution of instances. As a result, they fail to capture how instances are positioned relative to class centroids, for example, whether they concentrate in particular regions or directions around the centroid, thereby overlooking critical spatial information needed for more accurately characterizing class separability.

\begin{figure}[!t]
\centering
\subfloat[]{\includegraphics[width=0.24\textwidth]{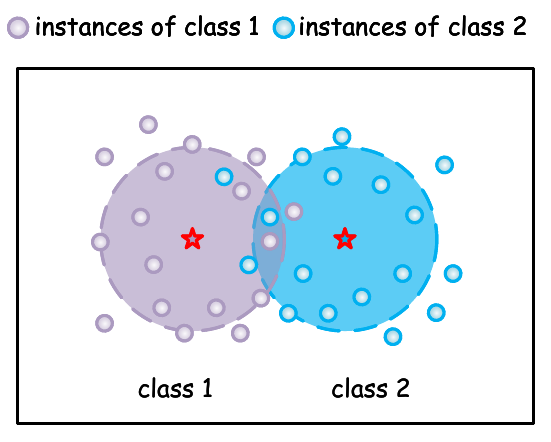}%
\label{fig:1a}}
\hfil
\subfloat[]{\includegraphics[width=0.24\textwidth]{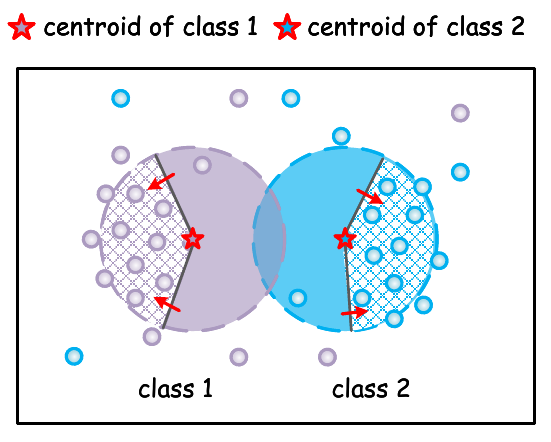}%
\label{fig:1b}}
\caption{Spatial distribution of instances from two decision classes. (a) \texttt{Isotropic distribution}. (b) \texttt{Two-sided concentrated distribution}.}
\label{fig:1}
\end{figure}

Consider the illustrative example shown in Fig. \ref{fig:1}, which presents two types of spatial distributions of instances from two decision classes: (a) an isotropic distribution and (b) a two-sided concentrated distribution. In Subfigs. 1(a) and 1(b), the centroid of each class is fixed at the same location. As a result, the inter-class dispersion \cite{HuM2022} or between-class margin \cite{XuS15250}, measured by the Euclidean distance between the two centroids, remains identical. Likewise, for each class, the intra-class aggregation \cite{HuM2022}, defined as the sum of membership degrees of instances to their class, or the within-class margin \cite{XuS15250}, defined as the mean distance of instances to the centroid, also remains the same across the two distributions. From the perspective of scalar-distance metrics, therefore, the structural relationships among instances appear equivalent under the two distributions. However, as Fig. \ref{fig:1} illustrates, the actual class separability differs: in the isotropic distribution (Subfigs. 1(a)), instances are uniformly spread around their centroid, producing a dense region of overlap between classes, whereas in the two-sided concentrated distribution (Subfigs. 1(b)), instances cluster along specific directions, thereby reducing inter-class overlap and yielding higher class separability. This contrast reveals that magnitude-only criteria overlook directional patterns of instances within classes and thus cannot adequately reflect the true degree of class separability. This limitation motivates the development of new separability criteria that explicitly incorporate spatial directional information into their quantification.

In this paper, we introduce a novel spatially-aware separability criterion that unifies within-class compactness and between-class separation. By jointly leveraging scalar-distance metrics and spatial directional information, this criterion provides a more comprehensive characterization of the structural relationships among instances. To effectively identify informative features, we adopt a forward greedy strategy and develop the \textbf{S}patially-aware \textbf{S}eparability-driven \textbf{F}eature \textbf{S}election (S$^2$FS) framework for FDSs. At each iteration, S$^2$FS selects the features that most effectively produce clear decision class boundaries. Specifically, we quantify within-class compactness with two complementary terms: the mean Euclidean distance from instances to their own class centroid, reflecting clustering tightness; and a directional consistency term that measures the angular deviation between each instance-to-own-centroid vector and the directions toward other class centroids, penalizing instances whose directions to their own centroid exhibit large misalignment with the directions toward other class centroids. Meanwhile, we quantify between-class separation with two complementary terms: the mean Euclidean distance from class centroids to their nearest centroid of another class, capturing overall separation across classes; and a directional discrepancy term that measures the angular deviation between each class-centroid–to–nearest-neighbor vector and the directions toward other class centroids, penalizing all other classes that exhibit substantial overlap with the nearest neighbor of a given class. Then, we formulate an objective that explicitly maximizes the ratio of between-class separation to within-class compactness, referred to as the spatially-aware separability criterion. A forward greedy strategy is employed to iteratively select the most discriminative features, progressively building a subset that best enhances class separability. Extensive experiments on eight publicly available high-dimensional datasets and two face recognition datasets (ORL and Yale) demonstrate that S$^2$FS achieves superior performance over eight state-of-the-art feature selection algorithms, as evaluated by both classification accuracy and clustering normalized mutual information. In addition, feature visualizations on the two face recognition datasets further validate the interpretability of the features selected by S$^2$FS.

To the best of our knowledge, this is the first work to integrate scalar-distance and directional information into modeling within-class compactness and between-class separation in FDSs, enabling a more separable and discriminative characterization of class structures for feature selection. In summary, our contributions are as follows.
\begin{itemize}
    \item We propose a spatially-aware separability criterion that jointly leverages distance and spatial directional information to effectively capture class separability.
    \item We develop a forward greedy feature selection framework (S$^2$FS) that exploits this criterion to identify the most informative features.
    \item We demonstrate the effectiveness and interpretability of S$^2$FS through extensive experiments and visualizations on ten real-world datasets.
\end{itemize}

\section{Related Works}
\label{sec:2}
Feature selection in FDSs should simultaneously consider the structural characteristics of fuzzy representations and the semantic ambiguity inherent in fuzzy data. Existing methods can generally be grouped into three categories, depending on the evaluation criteria they adopt. \textit{Fuzzy entropy-based criteria} aim to quantify the uncertainty associated with individual features using measures such as fuzzy entropy \cite{WangZ2023,WanJ2023}, fuzzy joint entropy \cite{WangZ2023,WanJ2023,DaiJ2024}, and fuzzy conditional entropy \cite{WangZ2023,WanJ2023,ZhangX2016,ZhangX2020,YangY2024}. Features with lower entropy values are typically considered more informative, enabling the elimination of redundant and irrelevant features in uncertain environments. \textit{Fuzzy similarity/correlation-based criteria} assess the dependency between features and decision classes using measures such as fuzzy mutual information \cite{WangZ2023,WanJ2023,WanJ2021} and fuzzy correlation coefficients. These methods are capable of capturing complex, nonlinear, and ambiguous relationships that traditional crisp correlation measures often fail to detect. \textit{Fuzzy rough set-based criteria} integrate fuzzy set theory with rough set approximations to simultaneously handle gradual membership uncertainty and boundary-region ambiguity. Specifically, lower and upper approximation operators characterize the certain and possible membership degrees of instances with respect to decision classes. Representative techniques include fuzzy rough dependency-based ranking \cite{JensenR2009,WangC2022,HuangZ2022,AnS2023}, fuzzy rough approximate quality evaluation \cite{HuQ2011,XuS2016}, and fuzzy rough discernibility matrix construction \cite{DaiJ2018,ChenJ2025FSS}.

Although these criteria provide effective evaluation of individual features through specific fuzzy measures, they often face challenges in high-dimensional settings because of the combinatorial complexity and the time-consuming nature of subset selection. To address this issue, heuristic and meta-heuristic search strategies have been introduced, which systematically explore the feature space by embedding fuzzy measures based criteria into optimization frameworks. Heuristic methods, such as greedy algorithms and recursive feature elimination, aim to yield fast, locally optimal solutions. In contrast, meta-heuristic methods, including genetic algorithms, particle swarm optimization, and ant colony optimization, simulate natural or social behaviors to conduct global searches while balancing fuzzy relevance and redundancy. These search strategies form the backbone of flexible feature selection frameworks that align closely with practical requirements, such as computational efficiency \cite{YangY2024,DaiJ2018,XiaD2023}, feature stability \cite{JiangZ2021}, and adaptability to diverse learning scenarios, e.g., online learning \cite{ZhangX2020,HuangW2023,ZhangC2025}, multi-label learning \cite{XuS2016,WangZ2024}, and weakly-supervised learning \cite{LiuK2023,ZhouN2025}.

In summary, studies on feature selection in FDSs have made substantial progress in both evaluation criteria and search strategies. Nevertheless, most existing methods fail to explicitly align evaluation criteria with learning performance. Although a few methods attempt to consider the performance by directly examining the structural characteristics of class distributions, they rely solely on non-directional Euclidean distances and ignores the spatial distribution of instances, which limits their ability to clarify decision boundaries. This drawback motivates us to develop a novel spatially-aware separability-driven feature selection framework that enhances class discrimination by integrating scalar-distances and spatial directional information into the separability criterion.

\section{Preliminaries}
\label{sec:3}

A FDS is formally defined as a 3-tuple $<U,F,L>$, where $U=\{x_1,x_2,\ldots,x_n\}$ represents a non-empty finite set of $n$ instances constituting the universe of discourse, $F=\{f_1,f_2,\ldots,f_m\}$ denotes a set of $m$ features characterizing these instances, and $L$ represents the decision function.

For each instance $x_{i} \in U$, its decision $L(x_i)\in \{l_1,l_2,\ldots,l_p\}$ is single and symbolic. Based on the decision function $L$, we establish an equivalence relation on $U$ as follows:
\begin{equation}
IND(L) = \{(x_i,x_j) \in U \times U: L(x_i)=L(x_j)\},
\label{eq:1}
\end{equation}
where $L$ partitions $U$ into $p$ boolean decision classes denoted by $U/IND(L)=\{LC_1,LC_2,\ldots,LC_p\}$. Each class $LC_q$ comprises all instances assigned to decision $l_q$, for $q = 1,2,\ldots,p$.

In practical applications, the belongingness of an instance to decision classes is often ambiguous, with fuzzy rather than crisp decision boundaries. Zadeh \cite{ZadehL1965} introduced the concept of fuzzy membership to quantify the degree to which each instance belongs to multiple decision classes. The fundamental principle is that the membership degree decreases as the distance between an instance and a decision class increases, and conversely, membership increases as this distance diminishes. Let $\mu_{iq}$ represent the fuzzy membership of instance $x_i$ to decision class $LC_q$, minimizing the following objective function formalizes this relationship:
\begin{equation}
Obj = \sum^{p}_{q=1}\sum^{n}_{i=1}\mu_{iq}dis(x_i,LC_q)^2
\label{eq:2}
\end{equation}
where $dis(x_i,LC_q)$ denotes the Euclidean distance from $x_i$ to decision class $LC_q$, with constraints $0 \leq \mu_{iq} \leq 1$ and $\sum^{p}_{q=1}\mu_{iq} = 1$ for all $q = 1,2,\ldots,p$ and $i = 1,2,\ldots,n$.

The method of Lagrange multipliers can be employed to transform the aforementioned constrained objective function into an unconstrained optimization problem as follows:
\begin{equation}
Obj' = \sum^{p}_{q=1}\sum^{n}_{i=1}\mu_{iq}dis(x_i,LC_q)^2 + \sum^{n}_{i=1}\lambda_i\Bigg(\sum^{p}_{q=1}\mu_{iq} - 1\Bigg)
\label{eq:3}
\end{equation}
where $\lambda_i$ are Lagrange multipliers for all $i = 1,2,\ldots,n$.

By the solution means similar to FCM clustering, it is not difficult to obtain:
\begin{equation}
\mu_{iq} = \frac{1}{\sum^{p}_{q^*=1}\Big(\frac{dis(x_i,LC_q)}{dis(x_i,LC_{q^*})}\Big)^2}
\label{eq:4}
\end{equation}

\section{Spatially-aware Separability-driven Feature Selection}

Given a FDS $<U,F,L>$, each $x_i \in U$ is an $m$-dimensional feature vector. By selecting a feature subset $F' \subseteq F$ with $m'$ features ($m' \leq m$), $x_i$ can be represented as a reduced $m'$-dimensional vector $\phi_{_{F'}}(x_i) = [f_{t_1}(x_{i}),f_{t_2}(x_{i}),\dots,f_{t_{m'}}(x_{i})]$, where $\{t_1,t_2,\ldots,t_{m'}\}$ are the indices of the selected features.

\subsection{Within-class Compactness}
The within-class compactness quantifies the degree of cohesion among instances within each decision class with respect to a feature subset $F' \subseteq F$. We discuss the within-class compactness by incorporating both distance and direction perspectives.

For any decision class $LC_q \in U/IND(L)$, where $q = 1,2,\ldots,p$, its centroid on feature subset $F'$ is defined as the mean feature vector of all instances belonging to $LC_q$:
\begin{equation}
\overline{LC}^{F'}_{q} = \frac{\sum^{n}_{i=1}\big(\phi_{_{F'}}(x_i)\omega_{iq}\big)}{\sum^{n}_{i=1}\omega_{iq}}
\label{eq:5}
\end{equation}
where $\omega_{iq}$ is an indicator function such that $\omega_{iq} = 1$ if $x_i \in LC_q$ and $\omega_{iq} = 0$ otherwise. The scatter of decision class $LC_q$ is then calculated as:
\begin{equation}
\sum^{n}_{i=1}\big({\Vert \phi_{_{F'}}(x_i) - \overline{LC}^{F'}_{q}\Vert}_{2}\omega_{iq}\big)
\label{eq:6}
\end{equation}

Furthermore, from a \textit{distance perspective}, the measure of within-class compactness of $U/IND(L)$ corresponding to any feature subset $F' \subseteq F$ is defined as:
\begin{equation}
\Theta^{F', L}_{dis} = \frac{1}{n}\sum^{p}_{q=1}\sum^{n}_{i=1}\big({\Vert \phi_{_{F'}}(x_i) - \overline{LC}^{F'}_{q}\Vert}_{2}\omega_{iq}\big)
\label{eq:7}
\end{equation}

From a direction perspective, we analyze the angular relationships between each instance and the centroids of all decision classes. For any instance $x_i \in LC_q$, we aim to maximize the directional consistency between the vector $\overrightarrow{\mathbf{V}_{iq}}$, which connects $x_i$ to its own class centroid, and the vector $\overrightarrow{\mathbf{V}_{iq'}}$, which links $x_i$ to the centroid of another class $LC_{q'}$, where $x_i \notin LC_{q'}$ and $q' = 1,2,\ldots,p$. In other words, the cosine similarity between $\overrightarrow{\mathbf{V}_{iq}}$ and $\overrightarrow{\mathbf{V}_{iq'}}$ is expected to increase. Given a feature subset $F'$, these vectors are defined as:
\begin{align}
\overrightarrow{\mathbf{V}_{iq}} = \overline{LC}^{F'}_{q} - \phi_{_{F'}}(x_i)\\
\overrightarrow{\mathbf{V}_{iq'}} = \overline{LC}^{F'}_{q'} - \phi_{_{F'}}(x_i)
\label{eq:9}
\end{align}

For a FDS containing only two decision classes, we seek to identify a feature subset $F'$ that maximizes the following cosine similarity:
\begin{equation}
d_{cos}(\overrightarrow{\mathbf{V}_{iq}},\overrightarrow{\mathbf{V}_{iq'}}) = \frac{\overrightarrow{\mathbf{V}_{iq}} \cdot \overrightarrow{\mathbf{V}_{iq'}}}{{\Vert\overrightarrow{\mathbf{V}_{iq}}\Vert}_{2} \cdot {\Vert \overrightarrow{\mathbf{V}_{iq'}} \Vert}_{2}}
\label{eq:10}
\end{equation}

However, in a FDS with more than two decision classes, the optimal positioning of any instance $x_i \in LC_q$ in the feature space is simultaneously influenced by all other classes $U/IND(L) \setminus LC_q$. The repelling effect exerted by each class $LC_{q'}$, where $q' = 1,2,\ldots,p$ and $q' \neq q$, often lead to conflicts when determining the optimal position of $x_i$. To address this, we introduce the fuzzy membership of $x_i \in LC_q$ to $LC_{q'}$, $q' = 1,2,\ldots,p$ and $q' \neq q$, as a means of quantifying the influence of $LC_{q'}$ on $x_i$. Specifically, a higher fuzzy membership $\mu_{iq'}$ indicates stronger influence from $LC_{q'}$ on $x_i$. Consequently, it is necessary to assign greater weight to the directional consistency between $\overrightarrow{\mathbf{V}_{iq}}$ and $\overrightarrow{\mathbf{V}_{iq'}}$.

For any instance $x_i \in LC_q$, considering all other classes $U/IND(L) \setminus LC_q$, we aim to identify a feature subset $F'$ that minimizes the follows:
\begin{equation}
\sum^{p}_{q=1}\Big(\sum^{p}_{q'=1}\mu_{iq'}\big(1 - d_{cos}(\overrightarrow{\mathbf{V}_{iq}},\overrightarrow{\mathbf{V}_{iq'}})\big)\Big)\omega_{iq}
\label{eq:11}
\end{equation}
where $\omega_{iq}$ is an indicator function such that $\omega_{iq} = 1$ if $x_i \in LC_q$ and $\omega_{iq} = 0$ otherwise.

To compute the fuzzy membership $\mu_{iq'}$, we replace the Euclidean distance from $x_i$ to $LC_{q'}$ in Eq.(\ref{eq:4}) with the Euclidean distance from $x_i$ to the centroid of $LC_{q'}$. Formally, $\mu_{iq'}$ can be calculated as follows:
\begin{equation}
\mu_{iq'} = \frac{1}{\sum^{p}_{q^*=1}\Big(\frac{{\Vert \phi_{_{F'}}(x_i) - \overline{LC}^{F'}_{q'} \Vert}_{2}}{{\Vert \phi_{_{F'}}(x_i) - \overline{LC}^{F'}_{q^*} \Vert}_{2}}\Big)^2}
\label{eq:12}
\end{equation}

Furthermore, from a \textit{direction perspective}, the measure of within-class compactness of $U/IND(L)$ corresponding to any feature subset $F' \subseteq F$ is defined as:
\begin{equation}
\Theta^{F', L}_{dir} = \frac{1}{n}\sum^{n}_{i=1}\sum^{p}_{q=1}\Big(\sum^{p}_{q'=1}\mu_{iq'}\big(1 - d_{cos}(\overrightarrow{\mathbf{V}_{iq}},\overrightarrow{\mathbf{V}_{iq'}})\big)\Big)\omega_{iq}
\label{eq:13}
\end{equation}

We integrate the within-class compactness from the distance perspective with that from the direction perspective:
\begin{equation}
\Theta^{F'}_L = \Theta^{F', L}_{dis} + \alpha \Theta^{F', L}_{dir} 
\label{eq:14}
\end{equation}
where $\alpha$ is a balancing parameter.

A smaller value of the within-class compactness measure indicates a higher degree of cohesion among instances within each decision class. An optimal feature subset $F' \subseteq F$ is expected to yield a smaller value of $\Theta^{F'}_L$.

\begin{figure}[!t]
\centering
\includegraphics[width=0.25\textwidth]{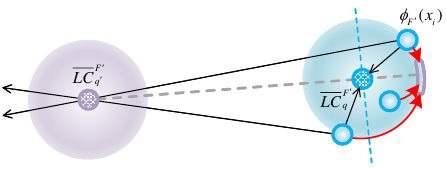}
\caption{An illustration of constructing within-class compactness from direction perspective}
\label{fig:2}
\end{figure}

\begin{figure}[!t]
\centering
\includegraphics[width=0.27\textwidth]{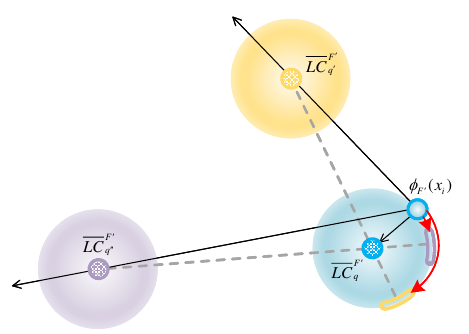}
\caption{An illustration of conflicts among multiple decision classes in constructing within-class compactness}
\label{fig:3}
\end{figure}

\subsection{Between-class Separation}
The between-class separation quantifies the degree of separation among instances belonging to distinct decision classes with respect to a feature subset $F' \subseteq F$. Well-defined class boundaries are crucial for enhancing the efficacy of classification, induction, and reasoning tasks in FDSs. Similarly, we analyze the between-class separation from both distance and direction perspectives.

Following previous work \cite{XuS15250}, two distance-based strategies are commonly used to measure between-class separation: a global strategy and a local strategy. The global strategy assesses each decision class’s deviation from the global centroid of all instances, whereas the local strategy considers pairwise deviations between two classes. However, both share a key limitation: when deviations are summarized, either relative to the global centroid or across class pairs, larger values can dominate the measure, causing smaller deviations for a given class to be overlooked.

To address this limitation, for each decision class $LC_q \in U/IND(L)$, where $q = 1,2,\ldots,p$, we identify its nearest neighboring class. Because nearest class pairs typically exhibit a higher degree of overlap, increasing the deviation between a class and its nearest neighbor is expected to improve overall between-class separation. Formally, we aim to identify a feature subset $F'$ that maximizes the follows:

\begin{equation}
\sum^{p}_{q'=1}\big({\Vert \overline{LC}^{F'}_{q} - \overline{LC}^{F'}_{q'}\Vert}_{2}\phi_{qq'}\big)
\label{eq:15}
\end{equation}
where $\phi_{qq'}$ is an indicator function such that $\phi_{qq'} = 1$ if $LC_{q'}$ is the nearest class to $LC_{q}$, and $\phi_{qq'} = 0$ otherwise.

Furthermore, from a \textit{distance perspective}, the measure of between-class separation of $U/IND(L)$ corresponding to any feature subset $F' \subseteq F$ is defined as:
\begin{equation}
\Lambda^{F', L}_{dis} = \frac{1}{p}\sum^{p}_{q=1}\sum^{p}_{q'=1}\big({\Vert \overline{LC}^{F'}_{q} - \overline{LC}^{F'}_{q'}\Vert}_{2}\phi_{qq'}\big)
\label{eq:16}
\end{equation}

From a direction perspective, we analyze the angular relationships between each decision class centroid and the centroids of all other classes. For any decision class $LC_q \in U/IND(L)$, let its nearest neighboring class be $LC_{q'}$, and any other class be $LC_{q''} \in U/IND(L) \setminus \{LC_q,LC_{q'}\}$. Our objective is to maximize the directional discrepancy between the vector $\overrightarrow{\mathbf{V}_{qq'}}$, which connects the centroid of $LC_{q}$ to the centroid of $LC_{q'}$, and the vector $\overrightarrow{\mathbf{V}_{qq''}}$, which links the centroid of $LC_{q}$ to the centroid of $LC_{q''}$. In other words, we aim to minimize the cosine similarity between $\overrightarrow{\mathbf{V}_{qq'}}$ and $\overrightarrow{\mathbf{V}_{qq''}}$. Given a feature subset $F'$, these vectors are defined as:
\begin{align}
\overrightarrow{\mathbf{V}_{qq'}} = \overline{LC}^{F'}_{q'} - \overline{LC}^{F'}_{q}\\
\overrightarrow{\mathbf{V}_{qq''}} = \overline{LC}^{F'}_{q''} - \overline{LC}^{F'}_{q}
\label{eq:18}
\end{align}

Similar to within-class compactness, for any decision class $LC_q \in U/IND(L)$, maximizing the directional discrepancy between $\overrightarrow{\mathbf{V}_{qq'}}$ and $\overrightarrow{\mathbf{V}_{qq''}}$ may lead to conflicts among different decision classes $U/IND(L) \setminus \{LC_q,LC_{q'}\}$. To address this, we introduce fuzzy membership to quantify the influence of each $LC_{q''}$, where $q'' = 1,2,\ldots,p$, $q'' \neq q$ and $q'' \neq q'$. Specifically, we consider the fuzzy membership $\overline{\mu}_{q'q''}$ of the centroid of $LC_{q'}$ with respect to $LC_{q''}$. 

A higher fuzzy membership $\overline{\mu}_{q'q''}$ indicates that the decision boundary between $LC_{q'}$ and $LC_{q''}$ is less distinct, suggesting that a greater weight should be assigned to the directional discrepancy between $\overrightarrow{\mathbf{V}_{qq'}}$ and $\overrightarrow{\mathbf{V}_{qq''}}$ to enhance the margin between decision classes. Conversely, if the decision boundary between $LC_{q'}$ and $LC_{q''}$ is clearer (i.e., $\overline{\mu}_{q'q''}$ is lower), a smaller weight is applied. Thus, we aim to identify a feature subset $F'$ that maximizes the follows:
\begin{equation}
\sum^{p}_{q''=1}\Big(\sum^{p}_{q'=1}\overline{\mu}_{q'q''}\big(1 - d_{cos}(\overrightarrow{\mathbf{V}_{qq'}},\overrightarrow{\mathbf{V}_{qq''}})\big)\phi_{qq'}\Big)(1-\tau_{qq''})
\label{eq:19}
\end{equation}
where $\phi_{qq'}$ and $\tau_{qq''}$ are indicator functions. Specifically, $\phi_{qq'} = 1$ if $LC_{q'}$ is the nearest decision class to $LC_{q}$, and $\phi_{qq'} = 0$ otherwise. Similarly, $\tau_{qq''} = 1$ if $LC_{q''}$ is either $LC_{q}$ itself or its nearest decision class, and $\tau_{qq''} = 0$ otherwise.

To compute the fuzzy membership $\overline{\mu}_{q'q''}$, we disregard the membership of $LC_{q'}$ to itself and directly set $\overline{\mu}_{q'q'} = 1$. Meanwhile, we ensure that the sum of the fuzzy memberships of $LC_{q'}$ to all other decision classes $LC_{q''}$, where $q'' = 1,2,\ldots,p$ and $q'' \neq q'$, equals $1$. Formally, $\overline{\mu}_{q'q''}$ is calculated as follows:
\begin{equation}
\overline{\mu}_{q'q''} = 
\begin{cases}
    1, & \text{if } q'' = q',\\
    \frac{1}{1 + \sum^{p}_{q^*=1,q^* \neq q'}\Big(\frac{{\Vert \overline{LC}^{F'}_{q'} - \overline{LC}^{F'}_{q''} \Vert}_{2}}{{\Vert \overline{LC}^{F'}_{q'} - \overline{LC}^{F'}_{q^*} \Vert}_{2}}\Big)^2}, & \text{if } q'' \neq q'.
\end{cases}
\label{eq:20}
\end{equation}

Furthermore, from a \textit{direction perspective}, the measure of between-class separation of $U/IND(L)$ corresponding to any feature subset $F' \subseteq F$ is defined as:
\begin{align}
\Lambda^{F', L}_{dir} = & \frac{1}{p}\sum^{p}_{q=1} \Bigg(\sum^{p}_{q''=1}\Big(\sum^{p}_{q'=1}\overline{\mu}_{q'q''} \nonumber\\
&\big(1 - d_{cos}(\overrightarrow{\mathbf{V}_{qq'}},\overrightarrow{\mathbf{V}_{qq''}})\big)\phi_{qq'}\Big)(1-\tau_{qq''})\Bigg)
\label{eq:21}
\end{align}

We integrate the between-class separation from the distance perspective with that from the direction perspective:
\begin{equation}
\Lambda^{F'}_L = \Lambda^{F', L}_{dis} + \beta \Lambda^{F', L}_{dir} 
\label{eq:22}
\end{equation}
where $\beta$ is a balancing parameter.

A larger value of the between-class separation measure indicates a higher degree of separation among instances belonging to distinct decision classes. An optimal feature subset $F' \subseteq F$ is expected to yield a larger value of $\Lambda^{F'}_L$.

\begin{figure}[!t]
\centering
\includegraphics[width=0.38\textwidth]{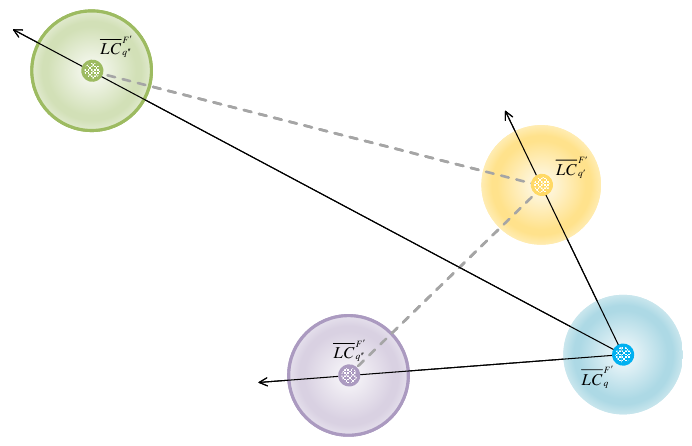}
\caption{An illustration of conflicts among multiple decision classes in constructing between-class separation}
\label{fig:4}
\end{figure}

\subsection{Spatially-aware Separability Criterion}

A feature subset $F' \subseteq F$ is considered beneficial in a FDS $<U,F,L>$ if it supports learning tasks such as classification or clustering by inducing well-defined decision boundaries through simultaneously enhancing within-class compactness and between-class separation. Specifically, a desirable subset should yield a smaller value of the within-class compactness measure $\Theta^{F'}_L$ and a larger value of the between-class separation measure $\Lambda^{F'}_L$. To achieve this, we propose a spatially-aware separability criterion that integrates $\Theta^{F'}_L$ and $\Lambda^{F'}_L$, taking into account both the scalar distances between instances and their spatial directional information.

Given a FDS $<U,F,L>$, $U/IND(L)=\{LC_1,LC_2,$ $\ldots,LC_p\}$, $\forall F' \subseteq F$, the spatially-aware separability criterion on $F'$ is defined as:
\begin{equation}
Sep^{F'}_L = \frac{\Lambda^{F'}_L}{\Theta^{F'}_L} = \frac{\Lambda^{F', L}_{dis} + \beta \Lambda^{F', L}_{dir}}{\Theta^{F', L}_{dis} + \alpha \Theta^{F', L}_{dir}}
\label{eq:23}
\end{equation}
where $\alpha$ and $\beta$ are balancing parameters.

A larger $Sep^{F'}_L$ indicates that both within-class compactness and between-class separation are moving in the desired direction, suggesting that the corresponding feature subset $F'$ is more effective for revealing well-structured groupings of instances.

\subsection{S$^2$FS Framework}

\label{sec:4.4}
During feature selection, we focus on the variations in within-class compactness and between-class separation. Leveraging the spatially-aware separability criterion introduced above, we quantitatively evaluate these variations to identify feature subsets $F' \subseteq F$ that achieve an effective trade-off between them.

Consider a FDS $<U,F,L>$, where $U/IND(L)=\{LC_1,LC_2,\ldots,LC_p\}$. For any feature subset $\forall F' \subseteq F$, if a candidate feature $f_t \in F {\setminus} F'$ is added to $F'$, the resulting improvement in separability can be computed as:
\begin{equation}
\varpi(f_t) = Sep^{F' \cup \{f_t\}}_L - Sep^{F'}_L
\label{eq:24}
\end{equation}

Building on this, Algorithm \ref{alg:algorithm1} introduces the \textbf{S}patially-aware \textbf{S}eparability-driven \textbf{F}eature \textbf{S}election (S$^2$FS) framework for FDSs. The framework adopts a forward greedy strategy that iteratively selects the most discriminative features.

\begin{algorithm}[tb]
\caption{S$^2$FS Framework}
\label{alg:algorithm1}
\textbf{Input}: $<U,F,L>$, number of target features $k$, balancing parameters $\alpha$ and $\beta$;\\
\textbf{Output}: One feature subset $F'$;
\begin{algorithmic}[1] 
\STATE $F'\leftarrow\emptyset$;
\STATE \textbf{do}
\STATE \quad Compute $\Theta^{F', L}_{dis}$, $\Theta^{F', L}_{dir}$ as Eqs.~\eqref{eq:7}, \eqref{eq:13};
\STATE \quad Compute $\Lambda^{F', L}_{dis}$, $\Lambda^{F', L}_{dir}$ as Eqs.~\eqref{eq:16}, \eqref{eq:21};
\STATE \quad Compute $Sep^{F'}_L$ as Eq.~\eqref{eq:23};
\STATE \quad \textbf{for} each $f_t \in F {\setminus} F'$ \textbf{do}
\STATE \quad \quad Compute $\Theta^{F' \cup \{f_t\}, L}_{dis}$, $\Theta^{F' \cup \{f_t\}, L}_{dir}$ as Eqs.~\eqref{eq:7}, \eqref{eq:13};
\STATE \quad \quad Compute $\Lambda^{F' \cup \{f_t\}, L}_{dis}$, $\Lambda^{F' \cup \{f_t\}, L}_{dir}$ as Eqs.~\eqref{eq:16}, \eqref{eq:21};
\STATE \quad \quad Compute $Sep^{F' \cup \{f_t\}}_L$ as Eq.~\eqref{eq:23};
\STATE \quad \quad Compute $\varpi(f_t)$ as Eq.~\eqref{eq:24};
\STATE \quad \textbf{end}
\STATE \quad Select a $f'_t$ as \\
\STATE \quad \quad $f'_t = \arg \max \{\varpi(f_t): \forall f_t \in F {\setminus} F'\}$;
\STATE \quad $F' \leftarrow F' \cup \{f'_t\}$;
\STATE \textbf{until} $|F'| = k$
\STATE \textbf{return} $F'$.
\end{algorithmic}
\end{algorithm}

\begin{figure}[!t]
\centering
\includegraphics[width=0.48\textwidth]{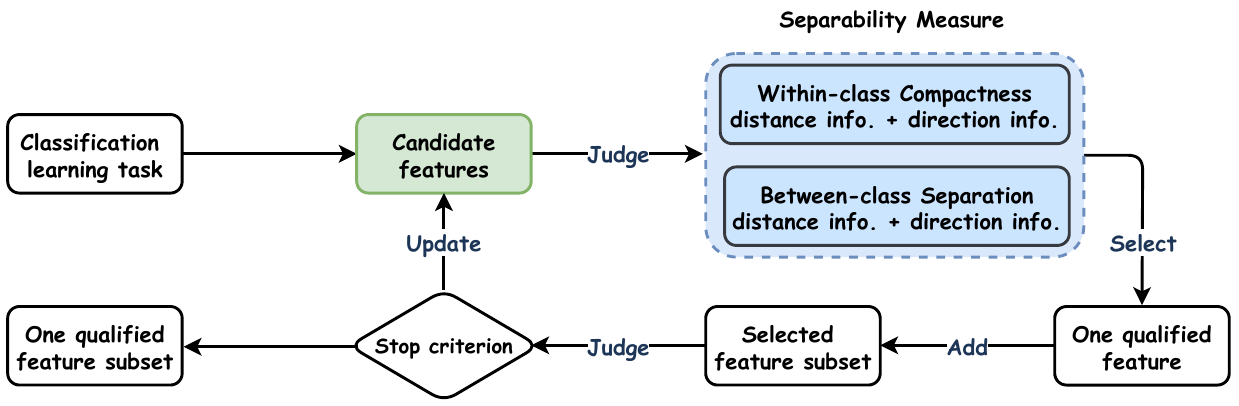}
\caption{Overall S$^2$FS framework and underlying processing}
\label{fig:5}
\end{figure}

S$^2$FS starts with an empty feature set and iteratively evaluates candidate features based on their contributions to within-class compactness and between-class separation, using spatially-aware separability criteria. Specifically, for each candidate feature $f_t \in F{\setminus}F'$, the algorithm first recomputes the within-class compactness of $U/IND(L)$ from both the \textit{distance} and \textit{direction} perspectives (Step 7). It then recomputes the between-class separation from these two perspectives (Step 8). Next, the spatially-aware separability score is derived (Step 9), and the separability gain $\varpi(f_t)$ is calculated (Step 10). The candidate feature $f_t$ that achieves the maximum separability gain is added to the selected subset $F'$ (Step 13). This procedure is repeated until the predefined number of features $k$ is reached, yielding the final feature subset $F'$. Fig. \ref{fig:5} presents the overall S$^2$FS framework and underlying processing.

\section{Experimental Analysis}

\subsection{Datasets}
To evaluate the effectiveness of S$^2$FS algorithm, we employ $10$ publicly available datasets, including eight small-sized high-dimensional datasets\footnote{\url{https://jundongl.github.io/scikit-feature/datasets.html}} and two face recognition datasets\footnote{\url{https://git-disl.github.io/GTDLBench/datasets/att_face_dataset/}}\footnote{\url{https://vision.ucsd.edu/datasets/yale-face-database}}. Tab.~\ref{tab:1} summarizes some statistics of these datasets, with $\#Inst.$, $\#Feat.$, and $\#Cls.$ denoting the number of instances, features, and decision classes, respectively.

\begin{table}[!t]
\caption{Statistics of the experimental datasets.}
\centering
\resizebox{0.46\textwidth}{!}{
\setlength{\tabcolsep}{4pt}
\begin{tabular}{ccccc}
\hline
Dataset & \#Inst. & \#Feat. & \#Cls. & Type \\
\hline
ALL-AML-3 & $72$ & $7129$ & $3$ & Small-sized high-dim\\
ALL-AML-4 & $72$ & $7129$ & $4$ & Small-sized high-dim\\
GLIOMA & $50$ & $4434$ & $4$ & Small-sized high-dim\\
Isolet & $1560$ & $617$ & $26$ & Small-sized high-dim\\
Lung & $203$ & $3312$ & $5$ & Small-sized high-dim\\
SRBCT & $84$ & $2308$ & $4$ & Small-sized high-dim\\
warpAR10P & $130$ & $2400$ & $10$ & Small-sized high-dim\\
warpPIE10P & $210$ & $2420$ & $10$ & Small-sized high-dim\\
\hline
ORL & $400$ & $1024$ & $40$ & Face recognition\\
Yale & $165$ & $1024$ & $15$ & Face recognition\\
\hline
\end{tabular}
}
\label{tab:1}
\end{table}

\subsection{Baselines and Settings}
In the experimental analysis, we compared eight feature selection algorithms: FDM (2009) \cite{JensenR2009}, MDP (2018) \cite{DaiJ2018}, RDSI (2021) \cite{WangC2021}, SFSS (2022) \cite{HuM2022}, IPD (2022) \cite{WangC2022}, N3Y (2023) \cite{LiuK2023ASOC}, FSNMER (2024) \cite{WuS2024}, and FGZEFSR (2025) \cite{YuanK2025}. Detailed descriptions of these algorithms can be found in \cite{XuS15250}. The implementations of all compared algorithms were obtained from the original authors.

Following the experimental settings in \cite{XuS15250}, three classifiers, i.e., Classification and Regression Tree (CART), Support Vector Machine (SVM), and $k$-Nearest Neighbors ($k$NN), were employed to evaluate the classification performance on the selected feature subsets. For SVM, a Gaussian kernel with a penalty factor of $C = 10$ was used, while the number of nearest neighbors in $k$NN was set to $5$. In addition, clustering analysis was performed using $k$-means on the selected feature subsets, and the normalized mutual information (NMI) \cite{EstevezP2009} was employed to quantitatively evaluate clustering performance. The number of clusters was set equal to the number of decision classes $p$. Since the initialization of cluster centroids can significantly affect clustering results, by default, initial centroids are randomly selected by uniformly sampling $p$ distinct instances without replacement from $U=\{x_1,x_2,\ldots,x_n\}$. To avoid unfair comparisons caused by such randomness, the initial cluster centroids \cite{YuanL2024} were uniformly determined as $\{x_{start},x_{\{start+int\}},\ldots,x_{\{start+(p-1)int\}}\}$, where $int = \lfloor \frac{n-1}{p-1} \rfloor$ and $start = \lfloor \frac{n-(p-1)int+1}{2} \rfloor$.

All experiments were conducted using 10-fold cross-validation to ensure statistical reliability. Since different feature selection algorithms may yield subsets of varying sizes, all outputs were converted into ranked feature lists based on their importance scores. To comprehensively evaluate the classification/clustering performance, the top-ranked features were incrementally included, starting from the top $1$ feature, then the top $2$ features, and so on, until all features were selected. Furthermore, the balancing parameters $\alpha$ and $\beta$ for S$^2$FS are tuned from \{$0.0100$, $0.0178$, $0.0316$, $0.0562$, $0.1000$, $0.1778$, $0.3162$, $0.5623$, $1.0000$\}.

\subsection{Results on Small-sized High-dimensional Data}

Tabs.~\ref{tab:2} and \ref{tab:3} summarize the classification and clustering performance of the proposed S$^2$FS algorithm compared with eight representative feature selection algorithms across eight small-sized high-dimensional datasets. For each dataset, both the maximum (Max) and the average (Ave) accuracy (for classification) or NMI (for clustering) are reported over 150 incrementally top-ranked feature subsets, with the best and second-best results highlighted in bold and underlined, respectively.

For the CART classifier, S$^2$FS achieves the best Max accuracy on five of the eight datasets (e.g., ALL-AML-3, ALL-AML-4, Isolet, SRBCT, and warpAR10P) and ranks second on two others (Lung and warpPIE10P), yielding top-two performance in 87.5\% of cases and demonstrating strong peak performance. Its average accuracy is slightly less dominant, ranking second on four datasets (ALL-AML-4, Isolet, warpAR10P, and warpPIE10P), but still reflecting relatively stable overall performance. SFSS performs better on average accuracy, surpassing S$^2$FS on five datasets (ALL-AML-3, ALL-AML-4, GLIOMA, Lung, and warpAR10P), but falls behind in Max accuracy (75\% top-two). RDSI shows competitive results on datasets such as ALL-AML-3 and warpPIE10P, while other algorithms exhibit only scattered strengths.

For the KNN classifier, S$^2$FS demonstrates clear superiority, ranking top-two in Max accuracy on all datasets and achieving the best results on seven (including ALL-AML-3, ALL-AML-4, GLIOMA, Isolet, Lung, SRBCT, and warpAR10P). It also secures three best (on ALL-AML-4, GLIOMA, and Lung) and five second-best average accuracies, reaching 100\% top-two coverage. This indicates that S$^2$FS consistently provides superior peak performance while maintaining high overall stability across datasets. SFSS remains competitive, attaining top-two results on half of the datasets for both Max and Ave accuracy, and outperforming S$^2$FS in average accuracy on ALL-AML-3, SRBCT, and warpAR10P. RDSI also shows satisfactory results, obtaining three best (or tied for best) and three second-best Max accuracies, with particularly notable performance on warpPIE10P. The algorithms N3Y, IPD, and FSNMER achieve only occasional top-two results, while FDM, MDP, and FGZEFSR fail to reach top-two performance in any case.

For the SVM classifier, S$^2$FS achieves the best Max accuracy on five datasets, namely ALL-AML-3, ALL-AML-4, GLIOMA, Lung, and SRBCT, and places top-two in average accuracy on four datasets (ALL-AML-3, ALL-AML-4, Lung, and SRBCT). This indicates strong peak performance but slightly lower consistency in average accuracy compared to its KNN results. RDSI is the strongest competitor, leading in Max accuracy on six datasets and achieving the best average accuracy on warpAR10P and warpPIE10P. N3Y often secures second-best Max accuracies and several top-two Ave results, indicating stable performance, while SFSS also achieves multiple top-two rankings across both Max and Ave accuracies. The remaining algorithms show limited competitiveness.

\begin{table*}[!t]
\caption{Maximum and average classification performance (Accuracy) with the top $1$, $2$,..., $150$ features across eight small-sized high-dimensional datasets.}
\centering
\resizebox{0.95\textwidth}{!}{
\setlength{\tabcolsep}{5pt}
\begin{tabular}{cc*{9}{cc}}
\hline
\multirow{2}{*}{\textbf{Cls.}} & 
\multirow{2}{*}{\textbf{SOTA}} & 
\multicolumn{2}{c}{\textbf{ALL-AML-3}} & 
\multicolumn{2}{c}{\textbf{ALL-AML-4}} & 
\multicolumn{2}{c}{\textbf{GLIOMA}} & 
\multicolumn{2}{c}{\textbf{Isolet}} & 
\multicolumn{2}{c}{\textbf{Lung}} & 
\multicolumn{2}{c}{\textbf{SRBCT}} & 
\multicolumn{2}{c}{\textbf{warpAR10P}} & 
\multicolumn{2}{c}{\textbf{warpPIE10P}} \\
& & \textit{Max} & \textit{Ave} & \textit{Max} & \textit{Ave} & \textit{Max} & \textit{Ave} & \textit{Max} & \textit{Ave} & \textit{Max} & \textit{Ave} & \textit{Max} & \textit{Ave} & \textit{Max} & \textit{Ave} & \textit{Max} & \textit{Ave} \\
\hline
\multirow{9}{*}{\rotatebox{90}{\textbf{CART}}} & SFSS & \textbf{0.9167} & \textbf{0.9004} & \underline{0.8750} & \textbf{0.8431} & \underline{0.7600} & \underline{0.6923} & 0.7577 & 0.7136 & \textbf{0.9310} & \textbf{0.9122} & 0.9036 & 0.8692 & \underline{0.7846} & \textbf{0.7401} & \textbf{0.8571} & 0.7957 \\
& FDM & 0.7917 & 0.7622 & 0.7917 & 0.7343 & 0.6600 & 0.5633 & 0.4923 & 0.4092 & 0.8522 & 0.8247 & 0.7711 & 0.7101 & 0.6000 & 0.4896 & 0.8048 & 0.7546 \\
& FSNMER & 0.6667 & 0.6166 & 0.6944 & 0.6247 & 0.6600 & 0.5152 & 0.7160 & 0.5943 & 0.8424 & 0.7927 & 0.6386 & 0.5428 & 0.6923 & 0.5963 & 0.6476 & 0.5672 \\
& IPD & \underline{0.8889} & 0.8223 & 0.7778 & 0.7307 & 0.7200 & 0.5980 & 0.5269 & 0.4750 & 0.8768 & 0.8443 & \textbf{0.9277} & \textbf{0.9093} & 0.5000 & 0.4510 & 0.7810 & 0.7329 \\
& MDP & 0.8056 & 0.7721 & 0.7917 & 0.7608 & \textbf{0.8400} & \textbf{0.7113} & 0.5731 & 0.5269 & 0.8522 & 0.8308 & 0.8072 & 0.7523 & 0.5462 & 0.5148 & 0.8238 & 0.7740 \\
& N3Y & 0.8750 & 0.8512 & 0.8333 & 0.7794 & 0.7200 & 0.6123 & \underline{0.7801} & \textbf{0.7351} & 0.8768 & 0.8580 & 0.8795 & 0.8413 & 0.7769 & 0.7174 & 0.8429 & 0.7936 \\
& RDSI & \textbf{0.9167} & \underline{0.8997} & 0.8611 & 0.8219 & 0.6600 & 0.5861 & 0.7667 & 0.6952 & \underline{0.9113} & \underline{0.8902} & \underline{0.9157} & \underline{0.8760} & 0.7692 & 0.6881 & \textbf{0.8571} & \textbf{0.7988} \\
& FGZEFSR & 0.8194 & 0.7547 & 0.7778 & 0.6794 & 0.6200 & 0.5049 & 0.7391 & 0.5891 & 0.8325 & 0.7801 & 0.6988 & 0.6280 & 0.4615 & 0.3727 & 0.7238 & 0.5789 \\
\cdashline{2-20}
& \textbf{S$^2$FS} & \textbf{0.9167} & 0.8693 & \textbf{0.8889} & \underline{0.8403} & 0.7200 & 0.6832 & \textbf{0.7917} & \underline{0.7189} & \underline{0.9113} & 0.8864 & \textbf{0.9277} & 0.8757 & \textbf{0.8000} & \underline{0.7364} & \underline{0.8524} & \underline{0.7984} \\
\hline
\multirow{9}{*}{\rotatebox{90}{\textbf{KNN}}} & SFSS & \textbf{0.9722} & \textbf{0.9505} & \textbf{0.9028} & \underline{0.8769} & \underline{0.8400} & 0.7879 & 0.8513 & 0.7695 & 0.9458 & 0.9322 & \textbf{1.0000} & \textbf{0.9875} & 0.7846 & \textbf{0.7525} & 0.9619 & 0.9257 \\
& FDM & 0.7639 & 0.6730 & 0.7083 & 0.6380 & 0.8200 & 0.7403 & 0.5244 & 0.4667 & 0.9212 & 0.9014 & 0.7349 & 0.6275 & 0.5308 & 0.4317 & 0.9476 & 0.9050 \\
& FSNMER & 0.6667 & 0.5994 & 0.6389 & 0.5799 & 0.7600 & 0.6871 & 0.7705 & 0.6237 & \underline{0.9557} & 0.9060 & 0.6988 & 0.6120 & 0.4769 & 0.3676 & 0.7619 & 0.6764 \\
& IPD & 0.7917 & 0.6840 & 0.8056 & 0.6768 & \textbf{0.8600} & \underline{0.7888} & 0.5372 & 0.5057 & 0.9261 & 0.8878 & 0.9157 & 0.7361 & 0.4923 & 0.4375 & 0.9333 & 0.8599 \\
& MDP & 0.8889 & 0.7185 & 0.7778 & 0.6755 & 0.8200 & 0.7557 & 0.6372 & 0.5732 & 0.9310 & 0.9006 & 0.8313 & 0.7284 & 0.5462 & 0.4614 & 0.9762 & 0.9397 \\
& N3Y & \underline{0.9306} & 0.9011 & \textbf{0.9028} & 0.8631 & 0.7800 & 0.7309 & \underline{0.8949} & \textbf{0.8283} & 0.9409 & 0.9112 & \underline{0.9759} & 0.9461 & 0.7538 & 0.7049 & 0.9714 & 0.9303 \\
& RDSI & \textbf{0.9722} & 0.9343 & \underline{0.8750} & 0.8087 & 0.7800 & 0.6953 & 0.8801 & 0.7823 & \underline{0.9557} & \underline{0.9329} & \textbf{1.0000} & 0.9671 & \underline{0.7923} & 0.7297 & \textbf{1.0000} & \textbf{0.9598} \\
& FGZEFSR & 0.8194 & 0.7009 & 0.7500 & 0.6686 & 0.7400 & 0.6047 & 0.8205 & 0.6258 & 0.9360 & 0.8785 & 0.7470 & 0.6085 & 0.3538 & 0.2709 & 0.5524 & 0.4668 \\
\cdashline{2-20}
& \textbf{S$^2$FS} & \textbf{0.9722} & \underline{0.9415} & \textbf{0.9028} & \textbf{0.8816} & \textbf{0.8600} & \textbf{0.8096} & \textbf{0.9019} & \underline{0.8053} & \textbf{0.9606} & \textbf{0.9466} & \textbf{1.0000} & \underline{0.9843} & \textbf{0.8154} & \underline{0.7514} & \underline{0.9857} & \underline{0.9513} \\
\hline
\multirow{9}{*}{\rotatebox{90}{\textbf{SVM}}} & SFSS & \textbf{0.9722} & \textbf{0.9545} & \underline{0.9167} & \underline{0.8994} & \textbf{0.8400} & 0.7553 & 0.9192 & 0.8142 & 0.9507 & 0.9350 & \textbf{1.0000} & \textbf{0.9884} & 0.8846 & 0.8573 & 0.9810 & 0.9475 \\
& FDM & 0.8056 & 0.7534 & 0.7778 & 0.7188 & 0.8000 & 0.7628 & 0.7179 & 0.6275 & 0.9409 & 0.9155 & 0.9759 & 0.8436 & 0.6923 & 0.6137 & 0.8714 & 0.8285 \\
& FSNMER & 0.7778 & 0.6815 & 0.7500 & 0.6520 & 0.8000 & 0.7291 & 0.8769 & 0.6935 & \underline{0.9557} & 0.9160 & 0.8795 & 0.7016 & 0.7769 & 0.6814 & 0.9000 & 0.8462 \\
& IPD & 0.8194 & 0.7297 & 0.8056 & 0.7634 & \underline{0.8200} & \textbf{0.7863} & 0.7378 & 0.6806 & 0.9163 & 0.8988 & 0.9639 & 0.9239 & 0.7231 & 0.6579 & 0.9381 & 0.8918 \\
& MDP & 0.8750 & 0.8094 & 0.8056 & 0.7507 & \underline{0.8200} & \underline{0.7807} & 0.8115 & 0.7556 & 0.9409 & 0.9161 & 0.9759 & 0.9173 & 0.7231 & 0.6631 & 0.9714 & 0.9348 \\
& N3Y & \underline{0.9444} & 0.9115 & \underline{0.9167} & 0.8844 & 0.7800 & 0.7280 & \underline{0.9391} & \textbf{0.8758} & 0.9409 & 0.9226 & \underline{0.9880} & 0.9635 & \underline{0.9538} & \underline{0.8933} & \underline{0.9905} & \underline{0.9638} \\
& RDSI & \textbf{0.9722} & 0.9481 & 0.9028 & 0.8669 & 0.7800 & 0.7151 & \textbf{0.9558} & \underline{0.8636} & \textbf{0.9655} & \underline{0.9420} & \textbf{1.0000} & 0.9764 & \textbf{0.9615} & \textbf{0.8962} & \textbf{1.0000} & \textbf{0.9664} \\
& FGZEFSR & 0.8611 & 0.7966 & 0.8056 & 0.7404 & 0.7600 & 0.6520 & 0.9090 & 0.7214 & 0.9409 & 0.8886 & 0.9277 & 0.7227 & 0.7154 & 0.5555 & 0.8857 & 0.7471 \\
\cdashline{2-20}
& \textbf{S$^2$FS} & \textbf{0.9722} & \underline{0.9509} & \textbf{0.9306} & \textbf{0.9102} & \textbf{0.8400} & 0.7784 & 0.9218 & 0.8302 & \textbf{0.9655} & \textbf{0.9531} & \textbf{1.0000} & \underline{0.9831} & 0.9385 & 0.8678 & 0.9857 & 0.9519 \\
\hline
\end{tabular}
}
\label{tab:2}
\end{table*}

For the $k$-means clustering, S$^2$FS demonstrates strong overall performance, achieving the best Max NMI on GLIOMA, Lung, and warpPIE10P, together with three second-best Max results (ALL-AML-4, Isolet, and warpAR10P). Regarding average NMI, it ranks first on GLIOMA, Lung, and warpPIE10P, while securing second-best results on ALL-AML-4, Isolet, and warpAR10P, reflecting a consistent top-two presence across most datasets. SFSS also shows competitive performance, attaining the best Max NMI on ALL-AML-4, SRBCT, and warpAR10P, and leading in average NMI on four datasets, particularly excelling on ALL-AML-3 and SRBCT. RDSI achieves the best Max NMI on ALL-AML-3 and second-best on SRBCT, with corresponding second-best average outcomes on both datasets. N3Y performs notably well on Isolet, where it obtains the best Max and average NMI, and also secures the second-best average on warpPIE10P. The remaining algorithms generally lag behind, rarely achieving competitive performance.

\begin{table*}[!t]
\caption{Maximum and average clustering performance (NMI) with the top $1$, $2$,..., $150$ features across eight small-sized high-dimensional datasets.}
\centering
\resizebox{0.95\textwidth}{!}{
\setlength{\tabcolsep}{5pt}
\begin{tabular}{cc*{9}{cc}}
\hline
\multirow{2}{*}{\textbf{Cls.}} & 
\multirow{2}{*}{\textbf{SOTA}} & 
\multicolumn{2}{c}{\textbf{ALL-AML-3}} & 
\multicolumn{2}{c}{\textbf{ALL-AML-4}} & 
\multicolumn{2}{c}{\textbf{GLIOMA}} & 
\multicolumn{2}{c}{\textbf{Isolet}} & 
\multicolumn{2}{c}{\textbf{Lung}} & 
\multicolumn{2}{c}{\textbf{SRBCT}} & 
\multicolumn{2}{c}{\textbf{warpAR10P}} & 
\multicolumn{2}{c}{\textbf{warpPIE10P}} \\
& & \textit{Max} & \textit{Ave} & \textit{Max} & \textit{Ave} & \textit{Max} & \textit{Ave} & \textit{Max} & \textit{Ave} & \textit{Max} & \textit{Ave} & \textit{Max} & \textit{Ave} & \textit{Max} & \textit{Ave} & \textit{Max} & \textit{Ave} \\
\hline
\multirow{9}{*}{\rotatebox{90}{\textbf{$k$-means}}} & SFSS & \underline{0.8805} & \textbf{0.8245} & \textbf{0.7978} & \textbf{0.7379} & \underline{0.7398} & \underline{0.6620} & 0.7655 & 0.7231 & \underline{0.6888} & \underline{0.6436} & \textbf{1.0000} & \textbf{0.9088} & \textbf{0.6845} & \textbf{0.6667} & \underline{0.6160} & 0.5797 \\
& FDM & 0.3448 & 0.1014 & 0.3545 & 0.1531 & 0.5662 & 0.5286 & 0.5044 & 0.4317 & 0.5072 & 0.4433 & 0.3662 & 0.1583 & 0.4466 & 0.3431 & 0.5678 & 0.5127 \\
& FSNMER & 0.1142 & 0.0949 & 0.1539 & 0.1325 & 0.4296 & 0.3636 & 0.7060 & 0.5984 & 0.5824 & 0.4594 & 0.1532 & 0.1180 & 0.3926 & 0.3032 & 0.3113 & 0.2898 \\
& IPD & 0.3984 & 0.1514 & 0.3159 & 0.1707 & 0.6292 & 0.5183 & 0.5114 & 0.4610 & 0.5073 & 0.4508 & 0.6929 & 0.2415 & 0.4356 & 0.3282 & 0.5096 & 0.4348 \\
& MDP & 0.4960 & 0.1319 & 0.4569 & 0.1888 & 0.6199 & 0.5474 & 0.5079 & 0.4727 & 0.5047 & 0.4566 & 0.5030 & 0.2028 & 0.4275 & 0.3580 & 0.5477 & 0.5085 \\
& N3Y & 0.6313 & 0.4896 & 0.6369 & 0.4991 & 0.6340 & 0.5588 & \textbf{0.8286} & \textbf{0.7799} & 0.5578 & 0.4982 & 0.8708 & 0.8199 & 0.5951 & 0.5710 & 0.6133 & \underline{0.5842} \\
& RDSI & \textbf{0.8860} & \underline{0.7742} & 0.7521 & 0.6545 & 0.5651 & 0.4975 & 0.7870 & 0.6921 & 0.6845 & 0.6381 & \underline{0.9913} & \underline{0.8703} & 0.5978 & 0.5559 & 0.5246 & 0.4717 \\
& FGZEFSR & 0.3539 & 0.3096 & 0.3182 & 0.2703 & 0.4503 & 0.3656 & 0.7060 & 0.5611 & 0.4527 & 0.4211 & 0.2321 & 0.1830 & 0.2696 & 0.1758 & 0.2126 & 0.1954 \\
\cdashline{2-20}
& \textbf{S$^2$FS} & 0.7909 & 0.7143 & \underline{0.7685} & \underline{0.6860} & \textbf{0.7812} & \textbf{0.7240} & \underline{0.8052} & \underline{0.7442} & \textbf{0.7987} & \textbf{0.7365} & 0.8857 & 0.8483 & \underline{0.6515} & \underline{0.6246} & \textbf{0.6668} & \textbf{0.6310} \\
\hline
\end{tabular}
}
\label{tab:3}
\end{table*}

\begin{figure*}[!t]
\centering
\subfloat[{\fontsize{7}{7}\selectfont \texttt{ALL-AML-3(CART)}}]{\includegraphics[width=0.16\textwidth]{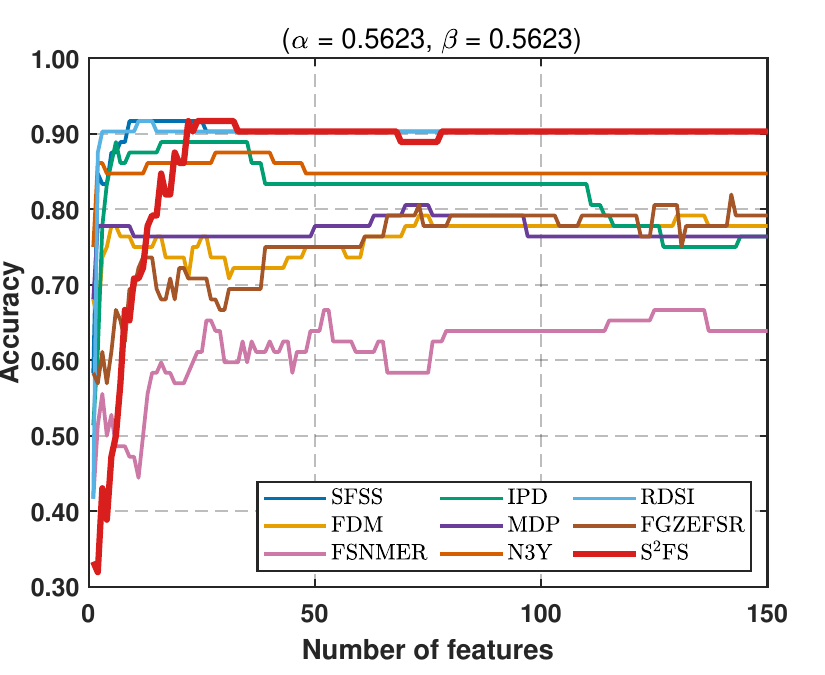}%
\label{fig:6a}}
\hfil
\subfloat[{\fontsize{7}{7}\selectfont \texttt{ALL-AML-4(CART)}}]{\includegraphics[width=0.16\textwidth]{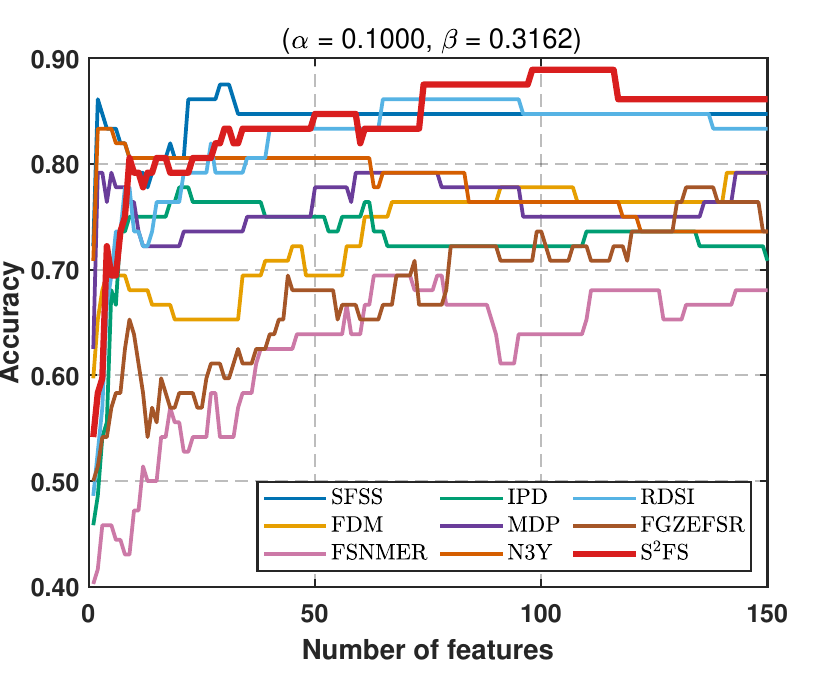}%
\label{fig:6b}}
\hfil
\subfloat[{\fontsize{7}{7}\selectfont \texttt{GLIOMA(CART)}}]{\includegraphics[width=0.16\textwidth]{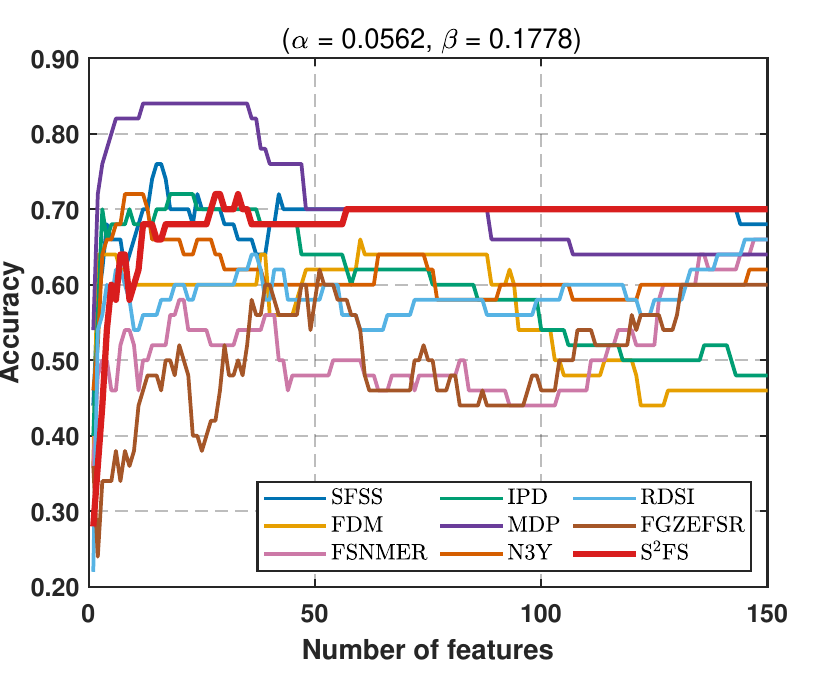}%
\label{fig:6c}}
\hfil
\subfloat[{\fontsize{7}{7}\selectfont \texttt{Isolet(CART)}}]{\includegraphics[width=0.16\textwidth]{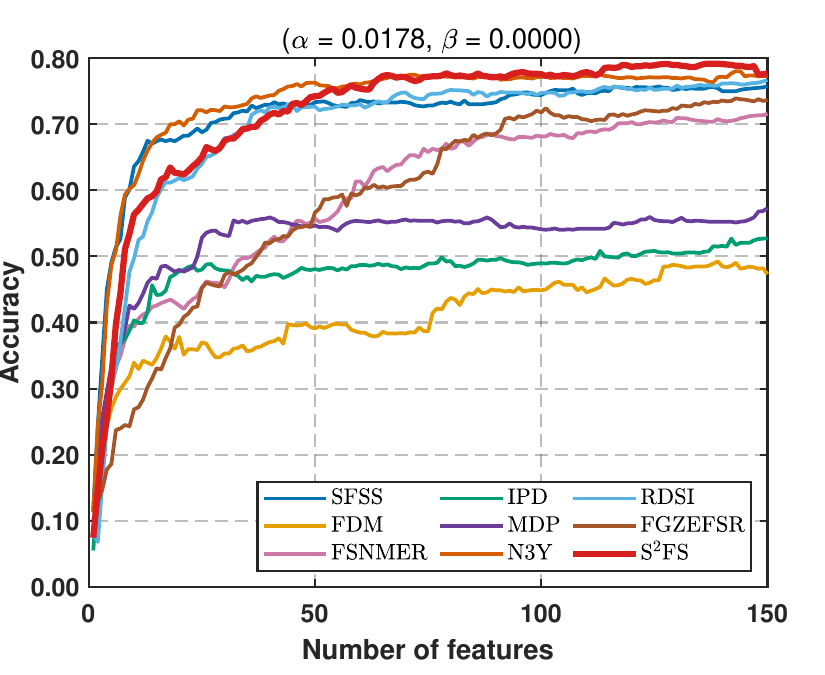}%
\label{fig:6d}}
\hfil
\subfloat[{\fontsize{7}{7}\selectfont \texttt{Lung(CART)}}]{\includegraphics[width=0.16\textwidth]{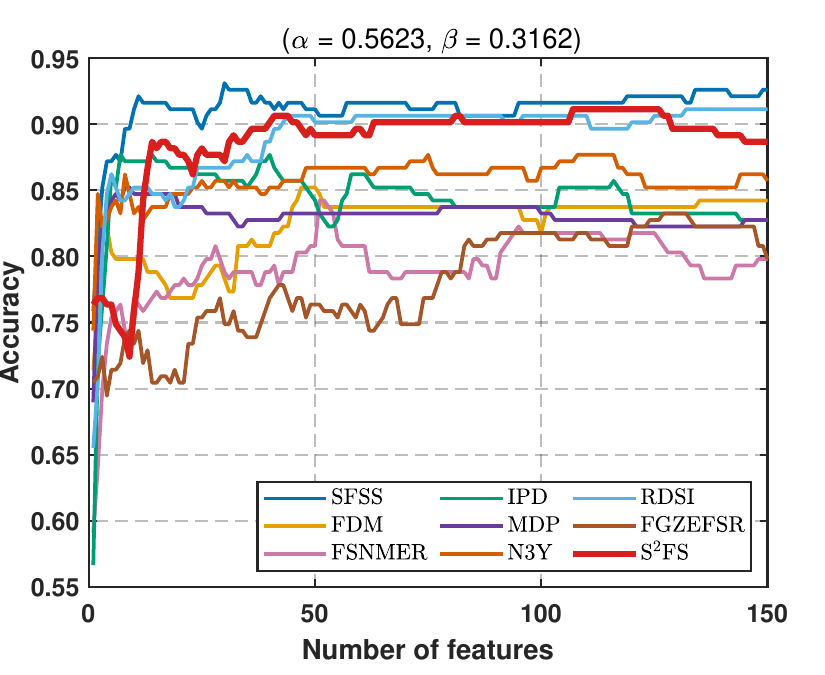}%
\label{fig:6e}}
\hfil
\subfloat[{\fontsize{7}{7}\selectfont \texttt{SRBCT(CART)}}]{\includegraphics[width=0.16\textwidth]{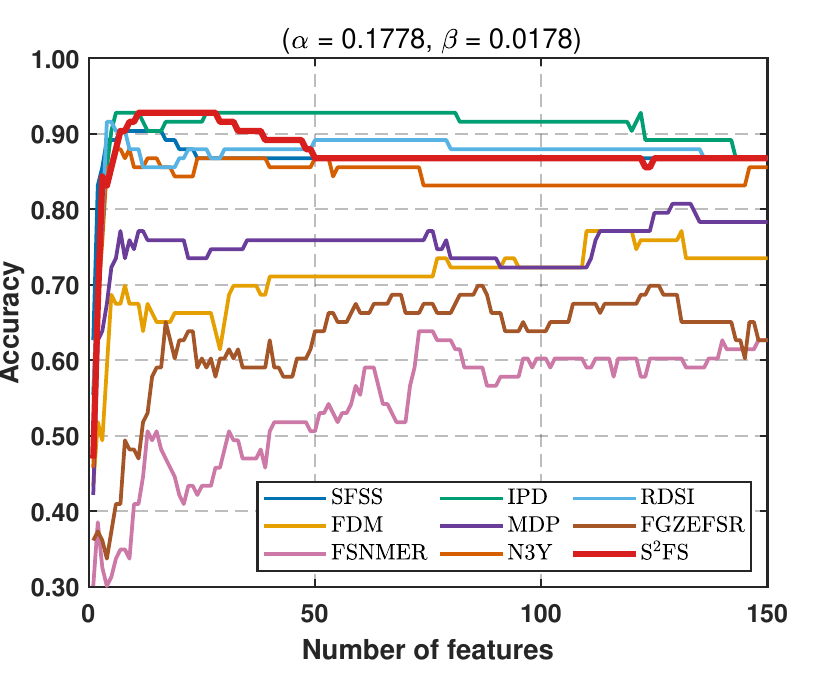}%
\label{fig:6f}}\\
\subfloat[{\fontsize{7}{7}\selectfont \texttt{warpAR10P(CART)}}]{\includegraphics[width=0.16\textwidth]{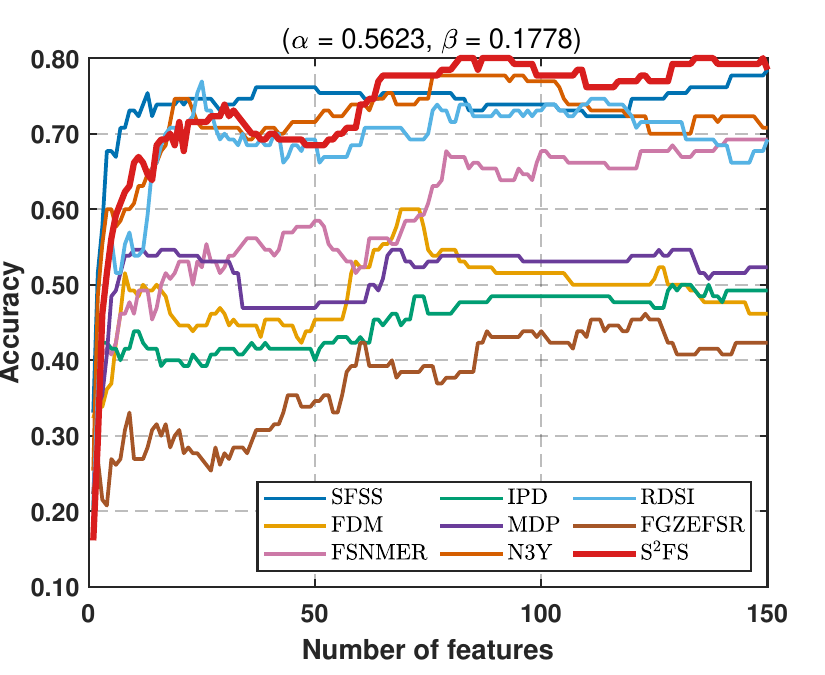}%
\label{fig:6g}}
\hfil
\subfloat[{\fontsize{7}{7}\selectfont \texttt{warpPIE10P(CART)}}]{\includegraphics[width=0.16\textwidth]{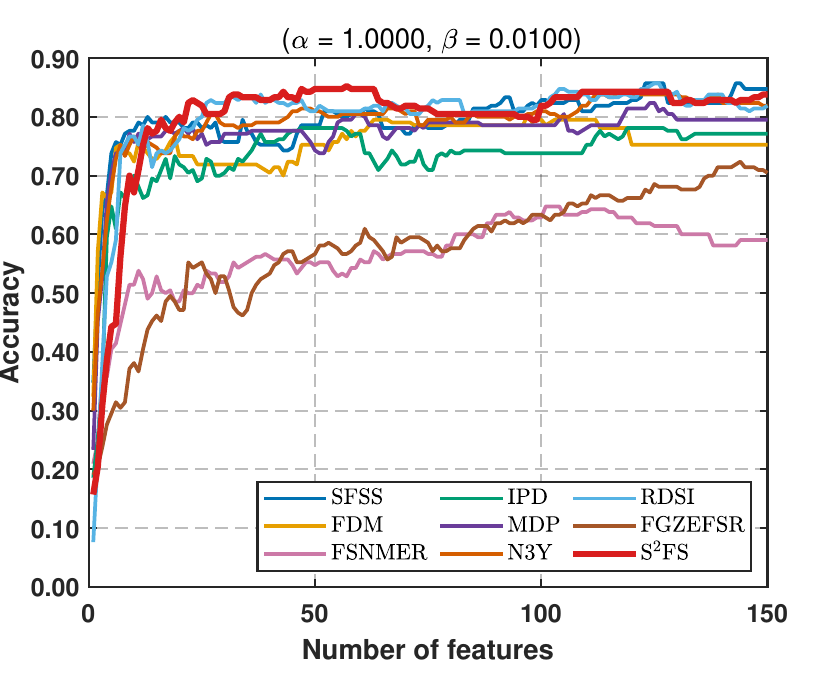}%
\label{fig:6h}}
\hfil
\subfloat[{\fontsize{7}{7}\selectfont \texttt{ALL-AML-3(KNN)}}]{\includegraphics[width=0.16\textwidth]{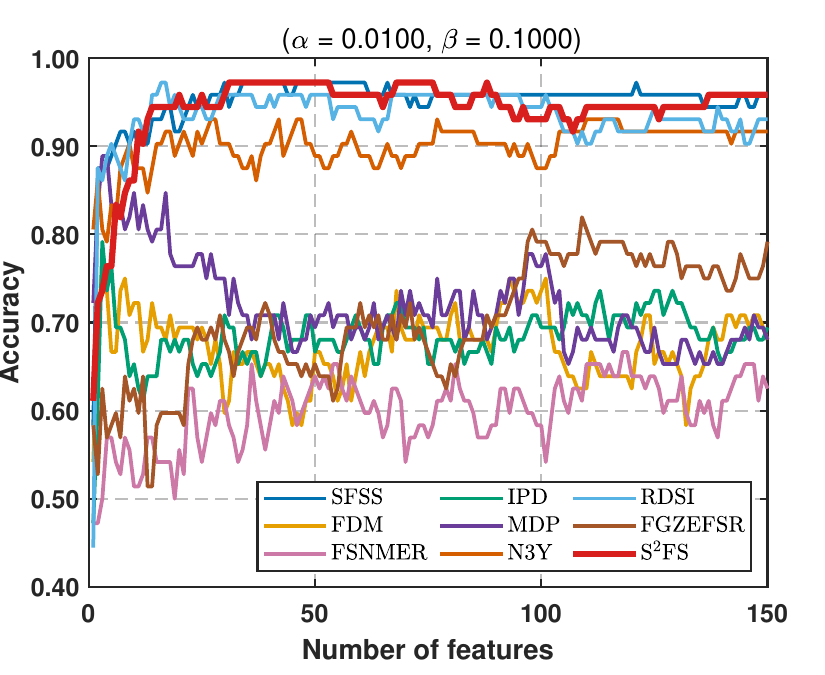}%
\label{fig:7a}}
\hfil
\subfloat[{\fontsize{7}{7}\selectfont \texttt{ALL-AML-4(KNN)}}]{\includegraphics[width=0.16\textwidth]{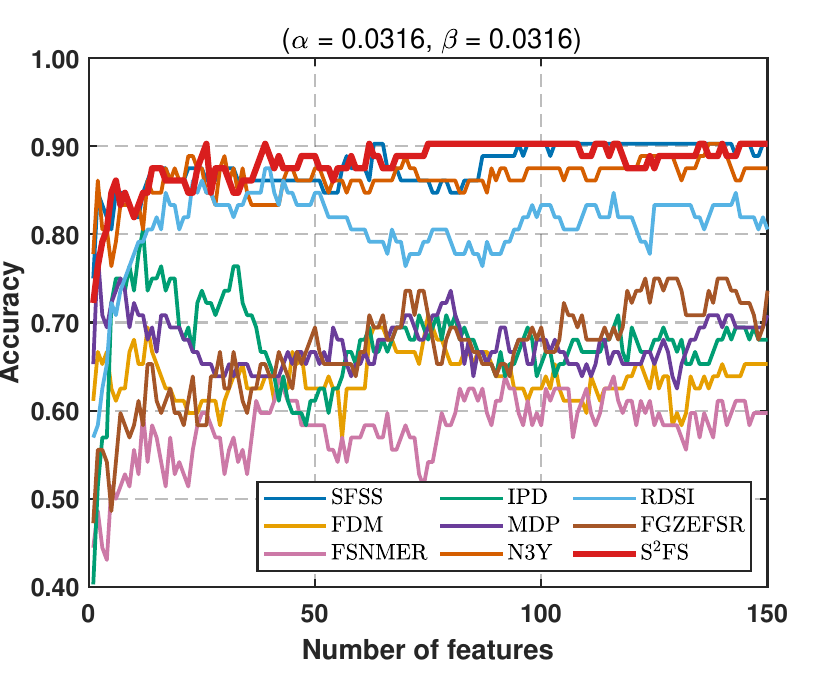}%
\label{fig:7b}}
\hfil
\subfloat[{\fontsize{7}{7}\selectfont \texttt{GLIOMA(KNN)}}]{\includegraphics[width=0.16\textwidth]{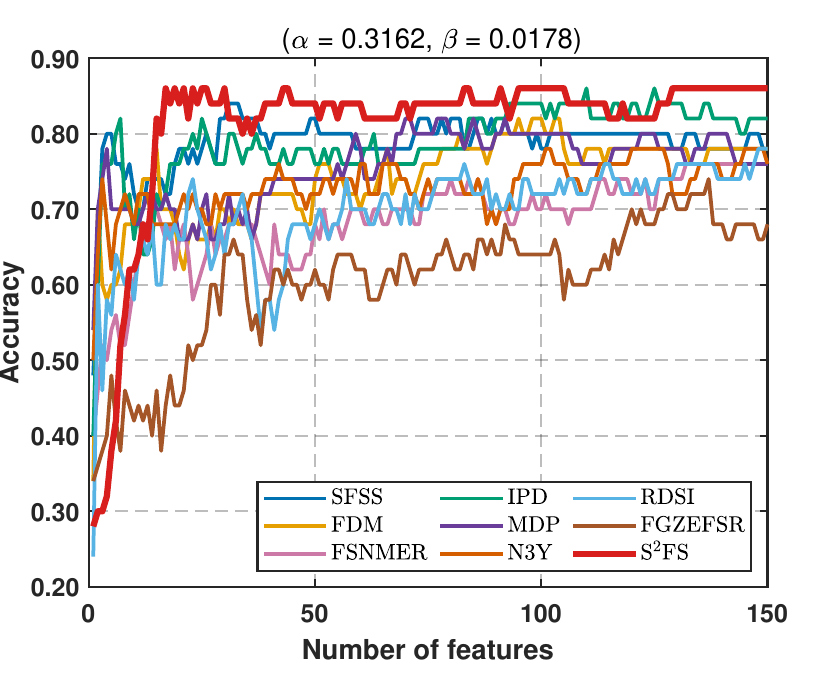}%
\label{fig:7c}}
\hfil
\subfloat[{\fontsize{7}{7}\selectfont \texttt{Isolet(KNN)}}]{\includegraphics[width=0.16\textwidth]{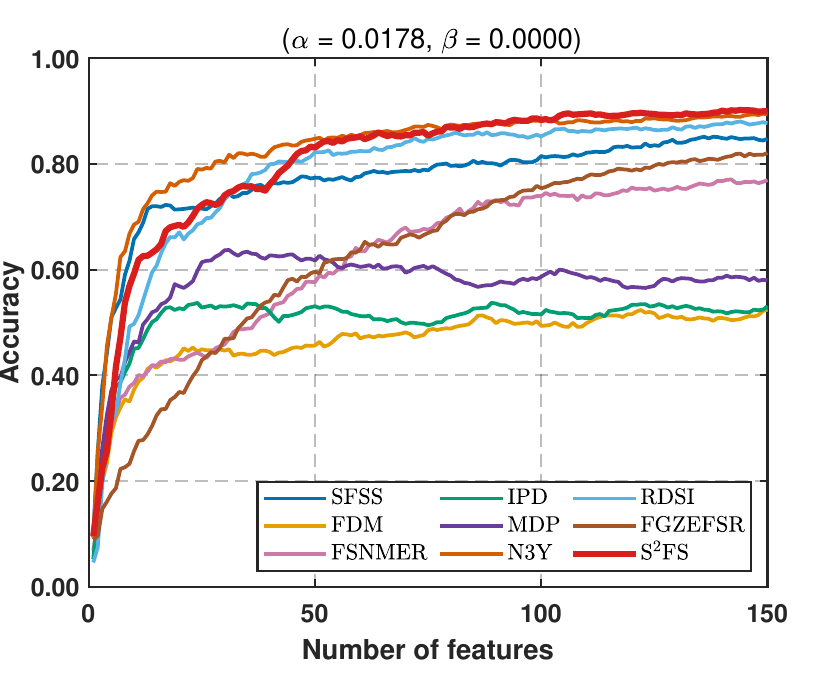}%
\label{fig:7d}}\\
\subfloat[{\fontsize{7}{7}\selectfont \texttt{Lung(KNN)}}]{\includegraphics[width=0.16\textwidth]{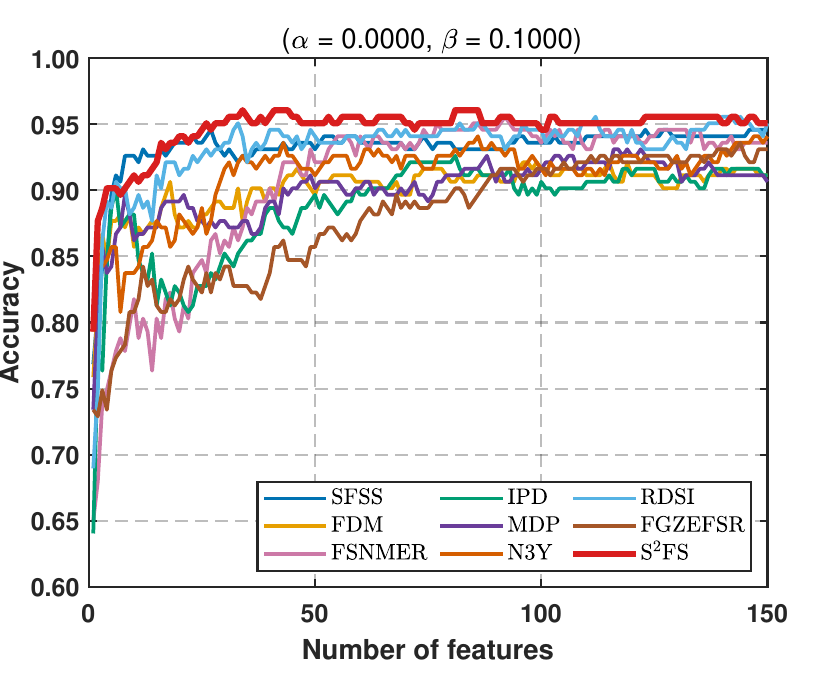}%
\label{fig:7e}}
\hfil
\subfloat[{\fontsize{7}{7}\selectfont \texttt{SRBCT(KNN)}}]{\includegraphics[width=0.16\textwidth]{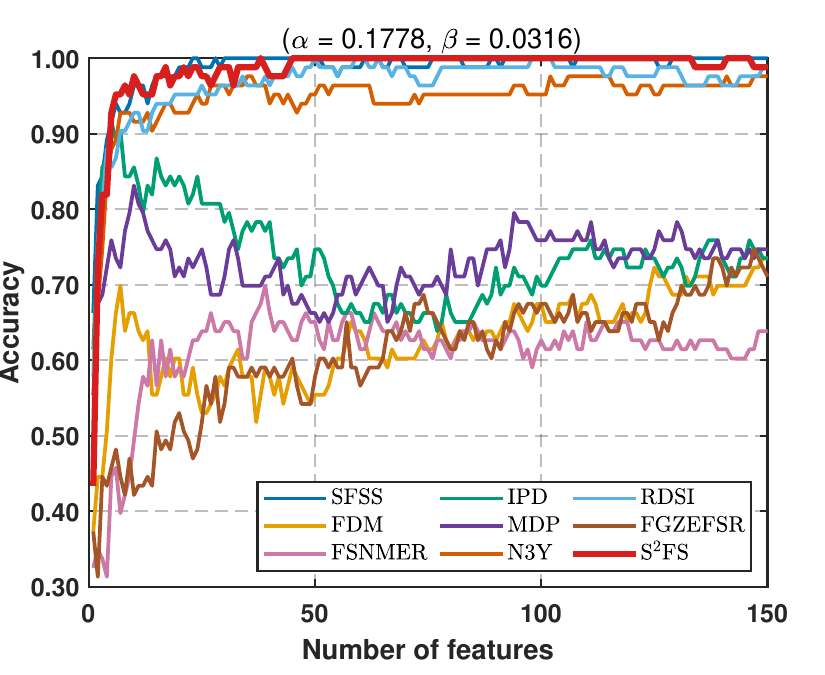}%
\label{fig:7f}}
\hfil
\subfloat[{\fontsize{7}{7}\selectfont \texttt{warpAR10P(KNN)}}]{\includegraphics[width=0.16\textwidth]{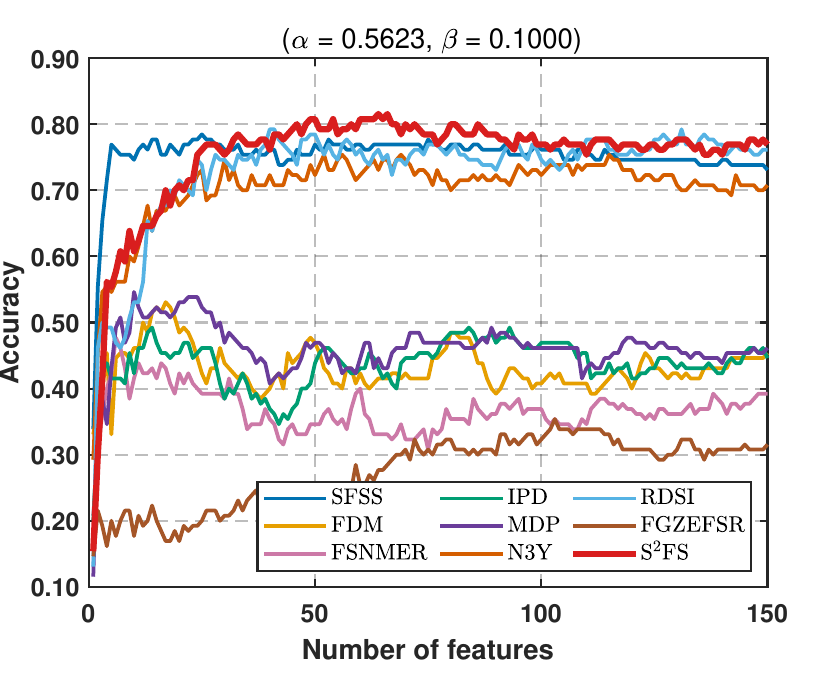}%
\label{fig:7g}}
\hfil
\subfloat[{\fontsize{7}{7}\selectfont \texttt{warpPIE10P(KNN)}}]{\includegraphics[width=0.16\textwidth]{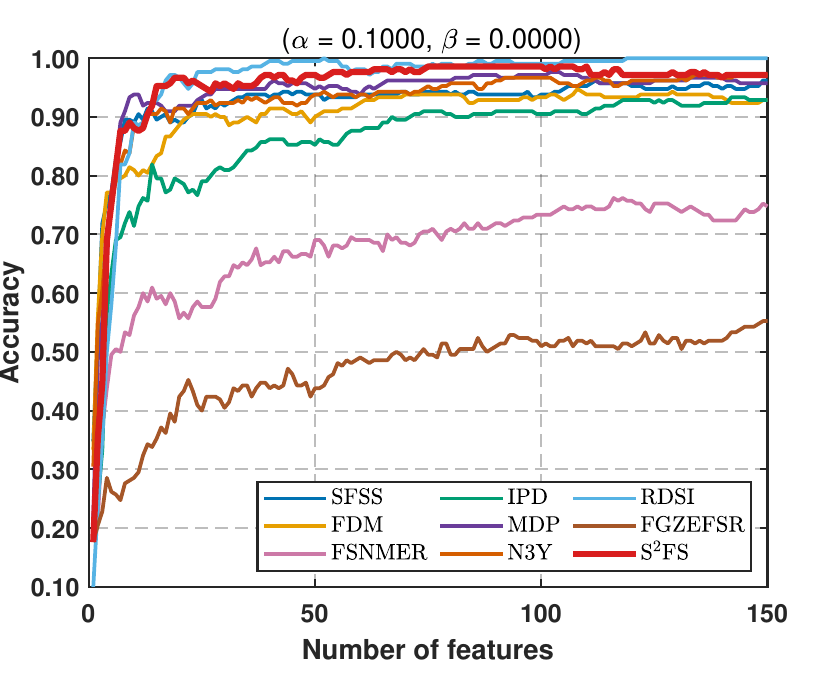}%
\label{fig:7h}}
\hfil
\subfloat[{\fontsize{7}{7}\selectfont \texttt{ALL-AML-3(SVM)}}]{\includegraphics[width=0.16\textwidth]{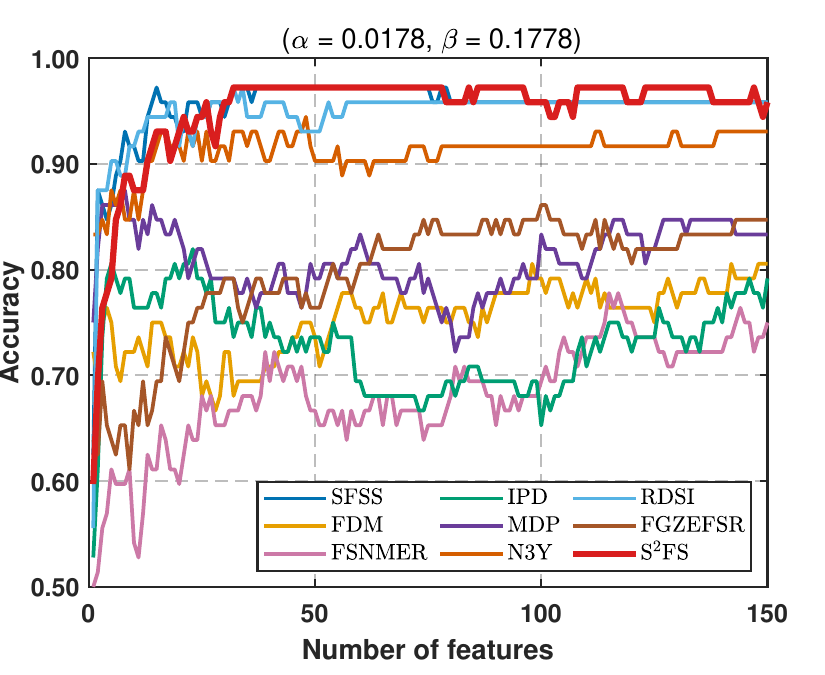}%
\label{fig:8a}}
\hfil
\subfloat[{\fontsize{7}{7}\selectfont \texttt{ALL-AML-4(SVM)}}]{\includegraphics[width=0.16\textwidth]{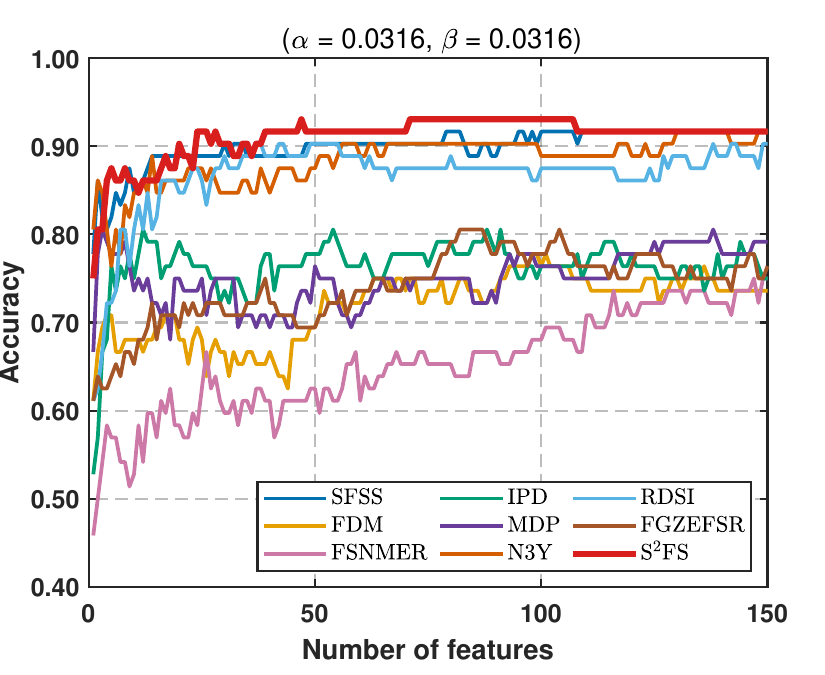}%
\label{fig:8b}}\\
\subfloat[{\fontsize{7}{7}\selectfont \texttt{GLIOMA(SVM)}}]{\includegraphics[width=0.16\textwidth]{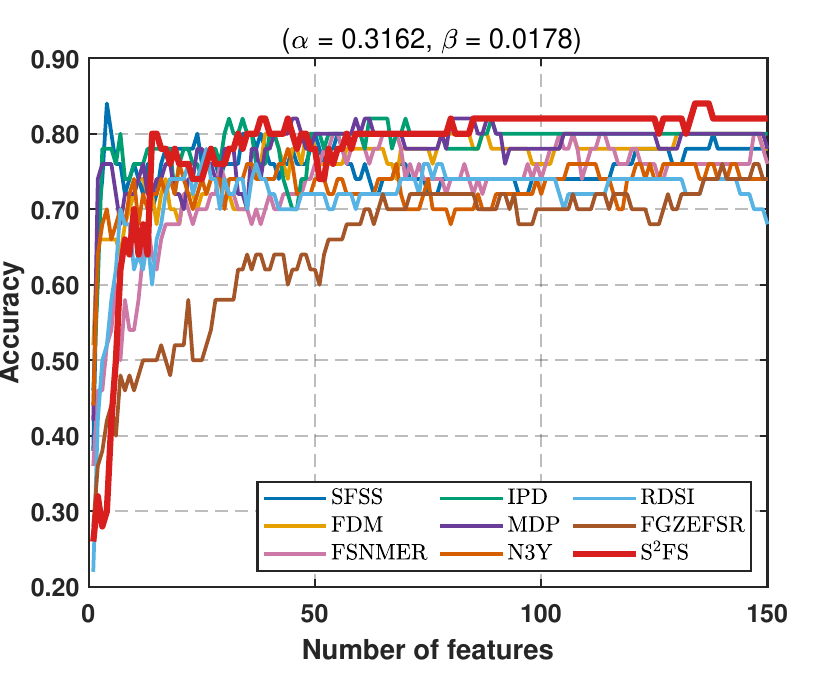}%
\label{fig:8c}}
\hfil
\subfloat[{\fontsize{7}{7}\selectfont \texttt{Isolet(SVM)}}]{\includegraphics[width=0.16\textwidth]{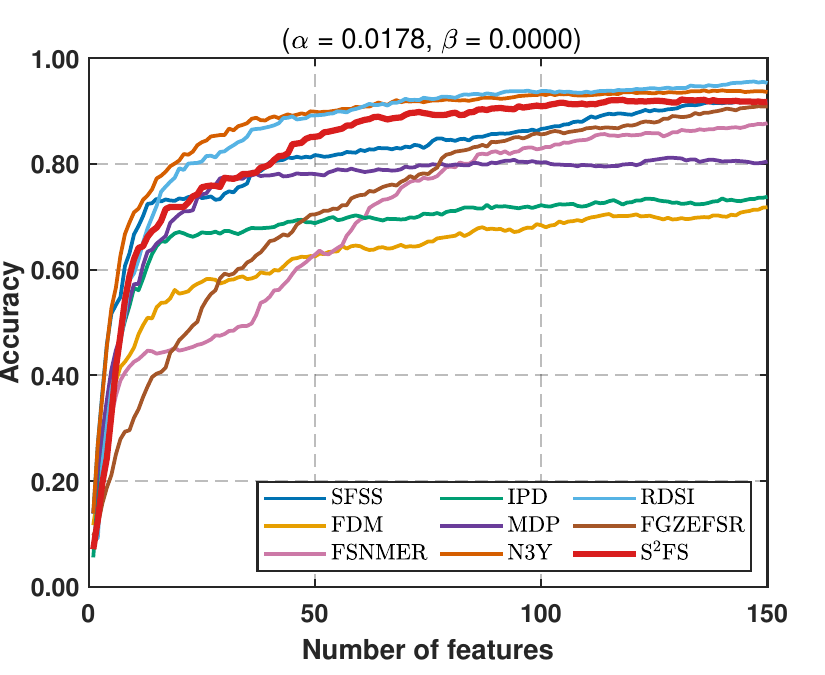}%
\label{fig:8d}}
\hfil
\subfloat[{\fontsize{7}{7}\selectfont \texttt{Lung(SVM)}}]{\includegraphics[width=0.16\textwidth]{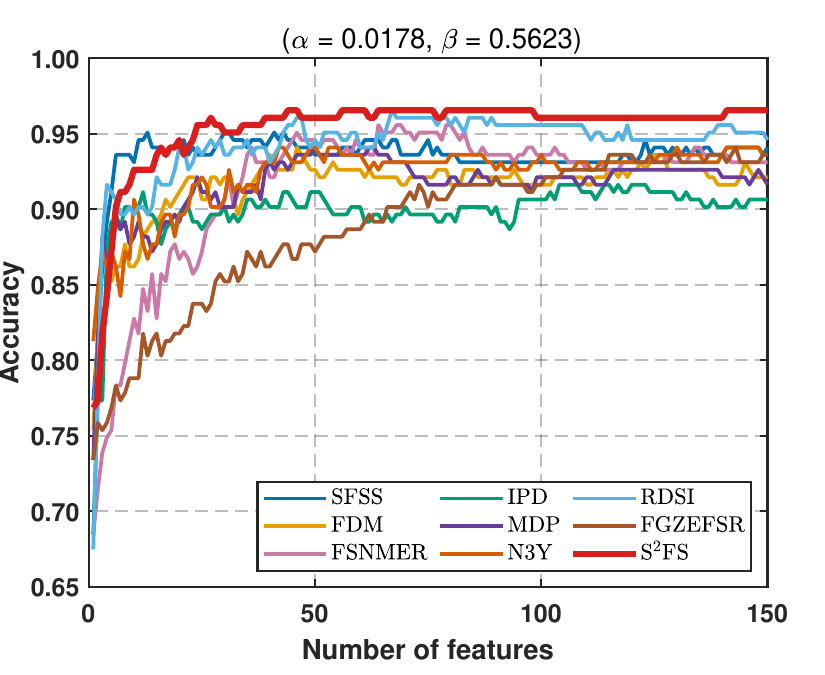}%
\label{fig:8e}}
\hfil
\subfloat[{\fontsize{7}{7}\selectfont \texttt{SRBCT(SVM)}}]{\includegraphics[width=0.16\textwidth]{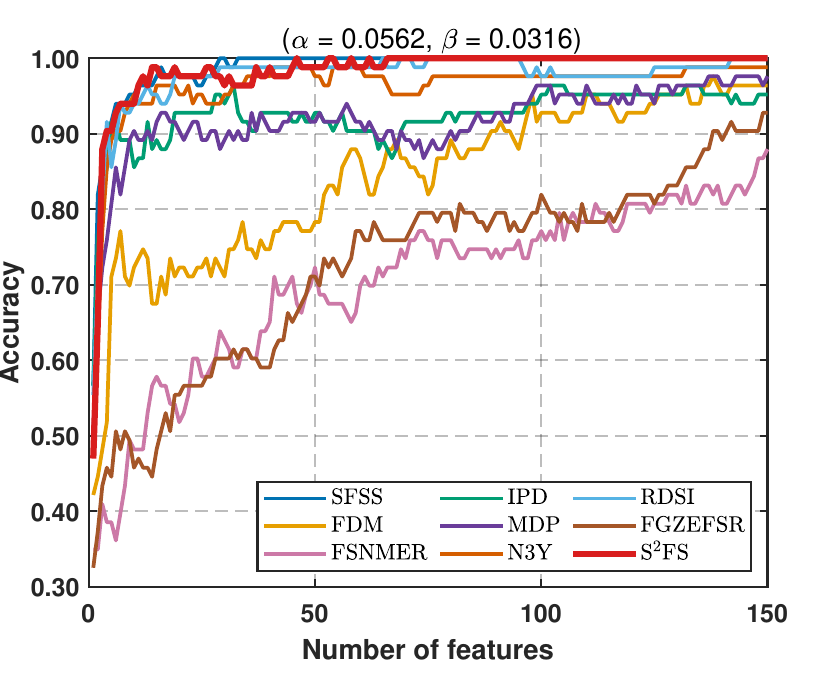}%
\label{fig:8f}}
\hfil
\subfloat[{\fontsize{7}{7}\selectfont \texttt{warpAR10P(SVM)}}]{\includegraphics[width=0.16\textwidth]{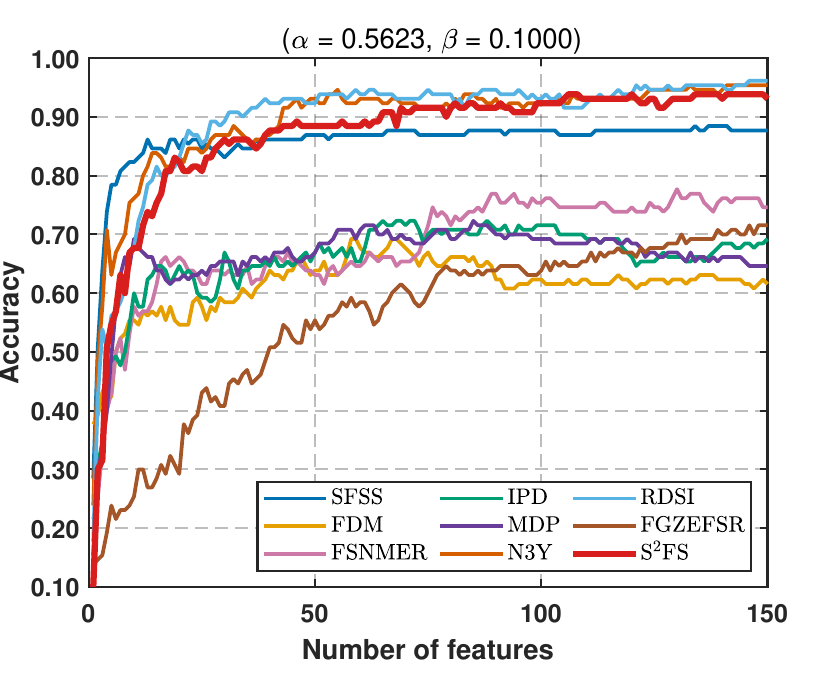}%
\label{fig:8g}}
\hfil
\subfloat[{\fontsize{7}{7}\selectfont \scriptsize \texttt{warpPIE10P(SVM)}}]{\includegraphics[width=0.16\textwidth]{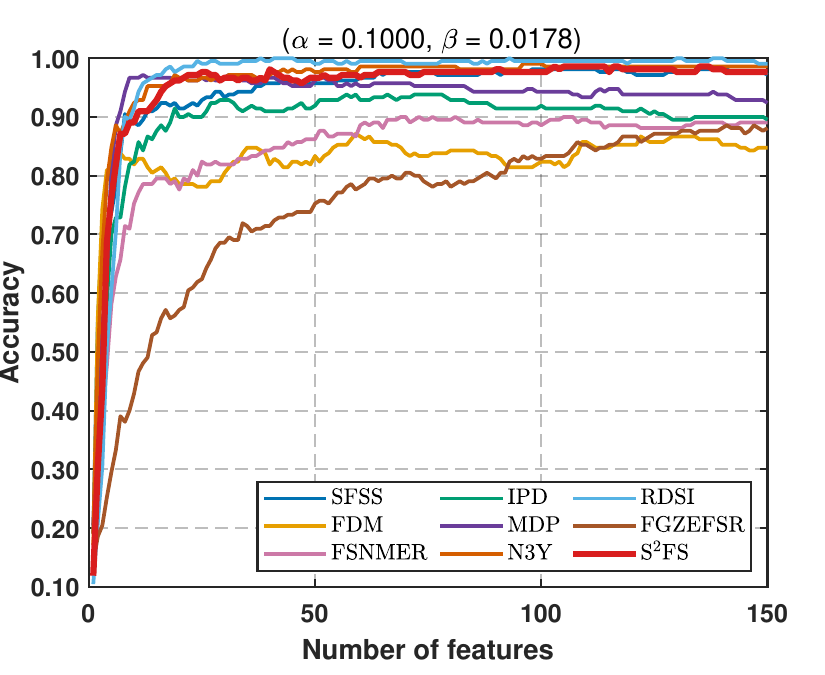}%
\label{fig:8h}}
\caption{Classification performance (Accuracy, CART/KNN/SVM) of nine algorithms across eight small-sized high-dimensional datasets.}
\label{fig:6}
\end{figure*}

\begin{figure*}[!t]
\centering
\subfloat[{\fontsize{7}{7}\selectfont \texttt{ALL-AML-3}}]{\includegraphics[width=0.16\textwidth]{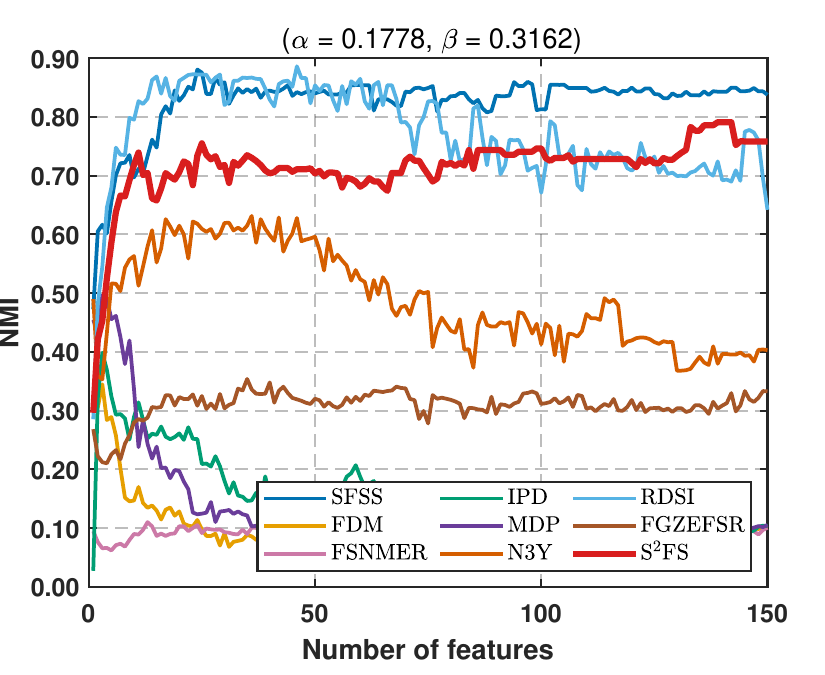}%
\label{fig:7a}}
\hspace{0.3mm}
\subfloat[{\fontsize{7}{7}\selectfont \texttt{ALL-AML-4}}]{\includegraphics[width=0.16\textwidth]{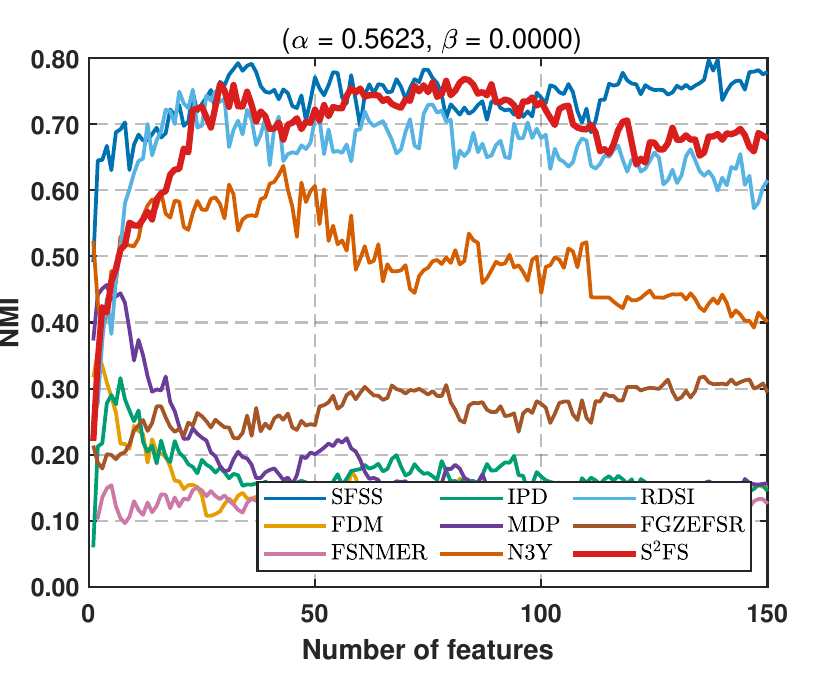}%
\label{fig:7b}}
\hspace{0.3mm}
\subfloat[{\fontsize{7}{7}\selectfont \texttt{GLIOMA}}]{\includegraphics[width=0.16\textwidth]{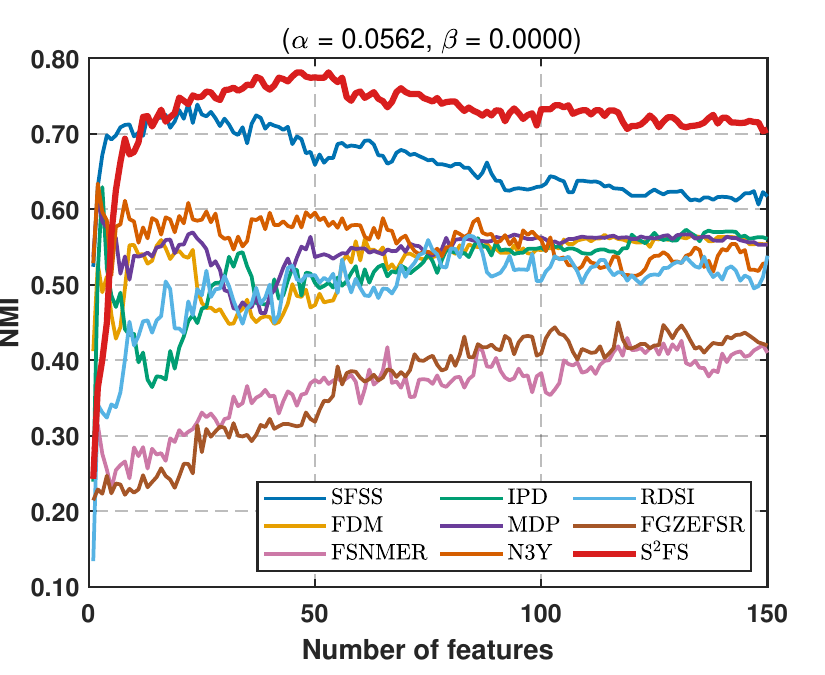}%
\label{fig:7c}}
\hspace{0.3mm}
\subfloat[{\fontsize{7}{7}\selectfont \texttt{Isolet}}]{\includegraphics[width=0.16\textwidth]{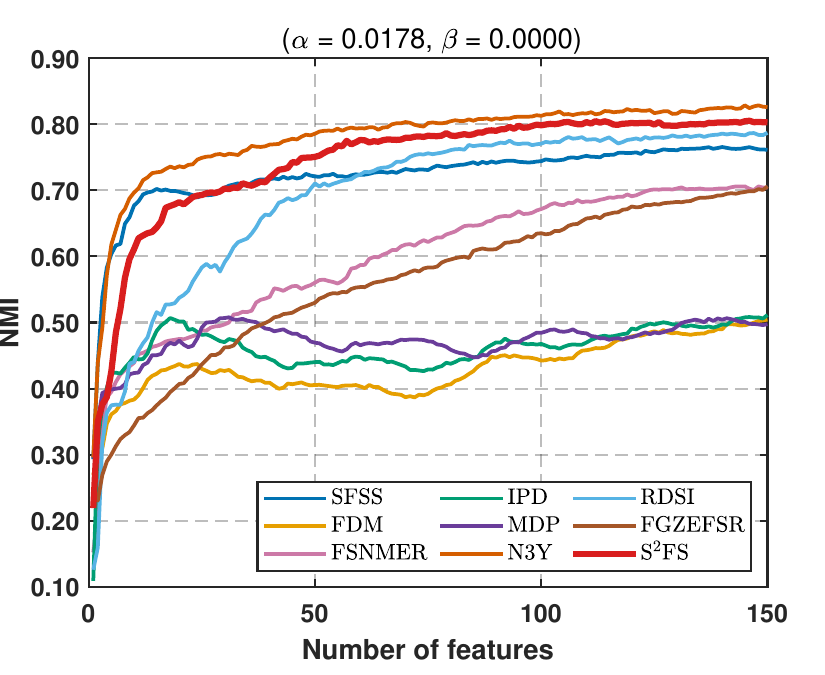}%
\label{fig:7d}}
\hspace{0.3mm}
\subfloat[{\fontsize{7}{7}\selectfont \texttt{Lung}}]{\includegraphics[width=0.16\textwidth]{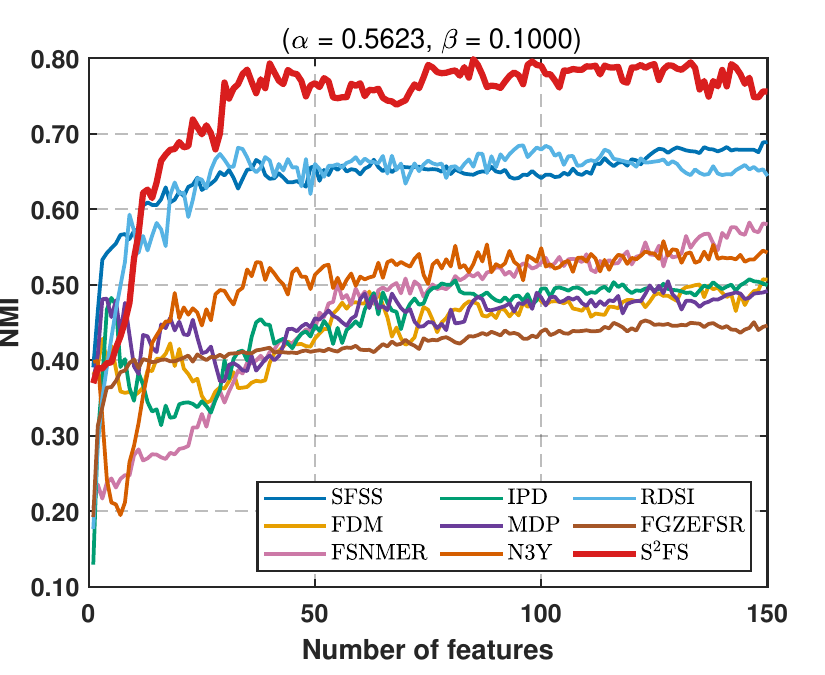}%
\label{fig:7e}}\\
\subfloat[{\fontsize{7}{7}\selectfont \texttt{SRBCT}}]{\includegraphics[width=0.16\textwidth]{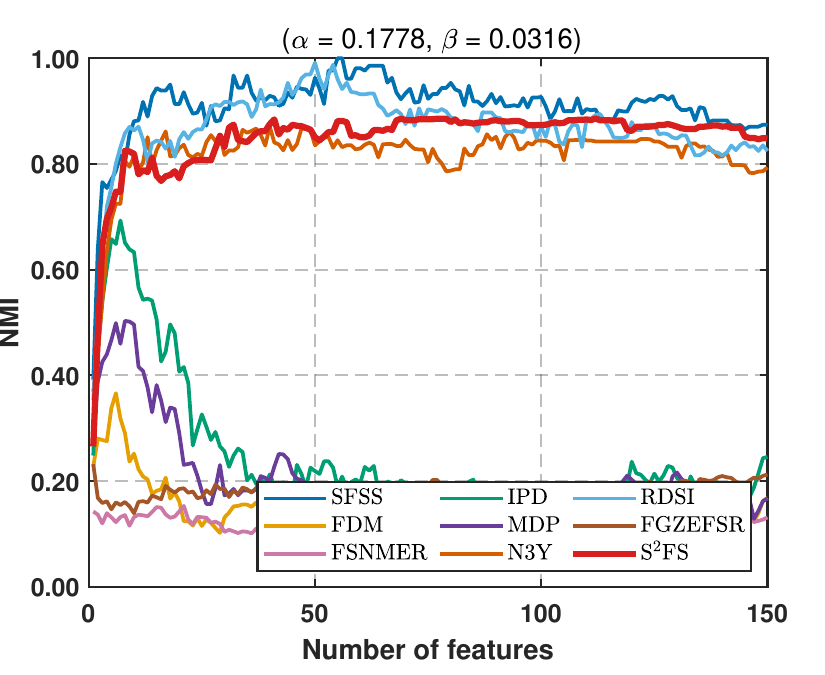}%
\label{fig:7f}}
\hspace{0.3mm}
\subfloat[{\fontsize{7}{7}\selectfont \texttt{warpAR10P}}]{\includegraphics[width=0.16\textwidth]{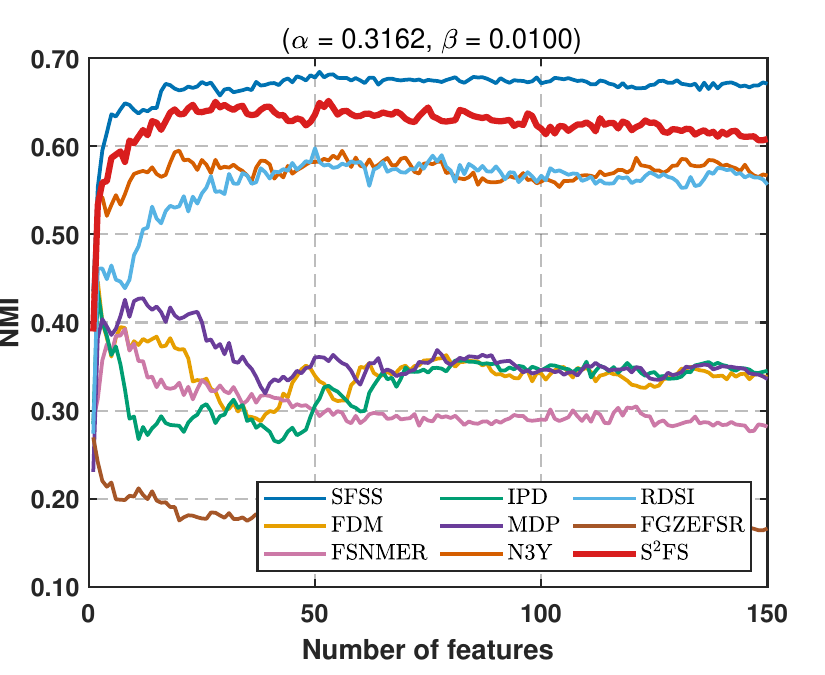}%
\label{fig:7g}}
\hspace{0.3mm}
\subfloat[{\fontsize{7}{7}\selectfont \texttt{warpPIE10P}}]{\includegraphics[width=0.16\textwidth]{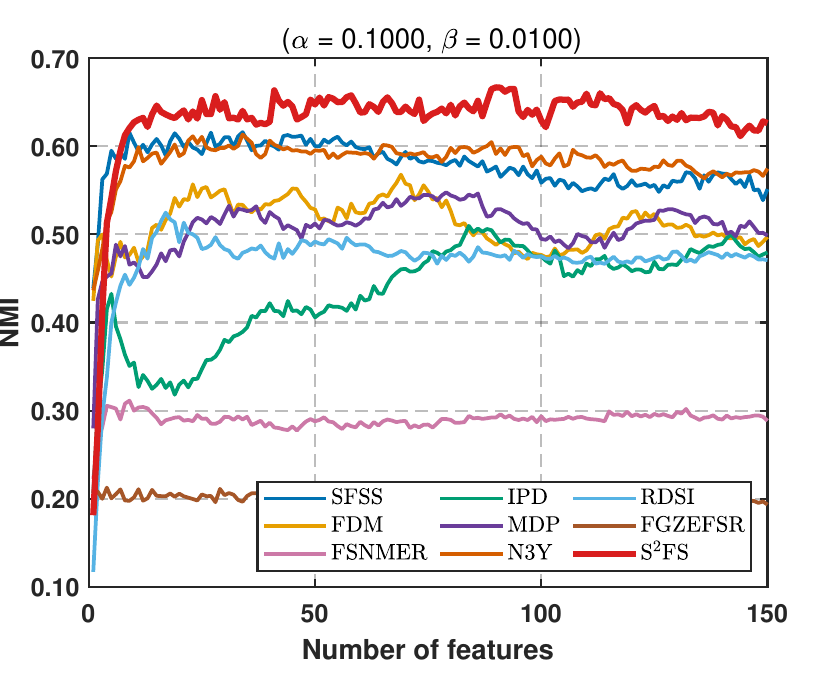}%
\label{fig:7h}}
\hspace{0.3mm}
\subfloat[{\fontsize{7}{7}\selectfont \texttt{ORL}}]{\includegraphics[width=0.16\textwidth]{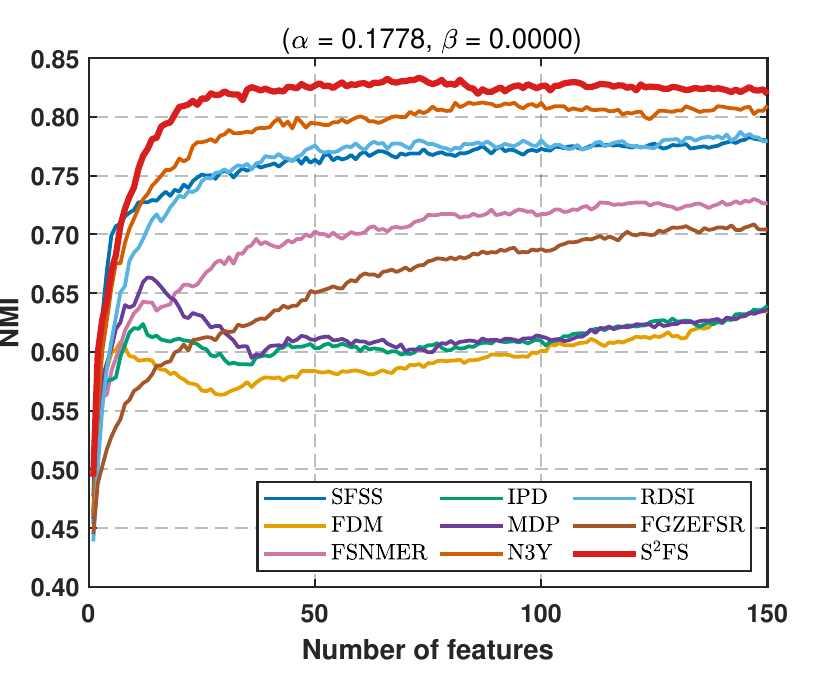}%
\label{fig:7i}}
\hspace{0.3mm}
\subfloat[{\fontsize{7}{7}\selectfont \texttt{Yale}}]{\includegraphics[width=0.16\textwidth]{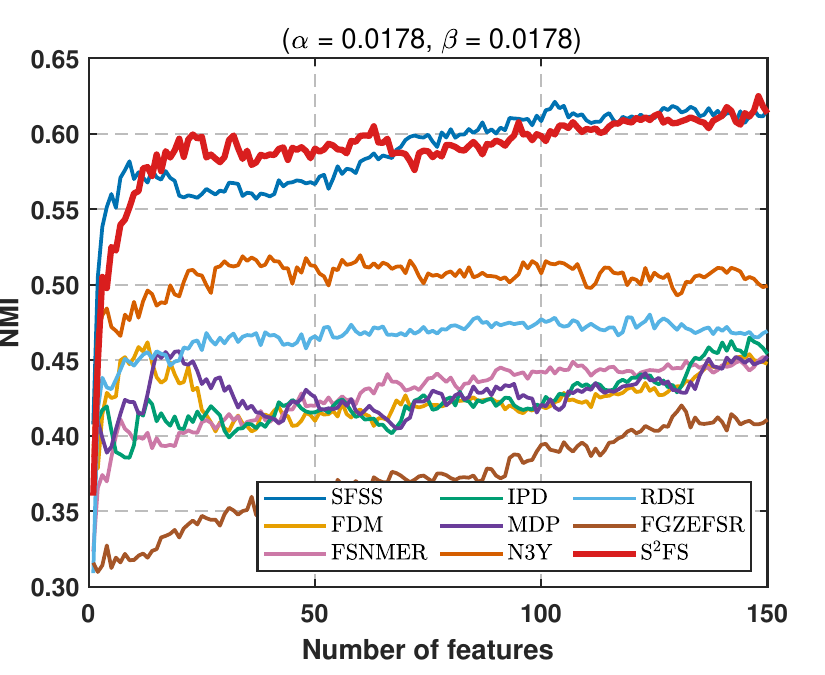}%
\label{fig:7j}}
\caption{Clustering performance (NMI, $k$-means) of nine algorithms across eight small-sized high-dimensional datasets and two face datasets.}
\label{fig:7}
\end{figure*}

Figs.~\ref{fig:6} and \ref{fig:7} illustrate the classification and clustering performance of nine feature selection algorithms across 150 incrementally top-ranked feature subsets. Specifically, Figs.~6(a)–6(h) report the results obtained with the CART classifier, Figs.~6(i)–6(p) correspond to the KNN classifier, and Figs.~6(q)–6(x) show the outcomes under the SVM classifier. For clustering, Figs.~7(a)–7(h) present the results based on $k$-means. Each curve depicts how performance changes as the number of selected features increases, highlighting not only the peak accuracy (or NMI for clustering) but also the consistency of each algorithm across varying sizes of feature subsets. The parameters $\alpha$ and $\beta$ adopted in S$^2$FS for each dataset are indicated above the corresponding figures.

As observed, the S$^2$FS curves (in red) generally lie above those of the competing algorithms across different datasets and learning models, reflecting a consistent advantage. Moreover, S$^2$FS attains relatively high performance with only a limited number of features, indicating that it effectively prioritizes and extracts the most discriminative features at an early stage of the selection process, thereby reducing redundancy and accelerating convergence toward optimal performance.

In summary, the results from Tabs.~\ref{tab:2} and \ref{tab:3} and Figs.~\ref{fig:6} and \ref{fig:7} demonstrate the superior performance of S$^2$FS on small-sized high-dimensional datasets, particularly its strengths in KNN classification and its strong generalization in $k$-means clustering. Its robustness and rapid convergence are clearly evident. These findings indicate that S$^2$FS not only achieves high peak accuracy (or NMI in clustering) but also delivers stable and efficient performance across varying feature subset sizes and learning paradigms.

\subsection{Results on Face Recognition Images}
We also evaluate the performance of the S$^2$FS algorithm on two widely used face recognition datasets: ORL and Yale. The ORL dataset contains $400$ images from $40$ subjects, while the Yale dataset includes $165$ grayscale images from $15$ subjects. Furthermore, we consider feature subsets consisting of the top-ranked features with sizes up to $150$. These subsets are then evaluated using CART, KNN, and SVM for classification, as well as $k$-means for clustering.

\begin{figure*}[!t]
\centering
\subfloat[{\fontsize{7}{7}\selectfont \texttt{ORL(CART)}}]{\includegraphics[width=0.16\textwidth]{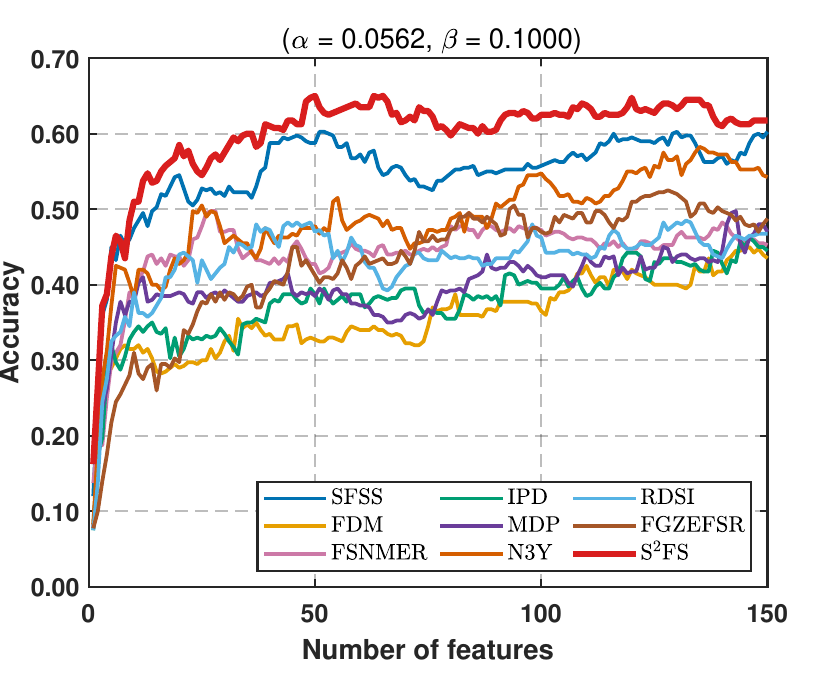}%
\label{fig:8a}}
\hfil
\subfloat[{\fontsize{7}{7}\selectfont \texttt{Yale(CART)}}]{\includegraphics[width=0.16\textwidth]{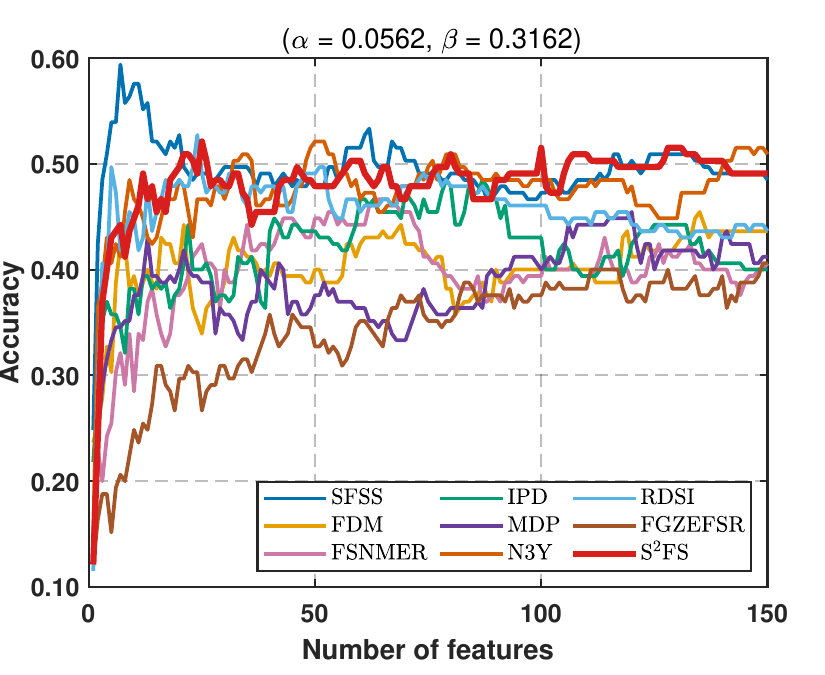}%
\label{fig:8b}}
\hfil
\subfloat[{\fontsize{7}{7}\selectfont \texttt{ORL(KNN)}}]{\includegraphics[width=0.16\textwidth]{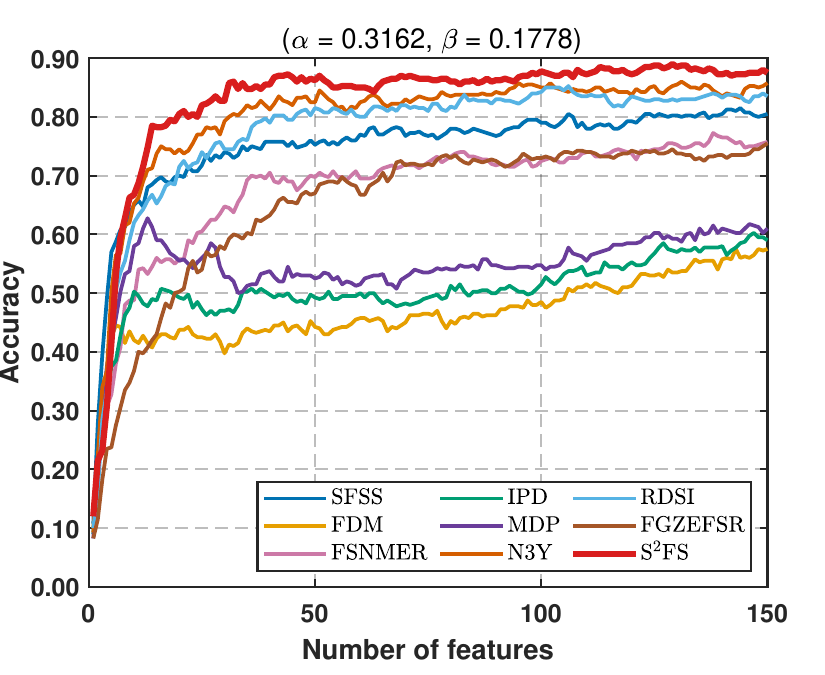}%
\label{fig:8c}}
\hfil
\subfloat[{\fontsize{7}{7}\selectfont \texttt{Yale(KNN)}}]{\includegraphics[width=0.16\textwidth]{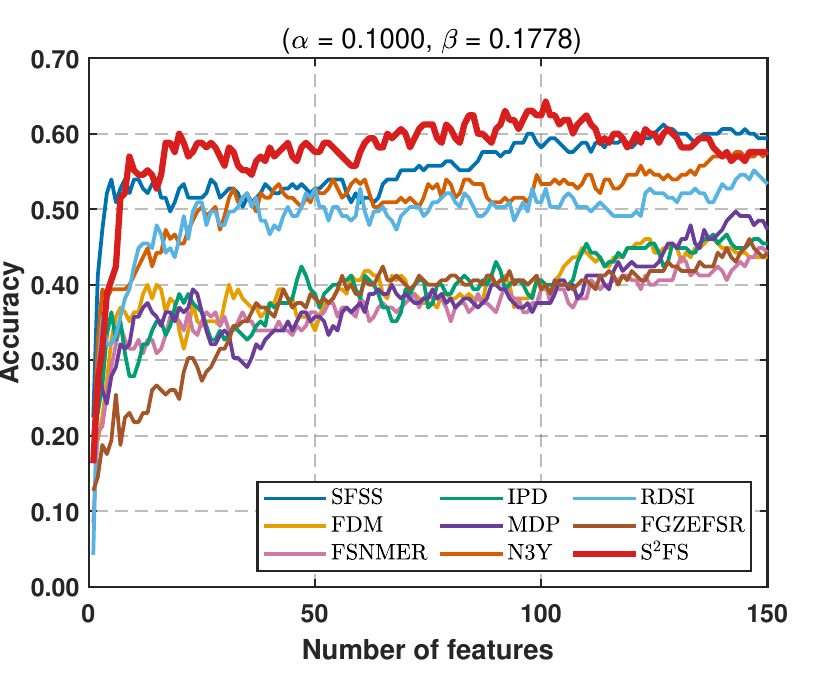}%
\label{fig:8d}}
\hfil
\subfloat[{\fontsize{7}{7}\selectfont \texttt{ORL(SVM)}}]{\includegraphics[width=0.16\textwidth]{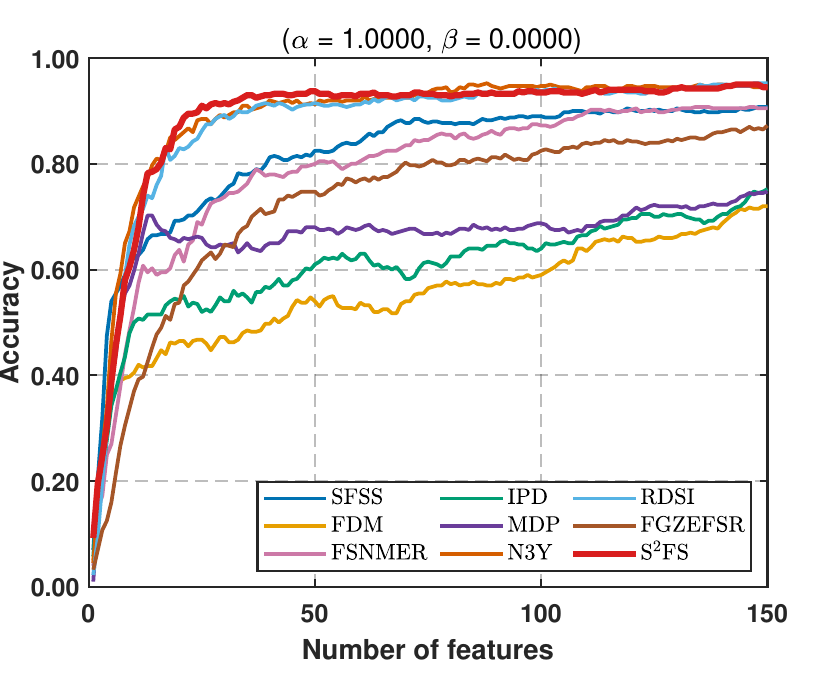}%
\label{fig:8e}}
\hfil
\subfloat[{\fontsize{7}{7}\selectfont \texttt{Yale(SVM)}}]{\includegraphics[width=0.16\textwidth]{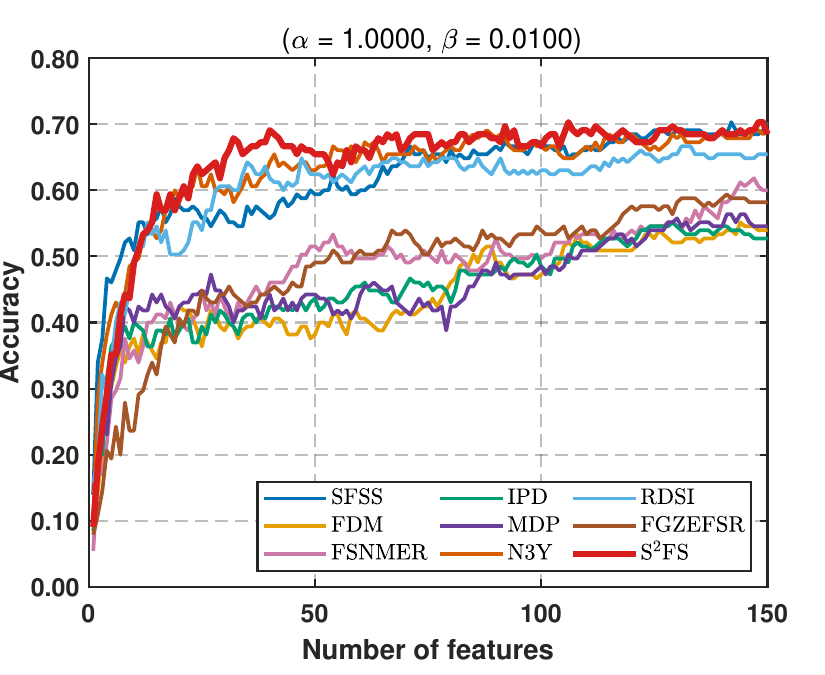}%
\label{fig:8f}}
\caption{Classification performance of nine algorithms on ORL and Yale datasets.}
\label{fig:8}
\end{figure*}

For the S$^2$FS algorithm and eight representative feature selection algorithms, the maximum and average performance achieved with the top $1$ to $150$ ranked features on the ORL and Yale datasets are reported in Tabs.~\ref{tab:4} and \ref{tab:5}, respectively. In addition, the fluctuation curves of classification accuracy and clustering NMI with varying numbers of selected features are illustrated in Fig.~\ref{fig:8} and Figs.~7(i)–7(j).

Tab.~\ref{tab:4} shows that S$^2$FS consistently outperforms the other algorithms in terms of maximum performance. On the ORL dataset, it achieves the best maximum CART and KNN accuracy, as well as the best maximum NMI for $k$-means clustering, while remaining highly competitive in SVM classification (0.9500 vs. the best 0.9525 obtained by N3Y and RDSI). On the Yale dataset, S$^2$FS again leads with the best maximum KNN accuracy (0.6424), maximum SVM accuracy (0.7030, tied with SFSS), and best maximum NMI for $k$-means clustering (0.6248), and is only slightly behind SFSS and RDSI under the CART classifier. These results highlight the strong advantage of the facial features selected by S$^2$FS in both classification and clustering tasks.

\begin{table}[!t]
\caption{Maximum performance with the top $1$, $2$,..., $150$ features on ORL and Yale datasets.}
\centering
\resizebox{0.46\textwidth}{!}{
\setlength{\tabcolsep}{5pt}
\begin{tabular}{c|cc|cc|cc|cc}
\hline
\multirow{2}{*}{\textbf{SOTA}} & \multicolumn{2}{c|}{\textbf{Accuracy (CART)}} & \multicolumn{2}{c|}{\textbf{Accuracy (KNN)}} & \multicolumn{2}{c|}{\textbf{Accuracy (SVM)}} & \multicolumn{2}{c}{\textbf{NMI ($k$-means)}} \\
& \textbf{ORL} & \textbf{Yale} & \textbf{ORL} & \textbf{Yale} & \textbf{ORL} & \textbf{Yale} & \textbf{ORL} & \textbf{Yale}\\
\hline
SFSS & \underline{0.6025} & \textbf{0.5939} & 0.8150 & \underline{0.6121} & 0.9075 & \textbf{0.7030} & 0.7829 & \underline{0.6211} \\
FDM & 0.4500 & 0.4545 & 0.5750 & 0.4606 & 0.7200 & 0.5515 & 0.6354 & 0.4619 \\
FSNMER & 0.4975 & 0.4667 & 0.7725 & 0.4485 & 0.9075 & 0.6182 & 0.7302 & 0.4521 \\
IPD & 0.4625 & 0.4848 & 0.6025 & 0.4667 & 0.7525 & 0.5515 & 0.6393 & 0.4650 \\
MDP & 0.4975 & 0.4545 & 0.6275 & 0.4970 & 0.7475 & 0.5636 & 0.6634 & 0.4560 \\
N3Y & 0.5825 & 0.5212 & \underline{0.8600} & 0.5758 & \textbf{0.9525} & \underline{0.6909} & \underline{0.8123} & 0.5195 \\
RDSI & 0.4850 & \underline{0.5273} & 0.8525 & 0.5515 & \textbf{0.9525} & 0.6667 & 0.7873 & 0.4803 \\
FGZEFSR & 0.5250 & 0.4061 & 0.7550 & 0.4606 & 0.8725 & 0.5939 & 0.7086 & 0.4201 \\
\cdashline{1-9}
\textbf{S$^2$FS} & \textbf{0.6500} & 0.5212 & \textbf{0.8900} & \textbf{0.6424} & \underline{0.9500} & \textbf{0.7030} & \textbf{0.8334} & \textbf{0.6248}\\
\hline
\end{tabular}
}
\label{tab:4}
\end{table}

\begin{table}[!t]
\caption{Average performance with the top $1$, $2$,..., $150$ features on ORL and Yale datasets.}
\centering
\resizebox{0.46\textwidth}{!}{
\setlength{\tabcolsep}{5pt}
\begin{tabular}{c|cc|cc|cc|cc}
\hline
\multirow{2}{*}{\textbf{SOTA}} & \multicolumn{2}{c|}{\textbf{Accuracy (CART)}} & \multicolumn{2}{c|}{\textbf{Accuracy (KNN)}} & \multicolumn{2}{c|}{\textbf{Accuracy (SVM)}} & \multicolumn{2}{c}{\textbf{NMI ($k$-means)}} \\
& \textbf{ORL} & \textbf{Yale} & \textbf{ORL} & \textbf{Yale} & \textbf{ORL} & \textbf{Yale} & \textbf{ORL} & \textbf{Yale}\\
\hline
SFSS & \underline{0.5489} & \textbf{0.4950}  & 0.7500  & \underline{0.5538} & 0.8188 & 0.6200 & 0.7589 & \underline{0.5888} \\
FDM & 0.3574 & 0.4023 & 0.4673 & 0.3956 & 0.5541 & 0.4442 & 0.5942 & 0.4252 \\
FSNMER & 0.4425 & 0.3980 & 0.6789 & 0.3703 & 0.7894 & 0.4841 & 0.6964 & 0.4274 \\
IPD & 0.3786 & 0.4185 & 0.5060 & 0.3936 & 0.6073 & 0.4568 & 0.6081 & 0.4231 \\
MDP & 0.3982 & 0.3874 & 0.5469 & 0.3828 & 0.6622 & 0.4606 & 0.6158 & 0.4278 \\
N3Y & 0.4880 & 0.4746 & \underline{0.7995} & 0.5123 & \textbf{0.8901} & \underline{0.6309} & \underline{0.7860} & 0.5045 \\
RDSI & 0.4329 & 0.4573 & 0.7770 & 0.4899 & 0.8775 & 0.6049 & 0.7569 & 0.4655 \\
FGZEFSR & 0.4282 & 0.3424 & 0.6509 & 0.3720 & 0.7264 & 0.4857 & 0.6572 & 0.3718 \\
\cdashline{1-9}
\textbf{S$^2$FS} & \textbf{0.5991} & \underline{0.4813} & \textbf{0.8292} & \textbf{0.5771} & \underline{0.8883} & \textbf{0.6421} & \textbf{0.8113} & \textbf{0.5902} \\
\hline
\end{tabular}
}
\label{tab:5}
\end{table}

As shown in Tab.~\ref{tab:5}, the trends in average performance are consistent with those observed for maximum performance. On the ORL dataset, S$^2$FS achieves the best results under CART, KNN, and $k$-means clustering, and ranks second in SVM classification (0.8883), just behind N3Y (0.8901). On the Yale dataset, it achieves the best average performance in KNN accuracy (0.5771), SVM accuracy (0.6421), and NMI for $k$-means clustering (0.5902), with CART remaining close to the best SFSS. Consistently, the fluctuation curves in Fig.~\ref{fig:8} and Figs.~7(i)–7(j) show that S$^2$FS remains above competing algorithms with varying feature subset sizes, highlighting its stable superiority.

Overall, S$^2$FS exhibits both robustness and generalizability, excelling in peak performance on two face recognition tasks.

\subsection{Statistical Tests}
To compare predictive performance in a statistically well-founded way, we adopt the \emph{Friedman test} \cite{FriedmanM1940}. Suppose $s$ algorithms are evaluated across $N$ datasets. Let $r_i^j$ denote the rank of the $j$-th algorithm on the $i$-th dataset, and define its average rank as: $R_j = \frac{1}{N}\sum_{i=1}^{N}r_i^j$.

Under the null hypothesis (no performance differences among algorithms), the Friedman statistic follows an approximate Fisher ($F$) distribution with $(s-1)$ and $(s-1)(N-1)$ degrees of freedom. Specifically, the Friedman statistic $F_F$ is computed as:

\begin{equation}
F_F = \frac{(N-1)\chi_F^2}{N(s-1)-\chi_F^2},
\label{eq:25}
\end{equation}
where
\begin{equation}
\chi_F^2 = \frac{12N}{s(s+1)}(\sum_{j=1}^{s}R_j^2-\frac{s(s+1)^2}{4}).
\label{eq:26}
\end{equation}

Tab.~\ref{tab:6} summarizes the Friedman statistics $F_F$ ($s = 9$, $N = 10$) along with the corresponding critical value (CV) for eight evaluation settings, i.e., Maximum (Max) and average (Ave) performance with the top $1$, $2$,..., $150$ features evaluated by CART, KNN, SVM, and $k$-means. At the $0.05$ significance level, all $F_F$ values exceed the CV, leading to rejection of the null hypothesis that all algorithms perform equally. To identify which algorithms differ significantly, we further conduct the post-hoc \emph{Nemenyi test}. Two algorithms are considered significantly different if the difference in their average ranks exceeds the critical difference (CD):
\begin{equation}
CD = q_\alpha \sqrt{\frac{s(s+1)}{6N}}
\label{eq:30}
\end{equation}
where $q_\alpha$ is derived from the Studentized range statistic divided by $\sqrt{2}$. For $\alpha = 0.05$, $q_\alpha = 3.1020$, yielding $CD = 3.7992$.

\begin{table}[!t]
\caption{Friedman statistic and critical value.}
\centering
\setlength{\tabcolsep}{10pt}
\begin{tabular}{ccccc}
\hline
Evaluation settings & $F_F$ & CV ($\alpha = 0.05$) \\
\hline
\multicolumn{1}{c}{Max (CART)} &  \multicolumn{1}{c}{$15.3958$} & \multirow{8}{*}{$2.0698$}\\
Max (KNN)       &        $18.4042$         &                   \\
Max (SVM)       &        $15.2697$         &                   \\
Max ($k$-means)       &        $24.8028$         &                   \\
Ave (CART)      &        $21.0668$         &                   \\
Ave (KNN)       &        $22.0702$         &                   \\
Ave (SVM)       &        $15.4123$         &                   \\
Ave ($k$-means)       &        $27.9357$         &                   \\
\hline
\end{tabular}
\label{tab:6}
\end{table}

\begin{figure}[!t]
\centering
\subfloat[\texttt{Max accuracy (CART)}]{\includegraphics[width=0.24\textwidth]{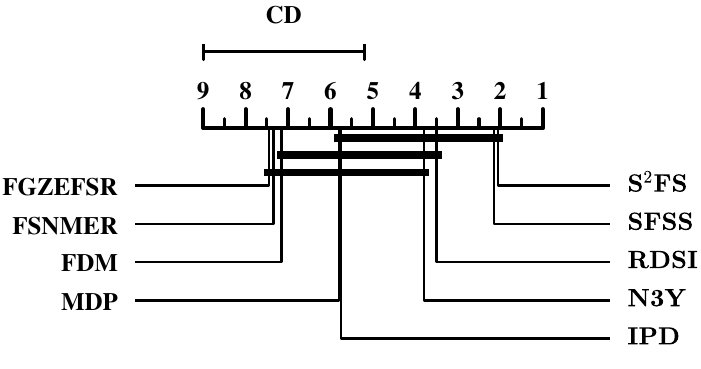}%
\label{fig:9a}}
\hspace{0.3mm}
\subfloat[\texttt{Max accuracy (KNN)}]{\includegraphics[width=0.24\textwidth]{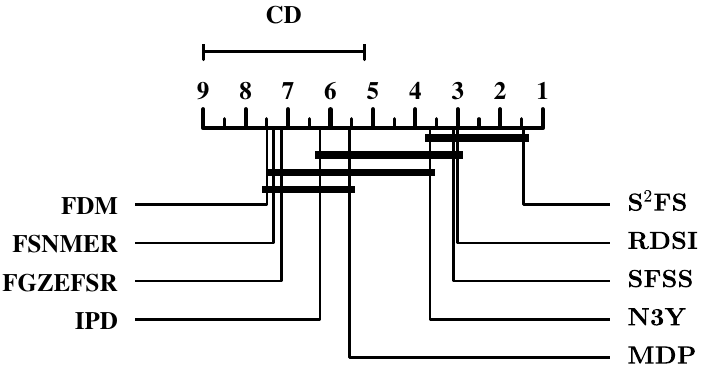}%
\label{fig:9b}}\\
\subfloat[\texttt{Max accuracy (SVM)}]{\includegraphics[width=0.24\textwidth]{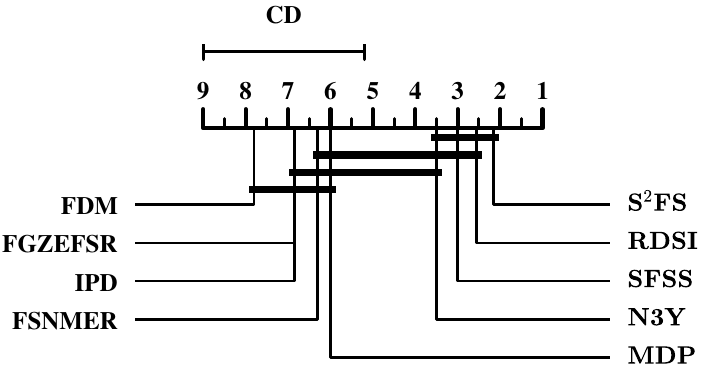}%
\label{fig:9c}}
\hspace{0.3mm}
\subfloat[\texttt{Max NMI ($k$-means)}]{\includegraphics[width=0.24\textwidth]{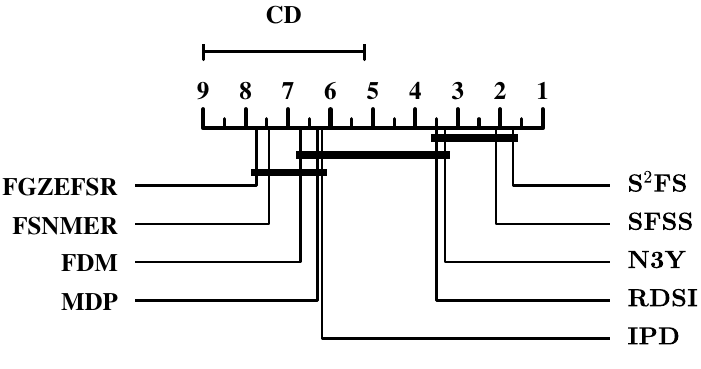}%
\label{fig:9d}}\\
\subfloat[\texttt{Ave accuracy (CART)}]{\includegraphics[width=0.24\textwidth]{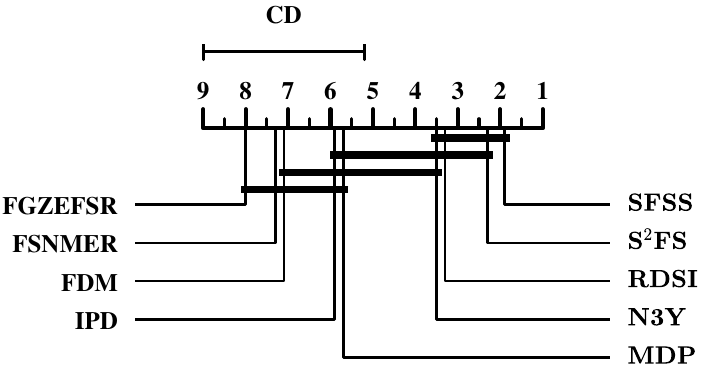}%
\label{fig:9e}}
\hspace{0.3mm}
\subfloat[\texttt{Ave accuracy (KNN)}]{\includegraphics[width=0.24\textwidth]{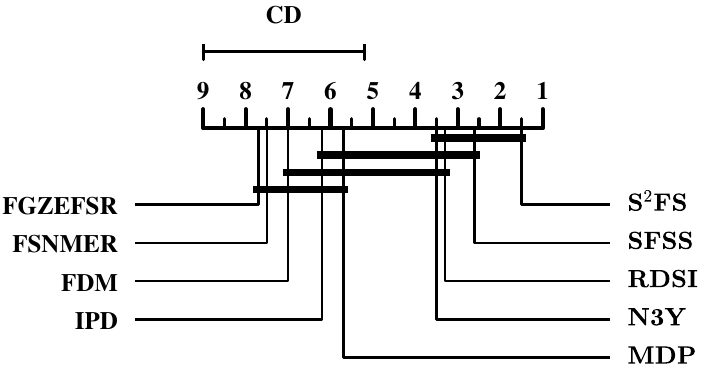}%
\label{fig:9f}}\\
\subfloat[\texttt{Ave accuracy (SVM)}]{\includegraphics[width=0.24\textwidth]{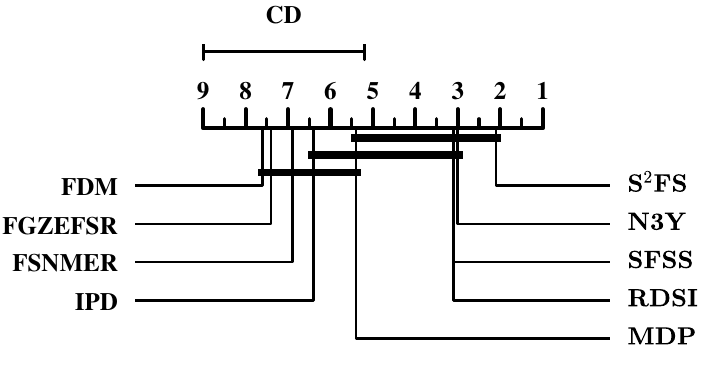}%
\label{fig:9g}}
\hspace{0.3mm}
\subfloat[\texttt{Ave NMI ($k$-means)}]{\includegraphics[width=0.24\textwidth]{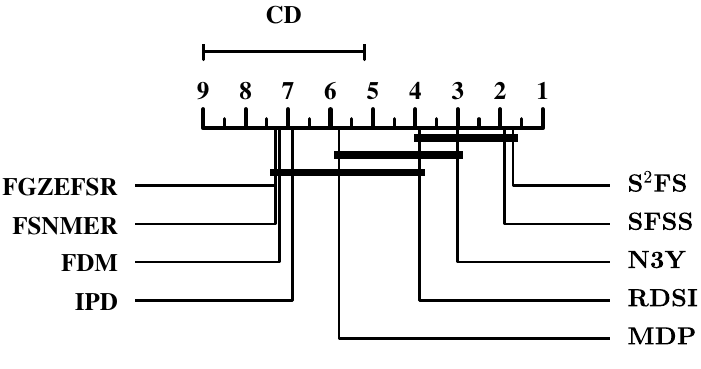}%
\label{fig:9h}}
\caption{CD diagrams of the Max and Ave performance with the top $1$, $2$,..., $150$ features.}
\label{fig:9}
\end{figure}

To visually highlight the actual differences among nine algorithms, Fig~\ref{fig:9} presents CD diagrams \cite{DemvsarJ2006} for the MAX and Ave performance, with the top $1$, $2$,..., $150$ features, of CART, KNN, SVM, and $k$-means. Algorithms are placed along a horizontal axis according to their average ranks, with higher-performing algorithms appearing to the right. Thick horizontal lines connect algorithms whose performances are not significantly different under the \emph{Nemenyi test}, with the CD value shown above each axis.

As illustrated in Fig.~\ref{fig:9}, across the $64$ pairwise predictive performance comparisons ($8$ competing algorithms $\times$ $8$ evaluation settings), S$^2$FS achieves statistically comparable performance in $42.19\%$ of the cases. Specifically, these $27$ cases include comparisons against SFSS, RDSI, and N3Y across all eight evaluation settings, against IPD under Max accuracy (CART) in Fig.~9(a), and against MDP under Max accuracy (CART) and Ave accuracy (SVM) in Figs.~9(a) and 9(g). In the remaining $57.81\%$ of the cases, S$^2$FS demonstrates statistically superior predictive performance. The only exception arises in Ave accuracy (CART), shown in Fig.~9(e), where SFSS surpasses S$^2$FS. Overall, these findings strongly confirm the superior predictive capability and robustness of S$^2$FS, while highlighting SFSS as a noteworthy competitor in the specific setting of average CART accuracy.

\subsection{Feature Visualization}

\begin{figure}[!t]
\centering
\subfloat[\texttt{Subject \#1}]{\includegraphics[width=0.445\textwidth]{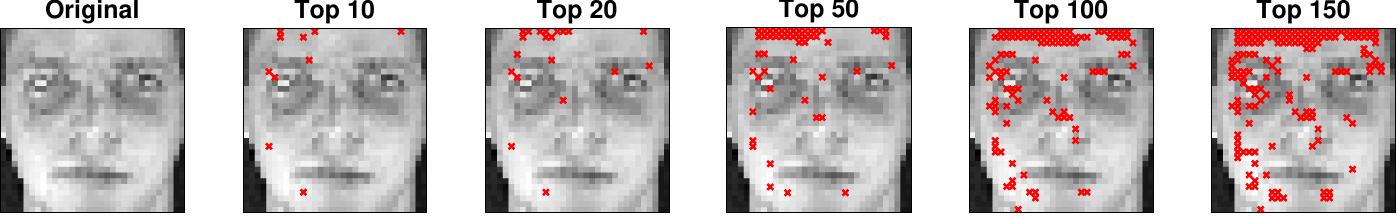}%
\label{fig:10a}}\\
\subfloat[\texttt{Subject \#2}]{\includegraphics[width=0.13\textwidth]{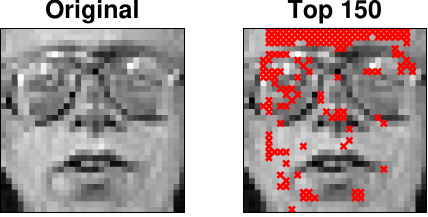}%
\label{fig:10b}}
\hfil
\subfloat[\texttt{Subject \#3}]{\includegraphics[width=0.13\textwidth]{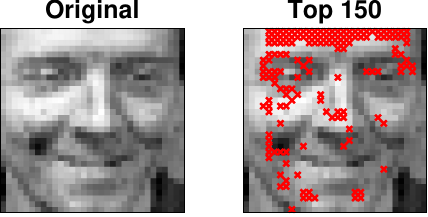}%
\label{fig:10c}}
\hfil
\subfloat[\texttt{Subject \#10}]{\includegraphics[width=0.13\textwidth]{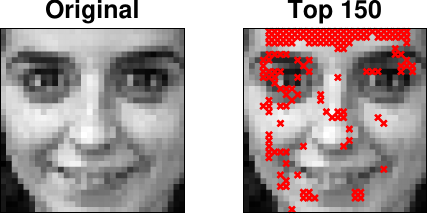}%
\label{fig:10d}}\\
\subfloat[\texttt{Subject \#20}]{\includegraphics[width=0.13\textwidth]{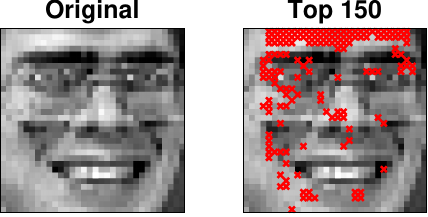}%
\label{fig:10e}}
\hfil
\subfloat[\texttt{Subject \#30}]{\includegraphics[width=0.13\textwidth]{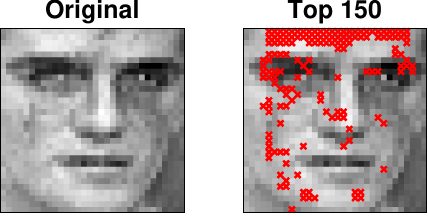}%
\label{fig:10f}}
\hfil
\subfloat[\texttt{Subject \#40}]{\includegraphics[width=0.13\textwidth]{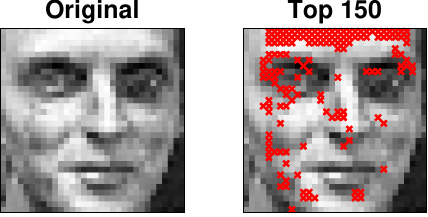}%
\label{fig:10g}}\\
\caption{Feature visualization on ORL dataset.}
\label{fig:10}
\end{figure}

\begin{figure}[!t]
\centering
\subfloat[\texttt{Subject \#1}]{\includegraphics[width=0.445\textwidth]{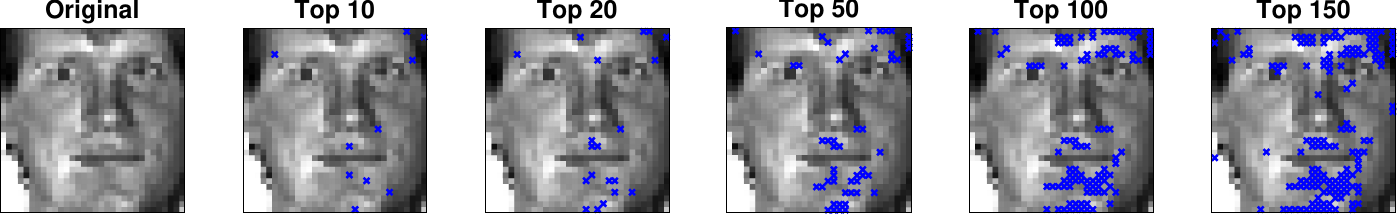}%
\label{fig:11a}}\\
\subfloat[\texttt{Subject \#2}]{\includegraphics[width=0.13\textwidth]{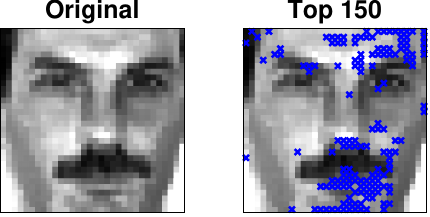}%
\label{fig:11b}}
\hfil
\subfloat[\texttt{Subject \#3}]{\includegraphics[width=0.13\textwidth]{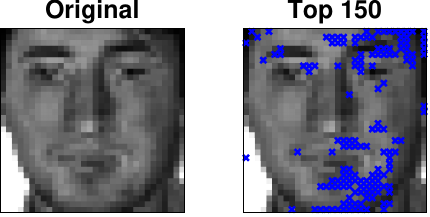}%
\label{fig:11c}}
\hfil
\subfloat[\texttt{Subject \#4}]{\includegraphics[width=0.13\textwidth]{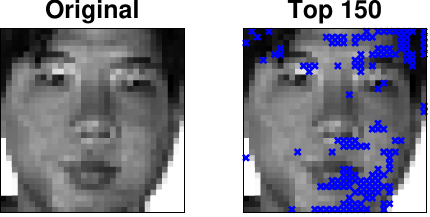}%
\label{fig:11d}}\\
\subfloat[\texttt{Subject \#5}]{\includegraphics[width=0.13\textwidth]{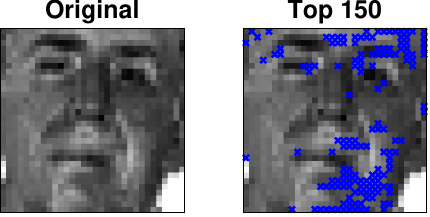}%
\label{fig:11e}}
\hfil
\subfloat[\texttt{Subject \#10}]{\includegraphics[width=0.13\textwidth]{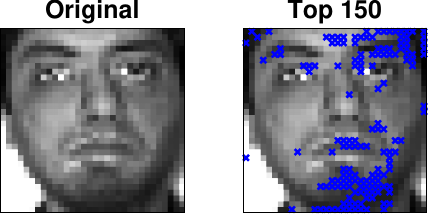}%
\label{fig:11f}}
\hfil
\subfloat[\texttt{Subject \#15}]{\includegraphics[width=0.13\textwidth]{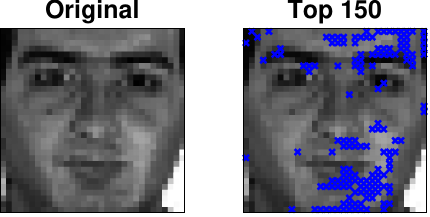}%
\label{fig:11g}}\\
\caption{Feature visualization on Yale dataset.}
\label{fig:11}
\end{figure}

On the two face recognition datasets, we further perform feature visualizations based on the features selected by S$^2$FS, as shown in Figs.~\ref{fig:10} and \ref{fig:11}, respectively. Due to space limitations, we display only the first image of seven subjects (\#1, \#2, \#3, \#10, \#20, \#30, \#40) from the ORL dataset and seven subjects (\#1, \#2, \#3, \#4, \#5, \#10, \#15) from the Yale dataset. In Figs.~10(a) and 11(a), the number of selected features varies over \{$10$, $20$, $50$, $100$, $150$\}, whereas in Figs.~10(b)-10(g) and Figs.~11(b)-11(g), the number of selected features is fixed at $150$. More detailed visualizations are provided in \textbf{Sec. A of the Supplementary Materials}.

From Figs.~\ref{fig:10} and \ref{fig:11}, particularly on the ORL dataset, it can be observed that the features selected by S$^2$FS are highly concentrated in spatially meaningful facial regions. These features predominantly cluster around distinctive sub-areas such as the eyebrows, nose, mouth, and cheeks. Moreover, S$^2$FS exhibits a clear tendency to prioritize features located along the contours of facial organs. Such a pattern is consistent with human perceptual strategies in face recognition and further confirms that contour- and edge-related features play an essential role in enhancing learning performance.

\subsection{Parameter Sensitivity Analysis}

\begin{figure}[!t]
\centering
\subfloat[\texttt{ALL-AML-3}]{\includegraphics[width=0.48\textwidth]{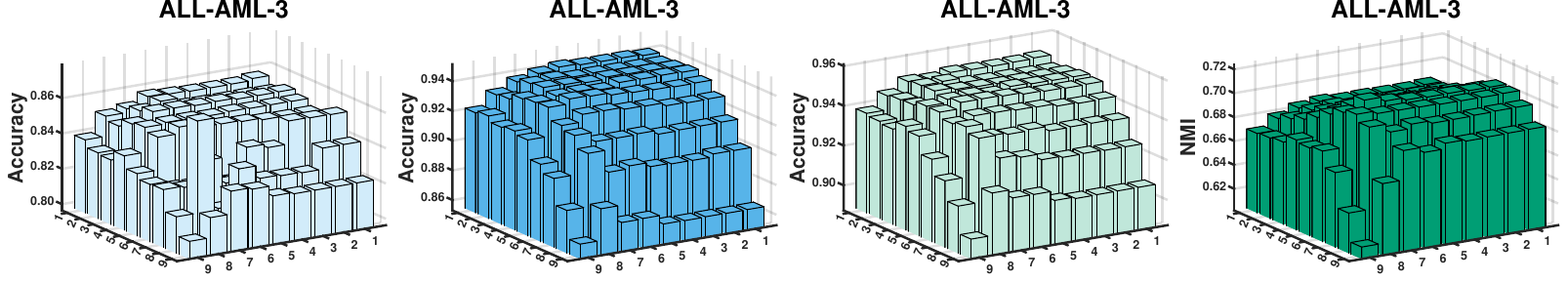}%
\label{fig:12a}}\\
\subfloat[\texttt{Isolet}]{\includegraphics[width=0.48\textwidth]{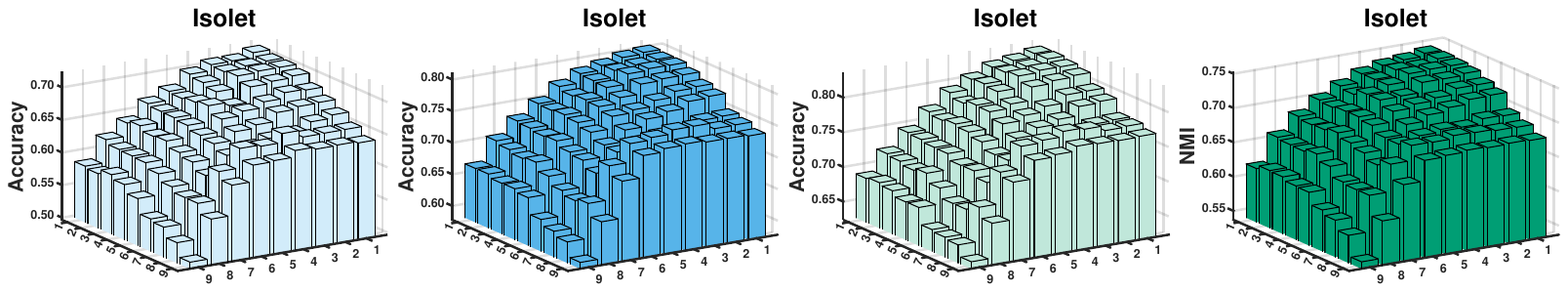}%
\label{fig:12b}}\\
\subfloat[\texttt{Lung}]{\includegraphics[width=0.48\textwidth]{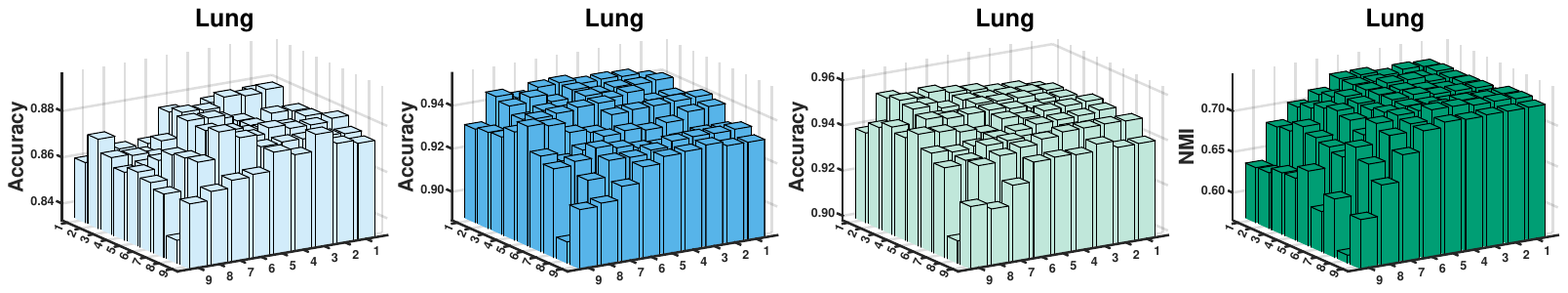}%
\label{fig:12c}}\\
\subfloat[\texttt{Yale}]{\includegraphics[width=0.48\textwidth]{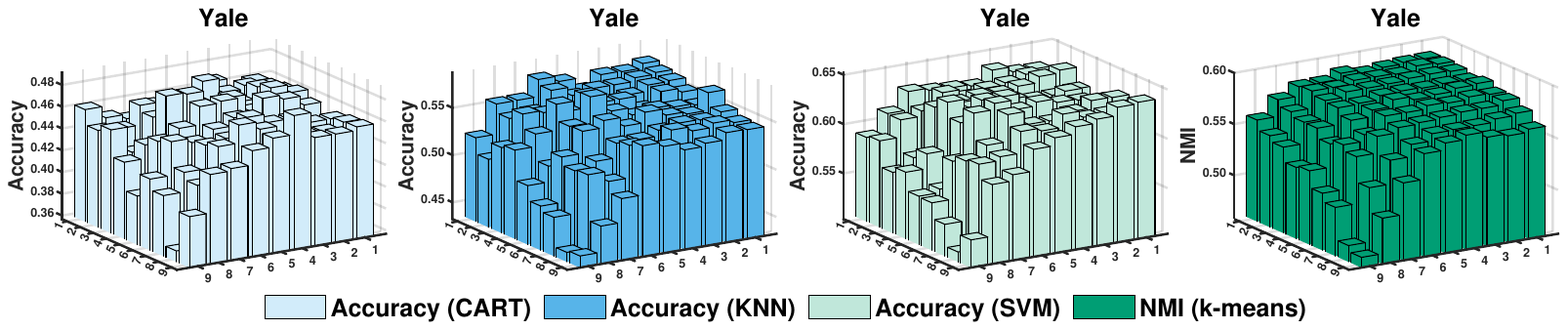}%
\label{fig:12d}}
\caption{Parameter sensitivity analysis on ALL-AML-3, Isolet, Lung, and Yale datasets.}
\label{fig:12}
\end{figure}

We conduct a parameter sensitivity analysis of S$^2$FS across all ten datasets. The balancing parameters $\alpha$ and $\beta$ in the spatially-aware separability criterion are chosen from the set \{$0.0100$, $0.0178$, $0.0316$, $0.0562$, $0.1000$, $0.1778$, $0.3162$, $0.5623$, $1.0000$\}. Due to space limitations, Fig.~\ref{fig:12} reports only the average performance of S$^2$FS with the top $1$, $2$,..., $150$ features, evaluated using the CART, KNN, and SVM classifiers as well as $k$-means clustering, on the ALL-AML-3, Isolet, Lung, and Yale datasets. Here, the scales $1$ to $9$ on the coordinate axis correspond to the parameter values in the above set. More comprehensive sensitivity analyses on the remaining six datasets are provided in \textbf{Sec. B of the Supplementary Materials}. Based on these results, we can conclude that:

(1) When $\alpha$ and $\beta$ are set to relatively small values (approximately $0.0100$–$0.0562$), S$^2$FS consistently achieves high and stable classification accuracy and clustering NMI. In contrast, larger values lead to gradual performance degradation, suggesting that spatial directional information should complement rather than dominate distance information.

(2) Although the absolute performance varies across classifiers, a consistent trend is observed: smaller parameter values yield superior results, and the similar pattern is reflected in clustering performance.

(3) For practical applications, setting $\alpha$ and $\beta$ within a relatively small range is recommended, as this not only ensures stable and competitive performance but also reduces the need for extensive parameter tuning.

\subsection{Comparison with Variants}
To further validate the effectiveness of the proposed spatially-aware separability criterion, we compare it with three variants obtained by removing specific directional components from Eq.~(\ref{eq:23}).

(1) \textbf{S$^2$--DWC} (\textbf{S}patially-aware \textbf{S}eparability criterion without \textbf{D}irectional \textbf{W}ithin-class \textbf{C}ompactness): This variant omits the directional within-class compactness term, making within-class compactness purely distance-based, while between-class separation still incorporates both distance and directional components. It is defined as:
\begin{equation}
Sep^{F'}_L = \frac{\Lambda^{F', L}_{dis} + \beta \Lambda^{F', L}_{dir}}{\Theta^{F', L}_{dis}}
\label{eq:28}
\end{equation}
 
(2) \textbf{S$^2$--DBS} (\textbf{S}patially-aware \textbf{S}eparability criterion without \textbf{D}irectional \textbf{B}etween-class \textbf{S}eparation): This variant excludes the directional between-class separation term, making between-class separation purely distance-based, while within-class compactness still incorporates both distance and directional components. It is defined as:
\begin{equation}
Sep^{F'}_L = \frac{\Lambda^{F', L}_{dis}}{\Theta^{F', L}_{dis} + \alpha \Theta^{F', L}_{dir}}
\label{eq:29}
\end{equation}

(3) \textbf{S$^2$--DWC--DBS} (\textbf{S}patially-aware \textbf{S}eparability criterion without \textbf{D}irectional \textbf{W}ithin-class \textbf{C}ompactness and \textbf{D}irectional \textbf{B}etween-class \textbf{S}eparation): This variant simultaneously removes both directional terms, reducing the separability criterion to a purely distance-based form. It is defined as:
\begin{equation}
Sep^{F'}_L = \frac{\Lambda^{F', L}_{dis}}{\Theta^{F', L}_{dis}}
\label{eq:30}
\end{equation}

\begin{figure}[!t]
\centering
\subfloat[\texttt{ALL-AML-3}]{\includegraphics[width=0.48\textwidth]{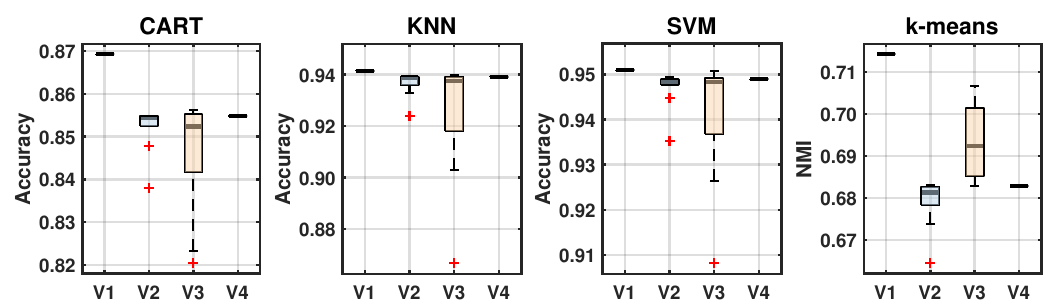}%
\label{fig:14a}}\\
\subfloat[\texttt{Isolet}]{\includegraphics[width=0.48\textwidth]{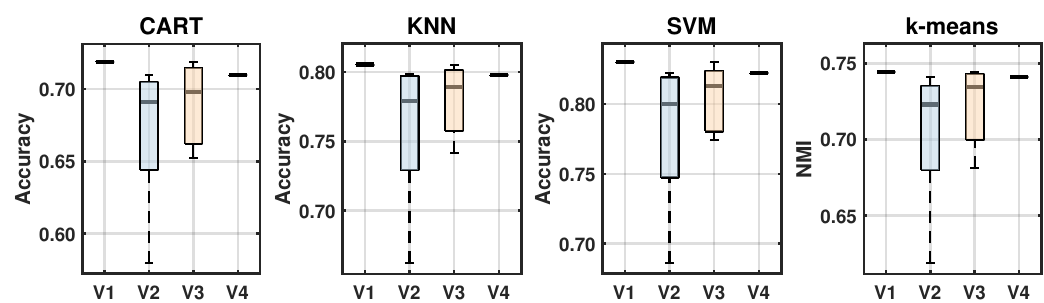}%
\label{fig:14b}}\\
\subfloat[\texttt{Lung}]{\includegraphics[width=0.48\textwidth]{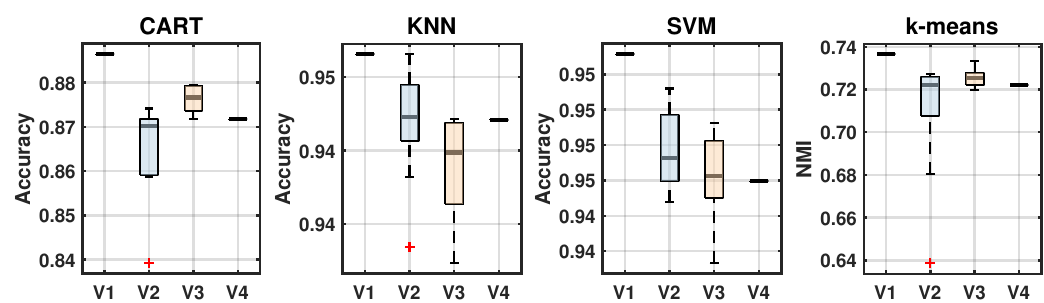}%
\label{fig:14c}}\\
\subfloat[\texttt{Yale}]{\includegraphics[width=0.48\textwidth]{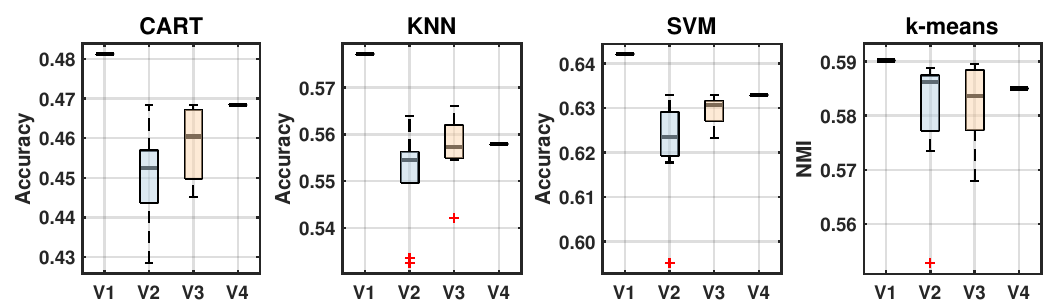}%
\label{fig:14d}}
\caption{Performance comparison of spatially-aware separability criterion (V1) and its three variants (V2: S$^2$--DWC, V3: S$^2$--DBS, V4: S$^2$--DWC--DBS) on ALL-AML-3, Isolet, Lung, and Yale datasets.}
\label{fig:13}
\end{figure}

By comparing the proposed spatially-aware separability criterion with its three variants, namely S$^2$--DWC, S$^2$--DBS, and S$^2$--DWC--DBS, we aim to isolate the effect of directional information on both within-class compactness and between-class separation. Fig.~\ref{fig:13} shows the performance of feature subsets selected by these criteria, evaluated with the CART, KNN, and SVM classifiers as well as $k$-means clustering on the ALL-AML-3, Isolet, Lung, and Yale datasets. The reported results correspond to the average performance of feature subsets containing the top $1$ to $150$ features. For the proposed criterion, we report the best performance across all combinations $(\alpha, \beta)$. For $\beta$ in S$^2$--DWC and $\alpha$ in S$^2$--DBS, values are varied within \{$0.0100$, $0.0178$, $0.0316$, $0.0562$, $0.1000$, $0.1778$, $0.3162$, $0.5623$, $1.0000$\}, and the corresponding results under the nine parameter settings are statistically analyzed. For S$^2$--DWC--DBS, which involves no parameters, only a single result is reported. Additional performance comparisons on the ALL-AML-4, GLIOMA, SRBCT, warpAR10P, warpPIE10P, and ORL datasets are provided in \textbf{Sec. C of the Supplementary Materials}.

From Fig.~\ref{fig:13}, it can be observed that removing any directional component (S$^2$--DWC, S$^2$--DBS, and S$^2$--DWC--DBS) generally leads to performance degradation, whereas the full criterion consistently achieves superior accuracy and NMI across datasets. These findings confirm that directional information plays a complementary role to distance information and is indispensable for constructing a more robust and discriminative feature selection criterion.

\subsection{Computational Complexity}

Tab.~\ref{tab:8} summarizes the computational complexity of six representative feature selection algorithms for FDSs, along with the proposed S$^2$FS. Here, $n$, $m$, and $p$ denote the numbers of instances, features, and decision classes, respectively. Compared with algorithms such as FDM ($\mathcal{O}(n^2 \cdot m^3)$) and FGZEFSR ($\mathcal{O}(n^2 \cdot m^2)$), S$^2$FS provides a more favorable trade-off between computational cost and selection effectiveness. Importantly, its complexity is of the same order as SFSS, while further incorporating spatially directional information, thereby enhancing class discriminability without incurring additional asymptotic overhead.

\begin{table}[!t]
\caption{Computational complexity}
\centering
\setlength{\tabcolsep}{10pt}
\begin{tabular}{ccccc}
\hline
Algorithms & Complexity  \\
\hline
SFSS       & \( \mathcal{O}(n \cdot m^2 \cdot p) \) \\
FDM        & \( \mathcal{O}(n^2 \cdot m^3) \) \\
FSNMER     & \( \mathcal{O}(n \cdot m \cdot (n + m)) \) \\
MDP        & \( \mathcal{O}(n^2 \cdot m) \) \\
N3Y        & \( \mathcal{O}(n^2 \cdot m \cdot p) \) \\
FGZEFSR    & \( \mathcal{O}(n^2 \cdot m^2) \) \\
S$^2$FS    & \( \mathcal{O}(n \cdot m^2 \cdot p) \) \\
\hline
\end{tabular}
\label{tab:8}
\end{table}

\section{Conclusion}
This paper proposes a spatially-aware separability-driven feature selection framework (S$^2$FS) for fuzzy decision systems. By jointly incorporating scalar-distance and spatial directional information, it introduces a novel separability criterion that simultaneously characterizes within-class compactness and between-class separation. Based on this criterion, S$^2$FS adopts a forward greedy strategy to iteratively identify the most discriminative features.

Extensive experiments on ten real-world datasets demonstrated that S$^2$FS consistently outperforms eight state-of-the-art feature selection algorithms in terms of classification accuracy and clustering NMI. Furthermore, feature visualizations on face recognition datasets confirmed the interpretability of the selected features.

\section*{Acknowledgment}

The authors would like to thank the anonymous reviewers and the editor for their constructive and valuable comments. This work is supported by the National Natural Science Foundation of China (Nos. 62076111, 51975294).

\bibliographystyle{IEEEtran}

\bibliography{bib}

\appendix

\subsection{Feature Visualization}

\begin{figure*}[htbp]
\centering

\subfloat[\texttt{Sub \#1}]{\includegraphics[width=0.15\textwidth]{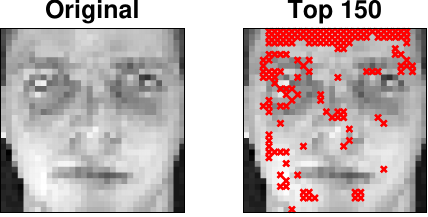}}
\hfill
\subfloat[\texttt{Sub \#2}]{\includegraphics[width=0.15\textwidth]{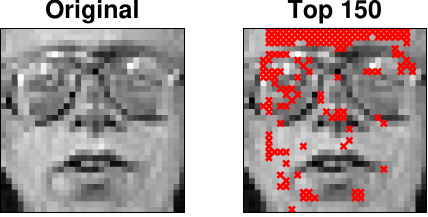}}
\hfill
\subfloat[\texttt{Sub \#3}]{\includegraphics[width=0.15\textwidth]{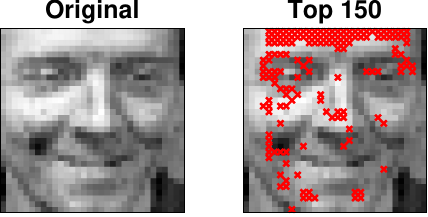}}
\hfill
\subfloat[\texttt{Sub \#4}]{\includegraphics[width=0.15\textwidth]{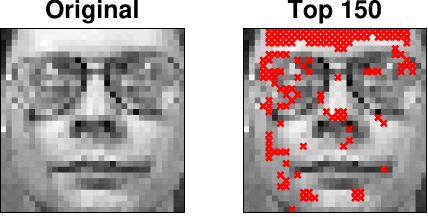}}
\hfill
\subfloat[\texttt{Sub \#5}]{\includegraphics[width=0.15\textwidth]{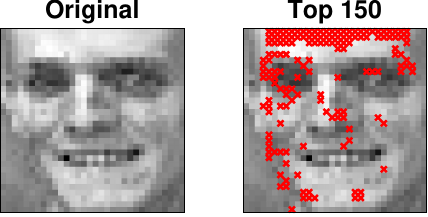}}\\

\subfloat[\texttt{Sub \#6}]{\includegraphics[width=0.15\textwidth]{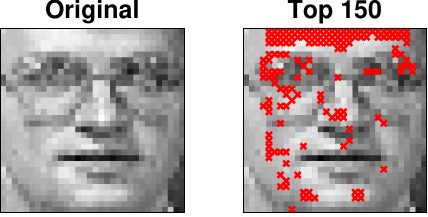}}
\hfill
\subfloat[\texttt{Sub \#7}]{\includegraphics[width=0.15\textwidth]{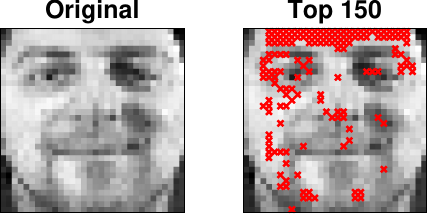}}
\hfill
\subfloat[\texttt{Sub \#8}]{\includegraphics[width=0.15\textwidth]{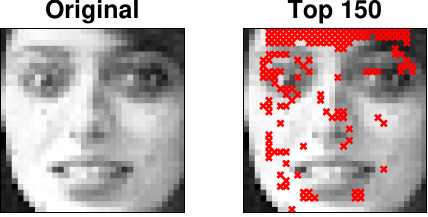}}
\hfill
\subfloat[\texttt{Sub \#9}]{\includegraphics[width=0.15\textwidth]{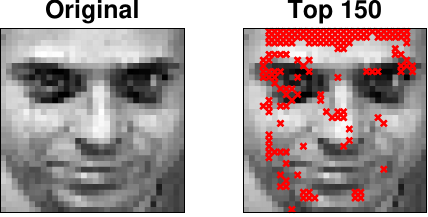}}
\hfill
\subfloat[\texttt{Sub \#10}]{\includegraphics[width=0.15\textwidth]{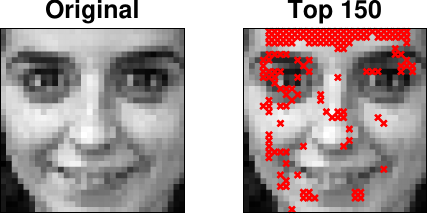}}\\

\subfloat[\texttt{Sub \#11}]{\includegraphics[width=0.15\textwidth]{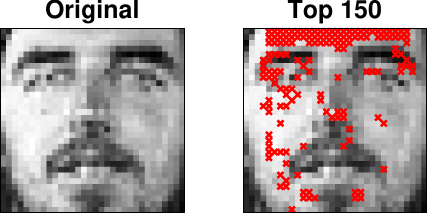}}
\hfill
\subfloat[\texttt{Sub \#12}]{\includegraphics[width=0.15\textwidth]{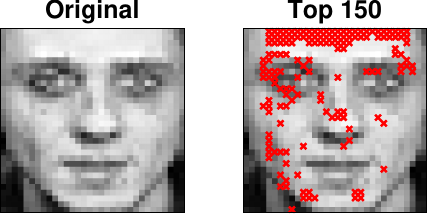}}
\hfill
\subfloat[\texttt{Sub \#13}]{\includegraphics[width=0.15\textwidth]{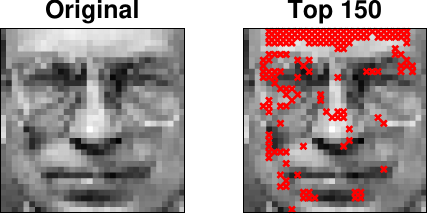}}
\hfill
\subfloat[\texttt{Sub \#14}]{\includegraphics[width=0.15\textwidth]{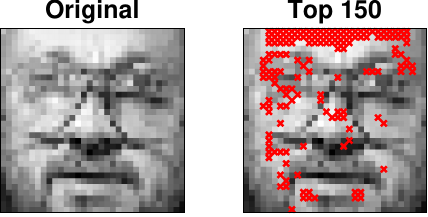}}
\hfill
\subfloat[\texttt{Sub \#15}]{\includegraphics[width=0.15\textwidth]{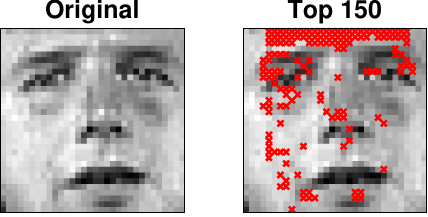}}\\

\subfloat[\texttt{Sub \#16}]{\includegraphics[width=0.15\textwidth]{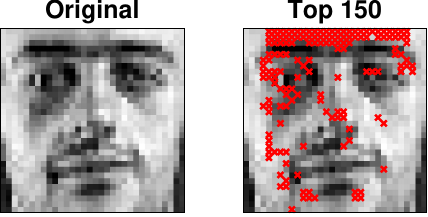}}
\hfill
\subfloat[\texttt{Sub \#17}]{\includegraphics[width=0.15\textwidth]{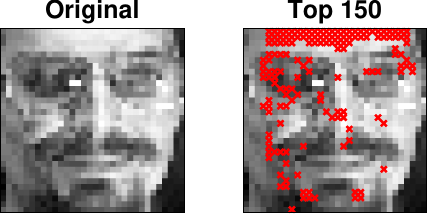}}
\hfill
\subfloat[\texttt{Sub \#18}]{\includegraphics[width=0.15\textwidth]{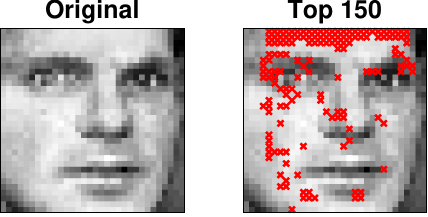}}
\hfill
\subfloat[\texttt{Sub \#19}]{\includegraphics[width=0.15\textwidth]{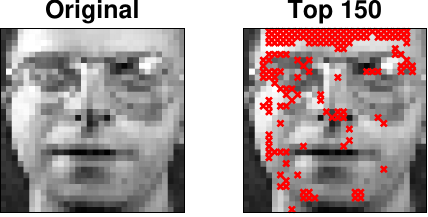}}
\hfill
\subfloat[\texttt{Sub \#20}]{\includegraphics[width=0.15\textwidth]{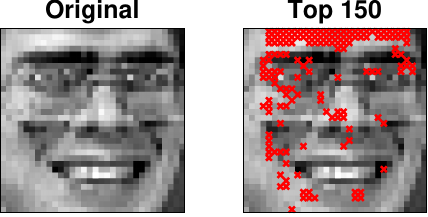}}\\

\caption{Feature visualization on ORL dataset - Part 1.}
\label{fig:14}
\end{figure*}

\begin{figure*}[htbp]
\centering

\subfloat[\texttt{Sub \#21}]{\includegraphics[width=0.15\textwidth]{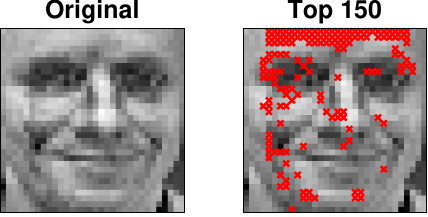}}
\hfill
\subfloat[\texttt{Sub \#22}]{\includegraphics[width=0.15\textwidth]{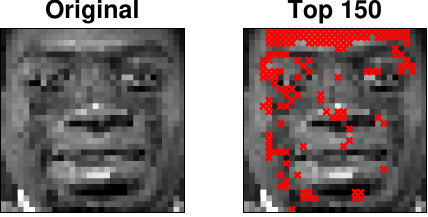}}
\hfill
\subfloat[\texttt{Sub \#23}]{\includegraphics[width=0.15\textwidth]{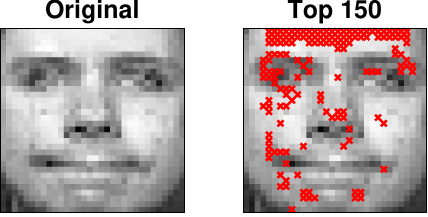}}
\hfill
\subfloat[\texttt{Sub \#24}]{\includegraphics[width=0.15\textwidth]{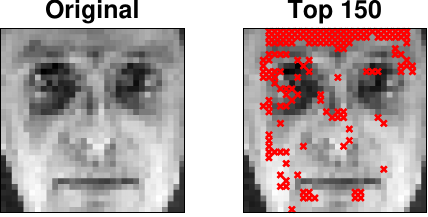}}
\hfill
\subfloat[\texttt{Sub \#25}]{\includegraphics[width=0.15\textwidth]{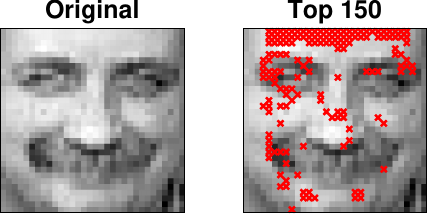}}\\

\subfloat[\texttt{Sub \#26}]{\includegraphics[width=0.15\textwidth]{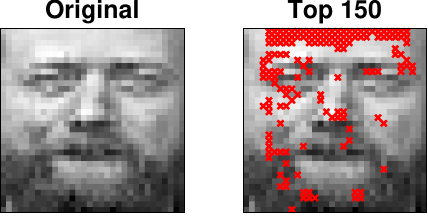}}
\hfill
\subfloat[\texttt{Sub \#27}]{\includegraphics[width=0.15\textwidth]{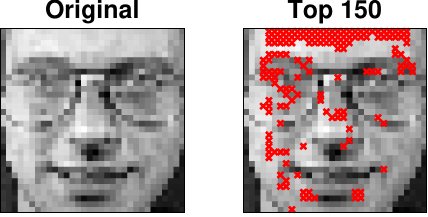}}
\hfill
\subfloat[\texttt{Sub \#28}]{\includegraphics[width=0.15\textwidth]{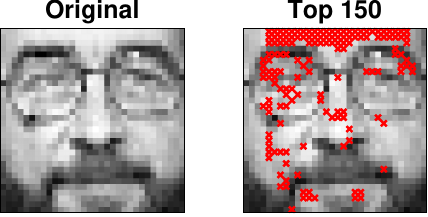}}
\hfill
\subfloat[\texttt{Sub \#29}]{\includegraphics[width=0.15\textwidth]{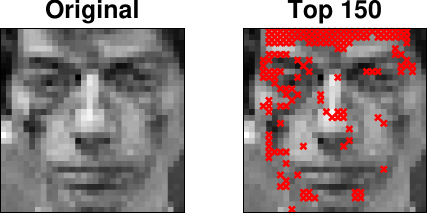}}
\hfill
\subfloat[\texttt{Sub \#30}]{\includegraphics[width=0.15\textwidth]{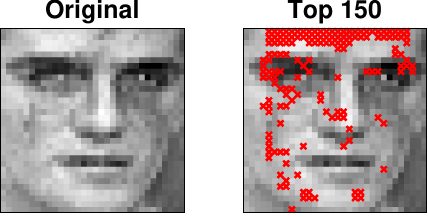}}\\

\subfloat[\texttt{Sub \#31}]{\includegraphics[width=0.15\textwidth]{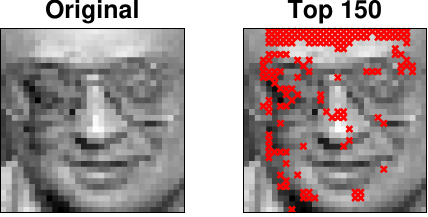}}
\hfill
\subfloat[\texttt{Sub \#32}]{\includegraphics[width=0.15\textwidth]{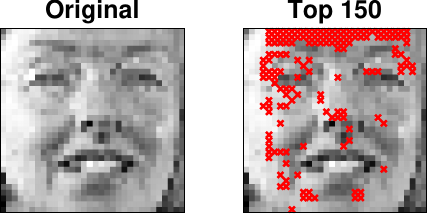}}
\hfill
\subfloat[\texttt{Sub \#33}]{\includegraphics[width=0.15\textwidth]{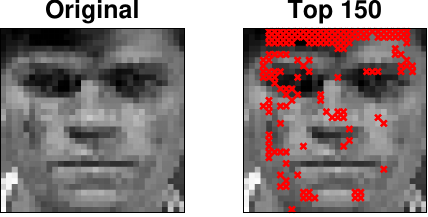}}
\hfill
\subfloat[\texttt{Sub \#34}]{\includegraphics[width=0.15\textwidth]{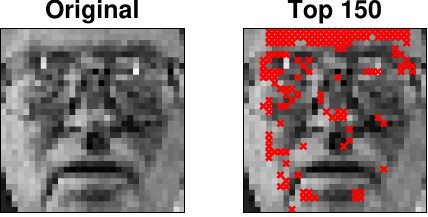}}
\hfill
\subfloat[\texttt{Sub \#35}]{\includegraphics[width=0.15\textwidth]{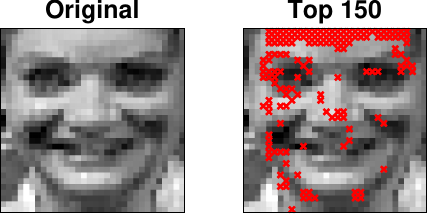}}\\

\subfloat[\texttt{Sub \#36}]{\includegraphics[width=0.15\textwidth]{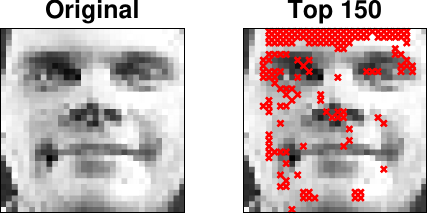}}
\hfill
\subfloat[\texttt{Sub \#37}]{\includegraphics[width=0.15\textwidth]{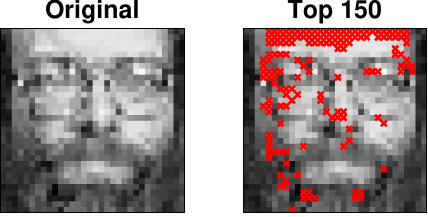}}
\hfill
\subfloat[\texttt{Sub \#38}]{\includegraphics[width=0.15\textwidth]{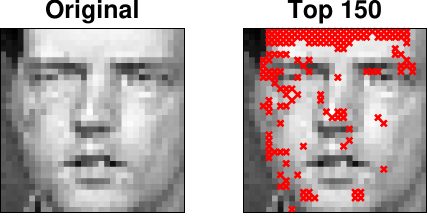}}
\hfill
\subfloat[\texttt{Sub \#39}]{\includegraphics[width=0.15\textwidth]{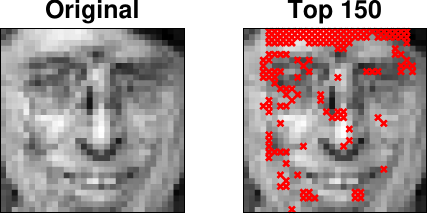}}
\hfill
\subfloat[\texttt{Sub \#40}]{\includegraphics[width=0.15\textwidth]{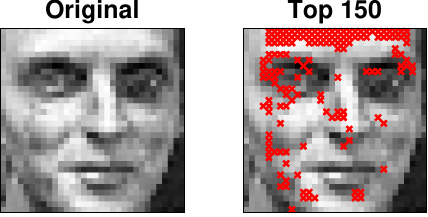}}\\

\caption{Feature visualization on ORL dataset - Part 2.}
\label{fig:15}
\end{figure*}

\begin{figure*}[htbp]
\centering

\subfloat[\texttt{Sub \#1}]{\includegraphics[width=0.15\textwidth]{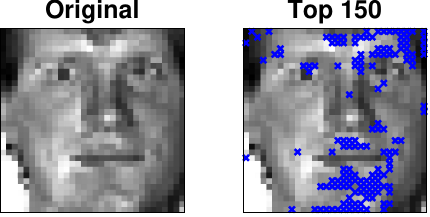}}
\hfill
\subfloat[\texttt{Sub \#2}]{\includegraphics[width=0.15\textwidth]{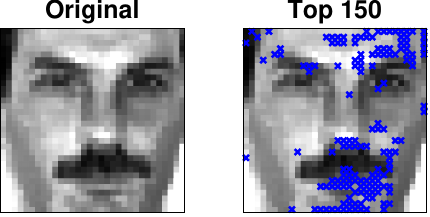}}
\hfill
\subfloat[\texttt{Sub \#3}]{\includegraphics[width=0.15\textwidth]{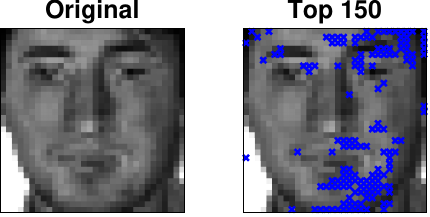}}
\hfill
\subfloat[\texttt{Sub \#4}]{\includegraphics[width=0.15\textwidth]{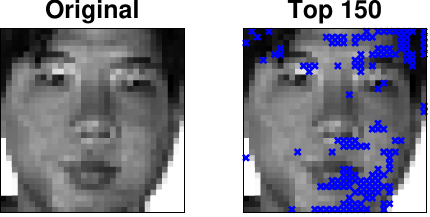}}
\hfill
\subfloat[\texttt{Sub \#5}]{\includegraphics[width=0.15\textwidth]{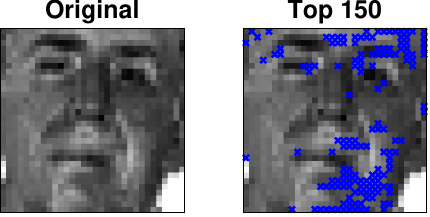}}\\

\subfloat[\texttt{Sub \#6}]{\includegraphics[width=0.15\textwidth]{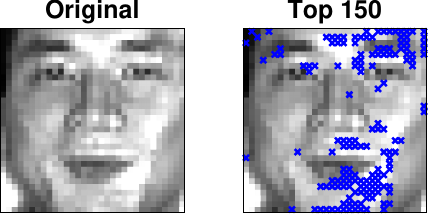}}
\hfill
\subfloat[\texttt{Sub \#7}]{\includegraphics[width=0.15\textwidth]{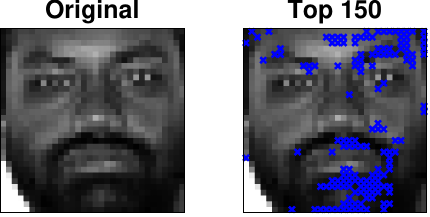}}
\hfill
\subfloat[\texttt{Sub \#8}]{\includegraphics[width=0.15\textwidth]{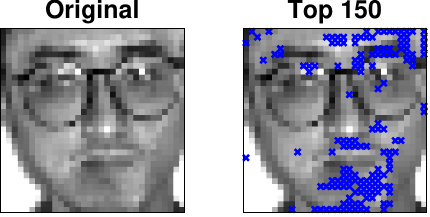}}
\hfill
\subfloat[\texttt{Sub \#9}]{\includegraphics[width=0.15\textwidth]{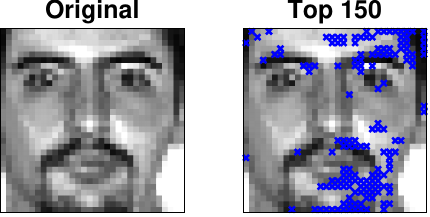}}
\hfill
\subfloat[\texttt{Sub \#10}]{\includegraphics[width=0.15\textwidth]{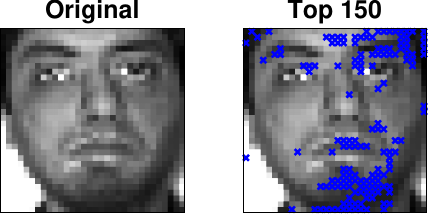}}\\

\subfloat[\texttt{Sub \#11}]{\includegraphics[width=0.15\textwidth]{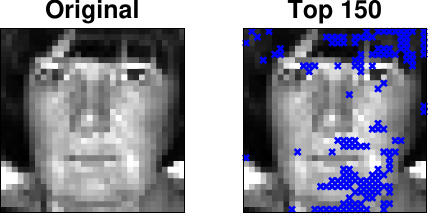}}
\hfill
\subfloat[\texttt{Sub \#12}]{\includegraphics[width=0.15\textwidth]{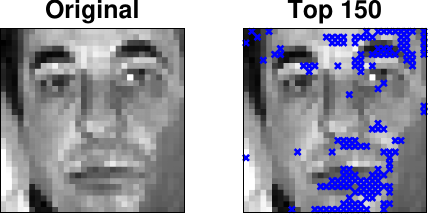}}
\hfill
\subfloat[\texttt{Sub \#13}]{\includegraphics[width=0.15\textwidth]{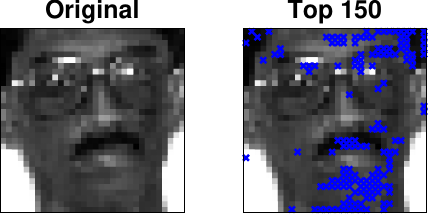}}
\hfill
\subfloat[\texttt{Sub \#14}]{\includegraphics[width=0.15\textwidth]{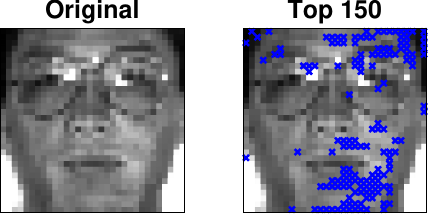}}
\hfill
\subfloat[\texttt{Sub \#15}]{\includegraphics[width=0.15\textwidth]{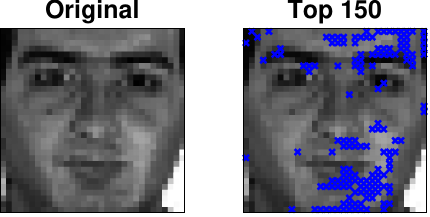}}\\

\caption{Feature visualization on Yale dataset.}
\label{fig:16}
\end{figure*}

Complete feature visualization results for the ORL and Yale datasets are presented in Figs.~\ref{fig:14}-\ref{fig:16}, respectively, based on the features selected by S$^2$FS. For each dataset, we display the first image of every subject. The number of selected features is fixed at $150$, corresponding to Figs.~14(a)-14(t), Figs.~15(a)-15(t), and Figs.~16(a)-16(o), respectively. We observe that the features selected by S$^2$FS are highly concentrated in spatially meaningful facial regions. These features predominantly cluster around distinctive sub-areas such as the eyebrows, nose, mouth, and cheeks.

\subsection{Parameter Sensitivity Analysis}

\begin{figure*}[!t]
\centering
\subfloat[\texttt{ALL-AML-4}]{\includegraphics[width=0.85\textwidth]{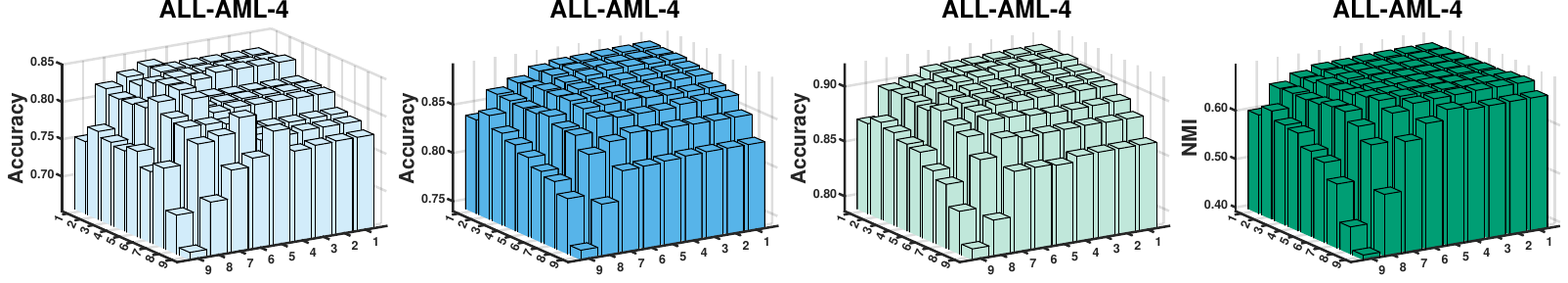}%
\label{fig:17a}}\\
\subfloat[\texttt{GLIOMA}]{\includegraphics[width=0.85\textwidth]{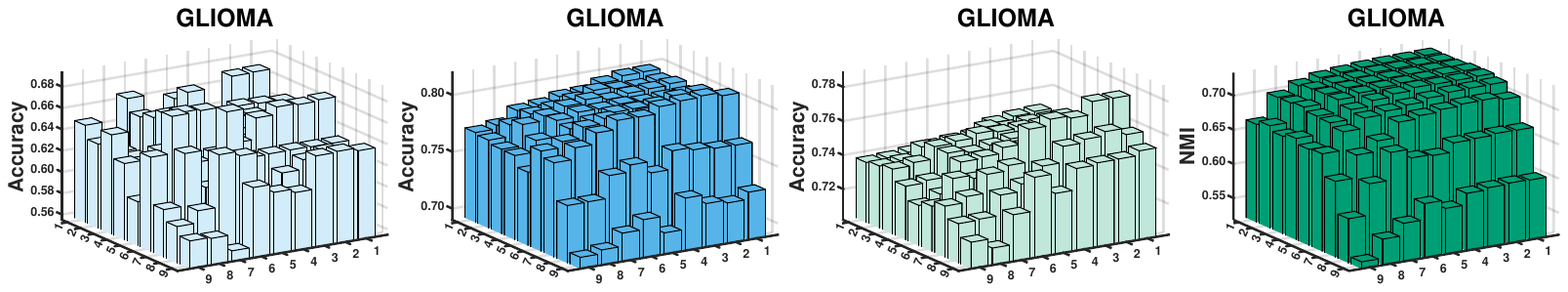}%
\label{fig:17b}}\\
\subfloat[\texttt{SRBCT}]{\includegraphics[width=0.85\textwidth]{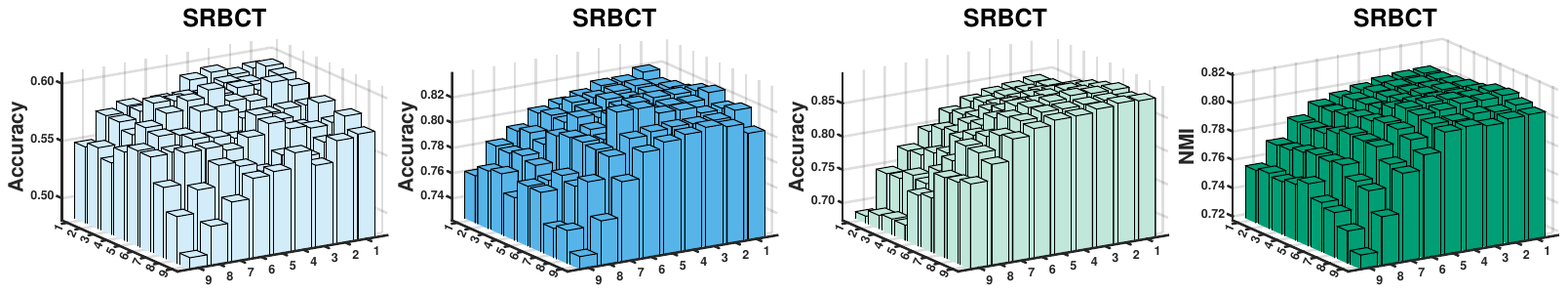}%
\label{fig:17c}}\\
\subfloat[\texttt{warpAR10P}]{\includegraphics[width=0.85\textwidth]{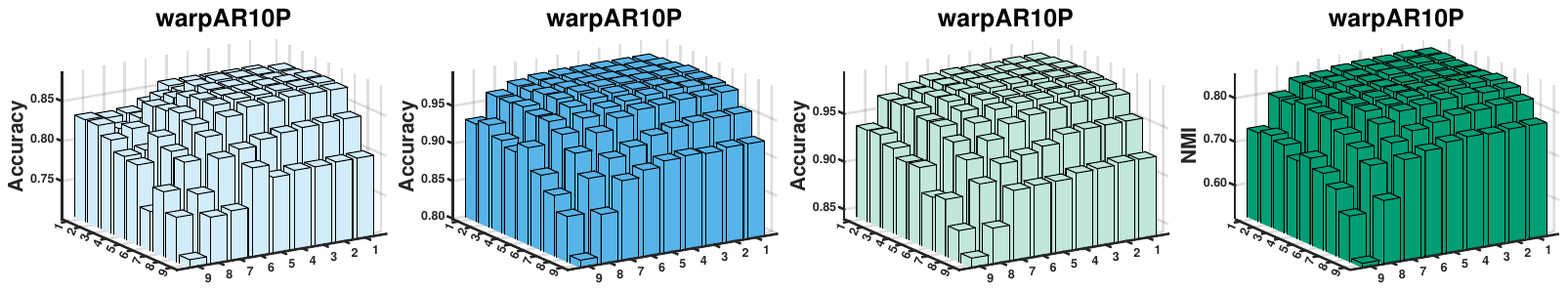}%
\label{fig:17d}}\\
\subfloat[\texttt{warpPIE10P}]{\includegraphics[width=0.85\textwidth]{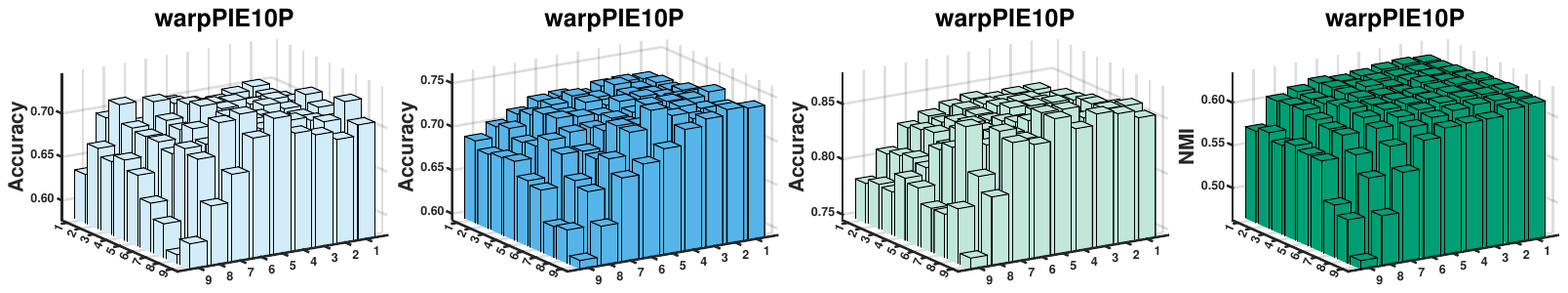}%
\label{fig:17e}}\\
\subfloat[\texttt{ORL}]{\includegraphics[width=0.85\textwidth]{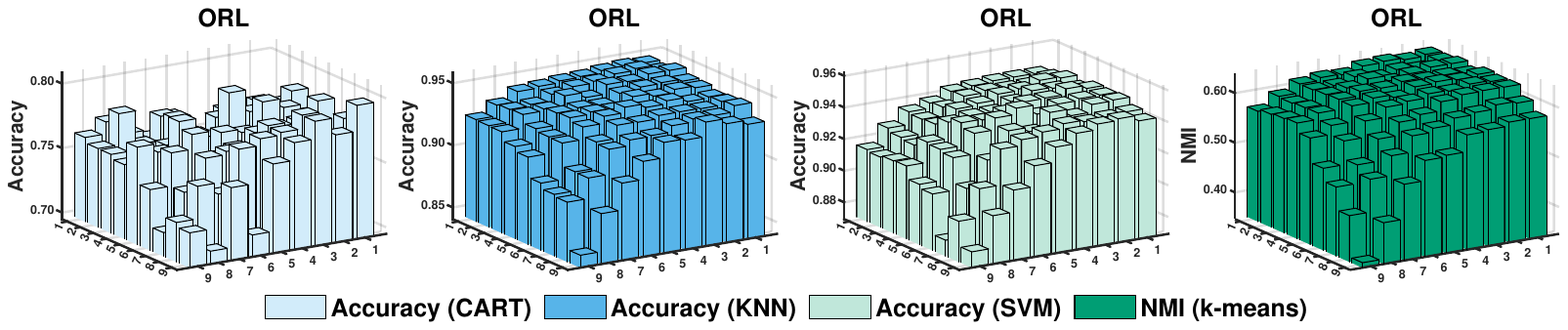}%
\label{fig:17f}}
\caption{Parameter sensitivity analysis on ALL-AML-4, GLIOMA, SRBCT, warpAR10P, warpPIE10P, and ORL datasets.}
\label{fig:17}
\end{figure*}

The parameter sensitivity analyses on the ALL-AML-4, GLIOMA, SRBCT, warpAR10P, warpPIE10P, and ORL datasets are presented in Fig.~\ref{fig:17}. The results lead to conclusions similar to those observed on the ALL-AML-3, Isolet, Lung, and Yale datasets.

\subsection{Comparison with Variants}

\textbf{S$^2$--DWC}: \textbf{S}patially-aware \textbf{S}eparability criterion without \textbf{D}irectional \textbf{W}ithin-class \textbf{C}ompactness;

\textbf{S$^2$--DBS}: \textbf{S}patially-aware \textbf{S}eparability criterion without \textbf{D}irectional \textbf{B}etween-class \textbf{S}eparation;

\textbf{S$^2$--DWC--DBS}: \textbf{S}patially-aware \textbf{S}eparability criterion without \textbf{D}irectional \textbf{W}ithin-class \textbf{C}ompactness and \textbf{D}irectional \textbf{B}etween-class \textbf{S}eparation.

\begin{figure*}[!t]
\centering
\subfloat[\texttt{ALL-AML-4}]{\includegraphics[width=0.48\textwidth]{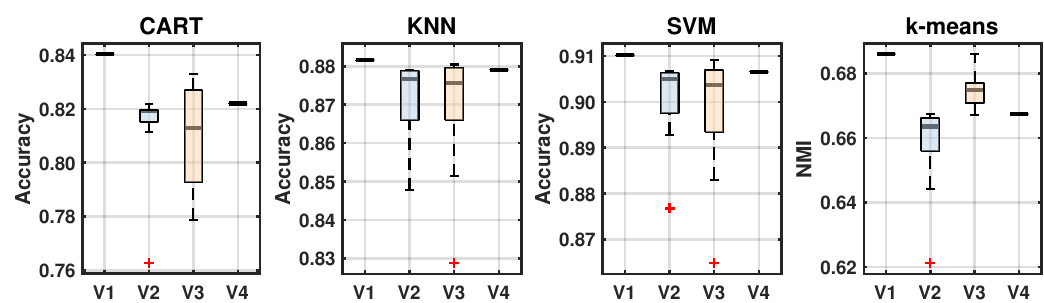}%
\label{fig:18a}}
\hfill
\subfloat[\texttt{GLIOMA}]{\includegraphics[width=0.48\textwidth]{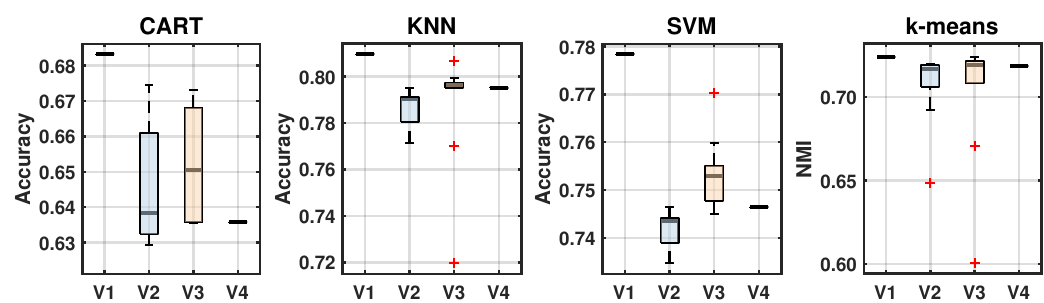}%
\label{fig:18b}}\\
\subfloat[\texttt{SRBCT}]{\includegraphics[width=0.48\textwidth]{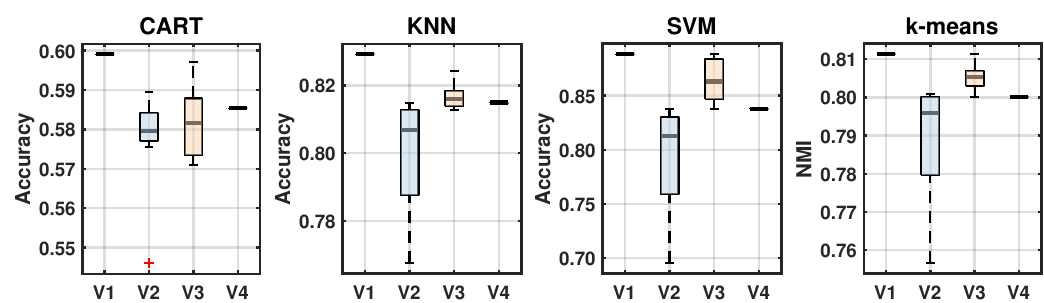}%
\label{fig:18c}}
\hfill
\subfloat[\texttt{warpAR10P}]{\includegraphics[width=0.48\textwidth]{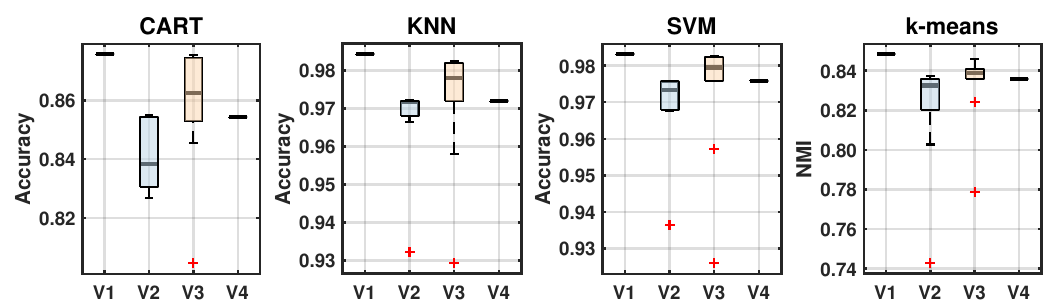}%
\label{fig:18d}}\\
\subfloat[\texttt{warpPIE10P}]{\includegraphics[width=0.48\textwidth]{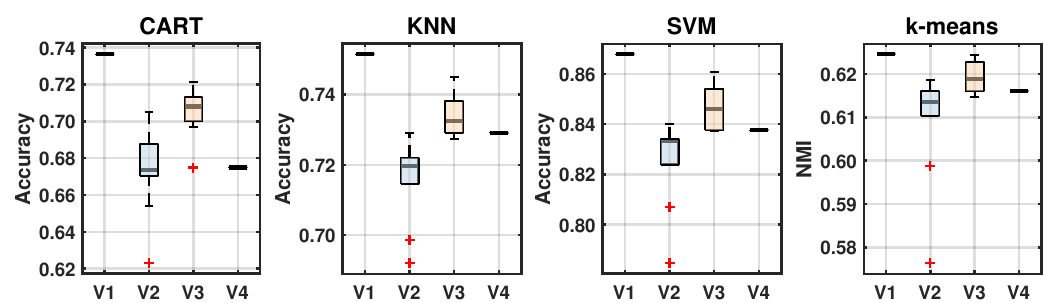}%
\label{fig:18e}}
\hfill
\subfloat[\texttt{ORL}]{\includegraphics[width=0.48\textwidth]{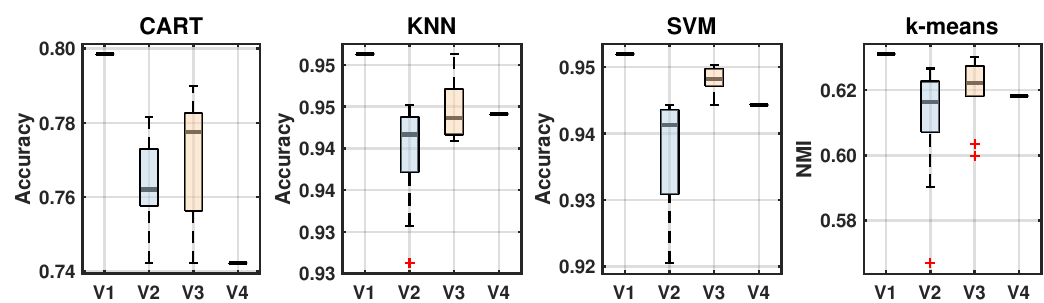}%
\label{fig:18f}}
\caption{Performance comparison of spatially-aware separability criterion (V1) and its three variants (V2: S$^2$--DWC, V3: S$^2$--DBS, V4: S$^2$--DWC--DBS) on ALL-AML-4, GLIOMA, SRBCT, warpAR10P, warpPIE10P, and ORL datasets.}
\label{fig:18}
\end{figure*}

The performance comparisons of spatially-aware separability criterion (V1) and its three variants (V2: S$^2$--DWC, V3: S$^2$--DBS, V4: S$^2$--DWC--DBS) on the ALL-AML-4, GLIOMA, SRBCT, warpAR10P, warpPIE10P, and ORL datasets are presented in Fig.~\ref{fig:18}. The results lead to conclusions similar to those observed on the ALL-AML-3, Isolet, Lung, and Yale datasets.

\end{document}